\documentclass[10pt,twocolumn,letterpaper]{article}

\usepackage{titling}
\usepackage{wacv}
\usepackage{times}
\usepackage{epsfig}
\usepackage{graphicx}
\usepackage{amsmath}
\usepackage{amssymb}
\usepackage{xcolor}
\usepackage{mathtools}
\usepackage{paralist}

\usepackage{multirow}
\usepackage{booktabs}
\usepackage{caption}
\usepackage{nicefrac, xfrac}
\usepackage{subfig}
\usepackage{colortbl}
\usepackage{arydshln}

\usepackage{algorithm}
\usepackage{algorithmic}

\newlength\paramargin
\newlength\figmargin
\newlength\subfigmargin
\newlength\secmargin
\newlength\subsecmargin
\newlength\tabmargin
\newlength\eqmargin
\newcommand {\walon}[1]{{\color{cyan}#1}\normalfont}

\newcommand{\comment}[1]{}
\newlength\myheight
\newlength\mydepth
\settototalheight\myheight{Xygp}
\def\wacvPaperID{150} 

\wacvalgorithmstrack   

\wacvfinalcopy 

\ifwacvfinal
\usepackage[breaklinks=true,bookmarks=false]{hyperref}
\else
\usepackage[pagebackref=true,breaklinks=true,colorlinks,bookmarks=false]{hyperref}
\fi

\newcommand{\modelName}{DiSS\xspace}

\usepackage{authblk}

\makeatletter
\renewcommand\AB@affilsepx{, \protect\Affilfont}
\makeatother

\begin{document}

\title{Adaptively-Realistic Image Generation \\from Stroke and Sketch with Diffusion Model}
\author[1]{Shin-I Cheng\protect\footnotemark[1]}
\author[1]{Yu-Jie Chen\protect\footnotemark[1]} 
\author[1]{Wei-Chen Chiu}
\author[2]{Hung-Yu Tseng}
\author[3]{Hsin-Ying Lee}
\affil[1]{National Chiao Tung University, Taiwan}
\affil[2]{Meta}
\affil[3]{Snap Inc.}

\twocolumn[{%
\renewcommand\twocolumn[1][]{#1}%
\maketitle

\begin{center}
    \vspace{-1em}
    \captionsetup{type=figure}
    \includegraphics[width=\linewidth]{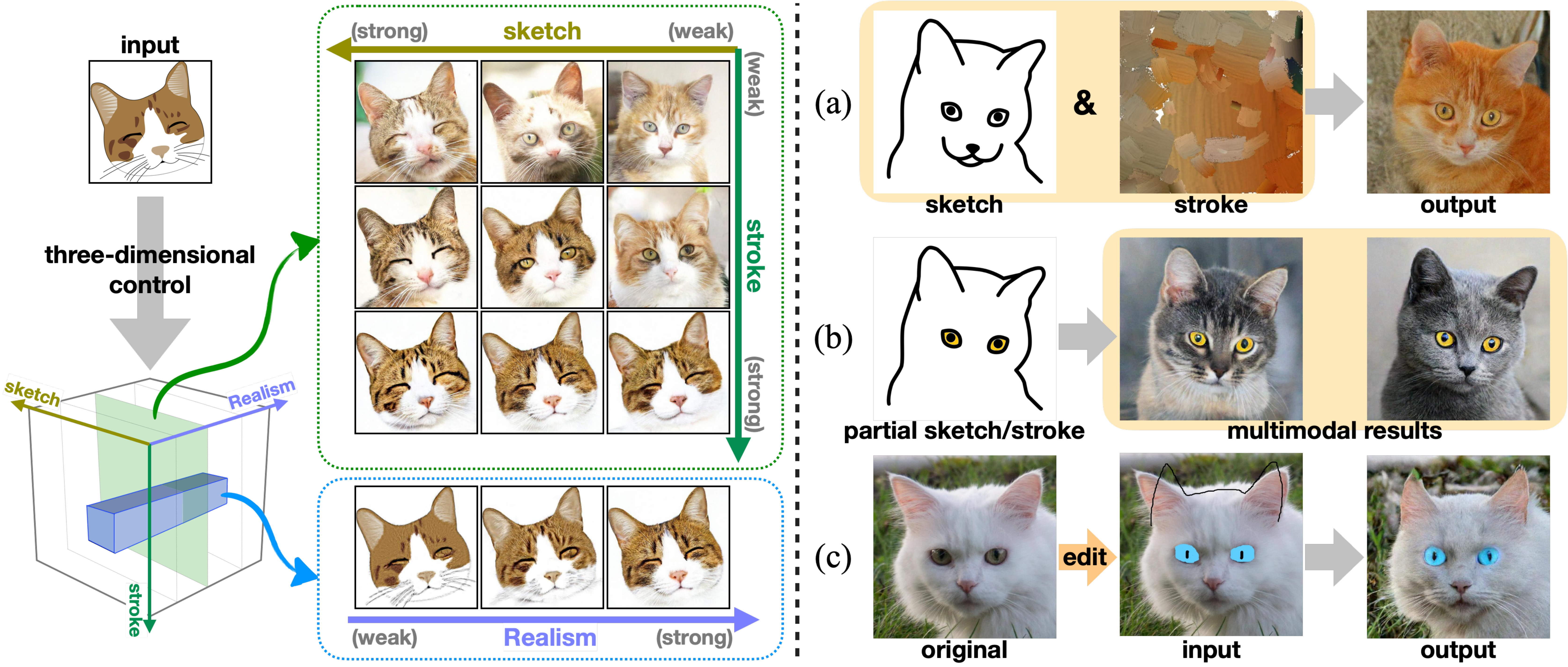}
    \captionof{figure}{
    \textbf{Three-dimension controls of image generation from stroke and sketch.} 
    {\color{black}
    (\textit{left}) Our proposed model is able provide three-dimension controls over image synthesis from stroke and sketch.
    Given sketch and stroke as input, we can control the scales of faithfulness for the synthesized output with respect to the sketch and stroke, as well as the degree of its realism.
    (\textit{right}) (a) Given sketch and strokes, we perform sketch/stroke-to-image translation.
    (b) We generate multimodal results with partial sketch/strokes as input.
    (c) Sketch/strokes conditioned local editing.
    }}
    \label{fig:teaser}
\end{center}%
}]


\renewcommand{\thefootnote}{\fnsymbol{footnote}}
\footnotetext[1]{Equal contribution.
\newline Project page: \href{https://cyj407.github.io/DiSS/}{https://cyj407.github.io/DiSS/}}

\begin{abstract}
Generating images from hand-drawings is a crucial and fundamental task in content creation. 
The translation is difficult as there exist infinite possibilities and the different users usually expect different outcomes. 
Therefore, we propose a unified framework supporting a three-dimensional control over the image synthesis from sketches and strokes based on diffusion models.
Users can not only decide the level of faithfulness to the input strokes and sketches, but also the degree of realism, as the user inputs are usually not consistent with the real images.
Qualitative and quantitative experiments demonstrate that our framework achieves state-of-the-art performance while providing flexibility in generating customized images with control over shape, color, and realism. 
Moreover, our method unleashes applications such as editing on real images, generation with partial sketches and strokes, and multi-domain multi-modal synthesis.
\comment{
{\color{blue} Current sketch/stroke-to-image methods mainly deal with either black-white sketches or colored-strokes, failing to simultaneously consider comprehensive factors involved in practical hand-drawings such as contour, color and realism. 
In this work, we propose introducing a three-scale controlled colorful sketch-to-image diffusion model considering the three important factors. 
The translation of shape and color information into image is handled with the two-directional classifier-free guidance while an additional iterative latent variable refinement is responsible for the trade-off between the realism and the consistency with respect to the user-input. 
Qualitative and quantitative experiments demonstrate that our framework achieves the state-of-the-art performance on colorful sketch-to-image task and presents the flexibility on generating customized images with the adjustment of shape, color, and consistency. 
Moreover, our method unleashes several valuable applications such as multi-conditioned local editing, region-sensitive stroke-to-images, and multi-domain sketch-to-images.}}
\end{abstract}

\vspace{-1em}
\section{Introduction}
\vspace{\secmargin}
Sketches and strokes are abstract depictions of objects and scenes. They represent different abstract illustrations that people have in mind and thus serve as important communication mediums.
Conceivably, image synthesis from hand-drawn inputs can bridge human creations with reality, unleashing potential applications and assistance toward the content creation process.

Image generation from sketches and strokes is difficult as the translation is ill-defined and multimodal.
For each sketch and stroke, different users will expect different outputs under different circumstances in terms of how faithful the results should be to the given inputs. 
Initially, the problem is formulated as image-to-image translation~\cite{sangkloy2017scribbler, chen2018sketchygan, liu2020unsupervised, wang2021sketch,sangkloy2017scribbler} with the help of generative adversarial networks (GANs)~\cite{goodfellow2014generative}.
This stream of works is usually task-dependent, requiring different models for various tasks (e.g. separate models for stroke-to-image and sketch-to-image translations).
Moreover, they lack flexibility and controllability in terms of the degree of faithfulness.
Recently, diffusion models~\cite{nichol2021improved,dhariwal2021diffusion} shed light on the tasks with high-quality image synthesis and stable training procedures.
Variants of diffusion models are proposed to handle conditions in different forms, such as category~\cite{dhariwal2021diffusion, ho2021classifier}, reference image~\cite{choi2021ilvr}, and stroke-based painting~\cite{meng2021sdedit}.
Therefore, utilizing diffusion models, we would like to explore the possibility of a unified framework that can consider all factors of interest, including contour, colors, consistency, and realism. 

In this paper, we introduce \modelName, a \textbf{DI}ffusion-based framework that generates images from \textbf{S}ketches and \textbf{S}trokes while enabling a three-dimensional (contour, color, realism) control over the degree of consistency to the input.
First, unlike previous works using either black-white sketches or stroke paintings, we propose to handle both factors simultaneously, which is not trivial because it often comes with a trade-off between faithfulness to shape and to color. 
To provide disentangled control for the consistency of sketch and stroke, we adopt classifier-free guidance~\cite{ho2021classifier} to support two-dimensional control. 
Upon disentangling the shape and color information, we can customize the generative process and separately adjust the sampling results depending on users' demands.
However, the input strokes and sketches from general users are often inconsistent with the distribution of real images.
Therefore, we propose the third control factor, the realism scale, to realize a trade-off between consistency and realism. Specifically, we apply iterative latent variable refinement~\cite{choi2021ilvr} and utilize a low-pass filter to adjust the coarse-to-fine features of the referred drawings.

With three-dimensional control, the proposed \modelName provides flexible editability, as shown in Fig.~\ref{fig:teaser}.
Users can decide to what extent the faithfulness should be to the input sketch and strokes, and to what degree the results are close to real images. 
\modelName naturally unleashes several applications.
First, multi-modal multi-domain translation (Fig.~\ref{fig:visual-multi}) can generate diverse results in multiple domains guided only by sketches and strokes without explicit labels.
Second, multi-conditioned local editing (Fig.~\ref{fig:visual-app}(a)) enables users to edit existing images by simply drawing contours and colors.
Third, region-sensitive stroke-to-images (Fig.~\ref{fig:visual-app}(b)) supports inputs that are not fully colored and provide variations on the blank regions.

We evaluate the proposed framework quantitatively and qualitatively on the three-dimension controllability.
We measure the realism and perceptual quality with Fréchet inception distance (FID)~\cite{heusel2017gans}, LPIPS~\cite{zhang2018unreasonable}, and subjective study on the AFHQ~\cite{choi2020stargan}, Oxford Flowers~\cite{nilsback2008automated}, and Landscapes-HQ~\cite{skorokhodov2021aligning} datasets.
Qualitatively, we present the diverse image synthesis conditioned on different kinds of drawings and demonstrate the adjustment of the three scales.
%

We summarize our contributions as follows:
 We present a unified framework of adaptively-realistic image generation from stroke and sketch that encodes the condition of the given stroke and sketch with the classifier-free guidance mechanism and adjusts the degree of realism with a latent variable refinement technique.
The proposed framework enables a three-dimensional control over image synthesis with flexibility and controllability over shape, color, and realism of the generation, given the input stroke and sketch. 
Moreover, our proposed work unleashes several interesting applications: multi-conditioned local editing, region-sensitive stroke-to-image, and multi-domain sketch-to-image.
\begin{figure*}[t]
    \centering
    \includegraphics[width=\linewidth]{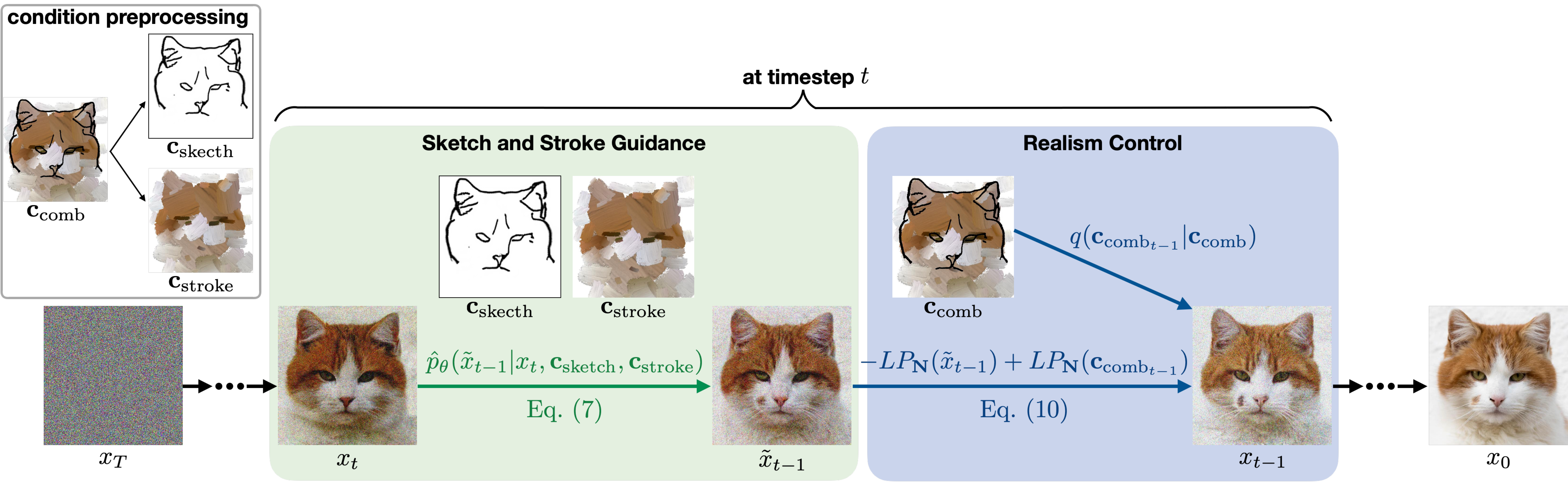}
    \vspace{-2em}
    \caption{
    \textbf{Conditional denoising process .} At each time-step $t$, our proposed pipeline first performs classifier-free guidance with $\textbf{c}_\text{sketch}$ and $\textbf{c}_\text{stroke}$, which are extracted from a single input of colorful drawing $\textbf{c}_\text{comb}$, and then controls the fidelity/realism by refining $x_{t-1}$ with the input $\textbf{c}_\text{comb}$, in which such realism control is realized by iterative latent variable refinement.
    }
    \label{fig:method}
    \vspace{\figmargin}
\end{figure*}
\vspace{-1em}
\section{Related Works}
\vspace{\secmargin}
\subsection{Image Generation from Hand-drawings}
\vspace{\subsecmargin}
Sketch-to-Image (S2I) generation aims at learning the mapping and eliminating the domain gap between hand-sketches and real images, which is usually modeled as an image-to-image translation task.
Early works on image-to-image translation~\cite{isola2017image,zhu2017unpaired,DRIT,zhu2017toward, cheng2020segvae,park2019SPADE} learn to map machine-generated edge maps or segmentation maps, where the distributions are quite different from real-world hand-drawn sketches ans scribbles, to real images.
In turn, Scribbler~\cite{sangkloy2017scribbler} and SketchyGAN~\cite{chen2018sketchygan} are among the pioneering works to specifically tackle the translation upon sketch inputs.
However, the training procedure requires datasets composed of various paired sketch-image data, which is not only hard to collect but also potentially limits the resultant translation model to tackle the general misalignment between sketches and images other than the one being seen in training set.
More recently, efforts are made to address the S2I task via unsupervised and self-supervised learning, where the gray-scale version of photos serve as an auxiliary intermediate representation~\cite{liu2020unsupervised} or an autoencoder is adopted to learn disentangled style and content factors~\cite{liu2021self}.
However, these GAN-based methods suffer from unstable training and quality.
Moreover, hand-drawings can be decomposed into sketches that model contours and strokes that model colors, yet the GAN-based models are usually task-specific, where different pipelines are adopted for different settings. Finally, it is difficult for GAN-based models to control the level of faithfulness to the input, which is a crucial property of S2I. 
Therefore, inspired by the recent success of diffusion models~\cite{ho2020denoising,dhariwal2021diffusion}, we propose a diffusion-based framework supporting flexible controls over the level of faithfulness to shapes, colors, and realism.
\subsection{Diffusion Models}
\vspace{\subsecmargin}
Diffusion models are flourishing in recent years as a powerful family of generative models with diversity, training stability, and easy scalability, which GANs commonly lack.
Fundamentally, diffusion models fulfill the sampling from a target distribution by reversing a progressive noise diffusion process, in which the process is defined as a Markov chain of diffusion steps for adding noise to the data. 
In addition to providing competitive or even superior capability on unconditional image generation~\cite{ho2020denoising,nichol2021improved} in comparison to GANs, diffusion models also make significant progress on various tasks of conditional generation. 
%
Given a target class, \cite{dhariwal2021diffusion} proposes a classifier-guidance mechanism that adopts a pretrained classifier to provide gradients as guidance toward generating images of the target class. 
More recently, \textit{classifier-free diffusion guidance}\cite{ho2021classifier} introduces a technique which jointly trains a conditional and an unconditional diffusion model without any pretrained classifier.
Other than directly modifying the network of an unconditional diffusion model for conditional generation, ILVR~\cite{choi2021ilvr} instead proposes to iteratively introduce the condition into the generative process via refining the intermediate latent images with a noisy reference image at each time-step during sampling. Therefore, ILVR is able to sample high-quality images while controlling the amount of high-level semantics being inherited from the given reference images.
%
%
As the nature of diffusion models for adopting a progressive denoising process, the generation/synthesis via sampling can start from a noisy input (similar to the intermediate stage of sampling) instead of always beginning from random noise.
SDEdit~\cite{meng2021sdedit} hence realizes stroke-based image synthesis by starting the sampling from a stroke input with noise injected, in which the generative model used in SDEdit is built upon stochastic differential equations where its mechanism is quite similar to diffusion models (e.g. sampling via iterative denoising).
%
%
In this work, we exploit both the techniques of classifier-free diffusion guidance and ILVR into our diffusion-based framework of image generation for fulfilling a three-dimensional control on the synthesized images in terms of their realism and the consistency with respect to the stroke and sketch conditions.

\section{Method}
\vspace{\secmargin}
As motivated above, our proposed framework, named \modelName, aims to perform image generation conditioned on the input of stroke and sketches with three-dimensional control over the faithfulness to the conditions and the realism of the synthesized output. In the following we sequentially describe our proposed method, starting from the preliminaries for diffusion models (Section~\ref{backgound}) and the modifications we make for realizing the conditional generation and the discussion for the sketch and stroke guidance (enabled by the technique of classifier-free diffusion guidance, Section~\ref{conditioning}), and the control over realism (achieved by the technique of iterative latent variable refinement, Section~\ref{realism}). 
\subsection{Preliminaries}
\vspace{\subsecmargin}
\label{backgound}
Denoising diffusion probabilistic models (DDPM)~\cite{ho2020denoising,nichol2021improved} are a class of generative models (and the diffusion models that our proposed framework is based on) which adopt a denoising process to formulate the mapping from a simple distribution (e.g., isotropic Gaussian) to the target distribution.
The forward diffusion process gradually adds noises to the data sampled from the target distribution, while the backward denoising process attempts to learn the reverse mapping.
Both processes are modeled as Markov chains. 
Here we briefly introduce the process following the formulations and notations in ~\cite{nichol2021improved}.

\comment{
Diffusion models are a class of generative models which performs a data distribution by learning a reverse gradual noising process. 
After Gaussian diffusion models firstly introduced by ~\cite{sohl2015deep}, ~\cite{ho2020denoising} adapts it and officially proposes Denoising diffusion probabilistic models (DDPM). 
Denoising diffusion probabilistic models (DDPM) aims to model a Markov transition process which progressively converts a simple distribution(e.g., isotropic Gaussian) into a target data distribution. Here we briefly go through the approach and follow the formulations and notations in ~\cite{nichol2021improved}, an improved DDPM with a few modifications, which is also the base model of our proposed method.}

Given a sample from the target data distribution $x_0 \sim q(x_0)$, the forward diffusion path of DDPM is a Markov chain produced by gradually adding Gaussian noise to $x_0$ with total $T$ steps:
\vspace{\eqmargin}
\begin{equation}
q(x_t \vert x_{t-1}) \vcentcolon= \mathcal{N}(x_t; \sqrt{1-\beta_t}x_{t-1}, \beta_t\textbf{I}),
\vspace{\eqmargin}
\end{equation}
where $t \sim [1, T]$ and $\beta_1$,...,$\beta_T$ is a fixed variance schedule with $\beta_t\in(0, 1)$.
Sampling $x_t$ at an arbitrary timestep $t$ can be expressed in a closed form:
\vspace{\eqmargin}
\begin{equation}
\begin{aligned}
\label{eqn:xt}
&q(x_t \vert x_0) \vcentcolon= \mathcal{N}(x_t; \sqrt{\bar{\alpha}_t}x_0, (1-\bar{\alpha}_t)\textbf{I}), \\
&x_t = \sqrt{\bar{\alpha}_t}x_0 + \sqrt{(1-\bar{\alpha}_t)}\epsilon,
\end{aligned}
\vspace{\eqmargin}
\end{equation}
where $\alpha_t \vcentcolon= 1-\beta_t$ and $\bar{\alpha}_t \vcentcolon= \prod^t_{i=1} \alpha_i$.
Consequently, $x_t$ can be viewed as a linear combination of the original data $x_0$ and $\epsilon \sim \mathcal{N}(0, \textbf{I})$.
The true posterior $q(x_{t-1} \vert x_t)$ can be well approximated by a diagonal Gaussian when the magnitude of noise $\beta_t$ added at each step is small enough. 
Moreover, $x_T$ is nearly an isotropic Gaussian $\mathcal{N}(0, \textbf{I})$ when $T$ is large enough. 
These behaviors facilitate a generative (denoising) process learning, the reverse of the forward path, to approximate the true posterior $q(x_{t-1} \vert x_t)$.
Specifically, DDPM adopts a deep neural network (typically U-Net is adopted) to predict the mean and the covariance of $x_{t-1}$ given $x_t$ as input and the generative process is expressed as parameterized Gaussian transitions:
\vspace{\eqmargin}
\begin{equation}
p_\theta(x_{t-1} \vert x_t) \vcentcolon= \mathcal{N}(x_{t-1}; \mu_\theta(x_t, t), \Sigma_\theta(x_t, t)). 
\vspace{\eqmargin}
\end{equation}
Ho \etal~\cite{ho2020denoising} propose to predict the noise $\epsilon_\theta(x_t, t)$ instead and derives $\mu_\theta(x_t, t)$ using Bayes's theorem:
\vspace{\eqmargin}
\begin{equation}
\mu_\theta(x_t, t) = \frac{1}{\sqrt{\alpha_t}}(x_t - \frac{\beta_t}{\sqrt{1-\bar{\alpha}_t}}\epsilon_\theta(x_t, t)).
\vspace{\eqmargin}
\end{equation}
To perform the learning of the denoising process, we first generate sample $x_t \sim q(x_t \vert x_0)$ by adding Gaussian noise $\epsilon$ to $x_0$(i.e. Eq.~\ref{eqn:xt}), then train a model $\epsilon_\theta(x_t, t)$ to predict the added noise using a standard MSE loss:
\vspace{\eqmargin}
\begin{equation}
L_\text{simple} \vcentcolon= E_{t \sim [1, T], x_0 \sim q(x_0), \epsilon \sim \mathcal{N}(0, \textbf{I})}[{\|\epsilon - \epsilon_\theta(x_t, t)\|}^2] .
\vspace{\eqmargin}
\end{equation}
For $\Sigma_\theta(x_t, t)$, Nichol \etal~\cite{nichol2021improved} presents an effective learning strategy as an improved version of DDPM with fewer steps needed and applies an additional loss term $L_{\text{vlb}}$ (the details are shown in ~\cite{nichol2021improved})
that interpolates between the upper and lower bounds for the fixed covariance proposed by the original DDPM.
The overall hybrid objective that we adopt is:
\vspace{\eqmargin}
\begin{equation}
L_\text{hybrid} \vcentcolon= L_\text{simple} + L_\text{vlb}.
\vspace{\eqmargin}
\end{equation}
%
\subsection{Sketch- and Stroke-Guided Diffusion Model}
\vspace{\subsecmargin}
\label{conditioning}
 To generate images based on the given sketches and strokes, our proposed method concatenates the sketch condition $\textbf{c}_\text{sketch}$ and the stroke condition $\textbf{c}_\text{stroke}$ along with $x_t$ as input for the U-Net model (which is responsible for posterior prediction). 
The modified parameterized Gaussian transition for conditioning generation is then represented as:
\vspace{\eqmargin}
\begin{equation} \label{eq-clf}
\begin{aligned}
 &\hat{p}_\theta(\tilde{x}_{t-1} \vert x_t, \textbf{c}_\text{sketch}, \textbf{c}_\text{stroke}) \\
 &\vcentcolon= \mathcal{N}(\tilde{x}_{t-1}; \mu_\theta(x_t, t, \textbf{c}_\text{sketch}, \textbf{c}_\text{stroke}),  \Sigma_\theta(x_t, t, \textbf{c}_\text{sketch}, \textbf{c}_\text{stroke})). 
\end{aligned}
\vspace{\eqmargin}
\end{equation} 
In practice, as following~\cite{ho2020denoising}, the conditioning denoising process learns the noise prediction with additional sketch and stroke information, denoted as $\hat{\epsilon}_\theta(x_t, t, \textbf{c}_\text{sketch}, \textbf{c}_\text{stroke})$:
\vspace{\eqmargin}
\begin{equation}
\hat{L}_\text{simple} \vcentcolon= E_{t, x_0, \epsilon}[{\|\epsilon - \hat{\epsilon}_\theta(x_t, t, \textbf{c}_\text{sketch}, \textbf{c}_\text{stroke})\|}^2].
\vspace{\eqmargin}
\end{equation}

\begin{figure*}[t]
    \centering
    \setlength\tabcolsep{1.5pt} 
    \begin{tabular}{cccc:ccccc}
    Sketch & Stroke & Combine &  &  & Ours & SDEdit~\cite{meng2021sdedit} & SSS2IS~\cite{liu2021self} & U-GAT-IT~\cite{kim2019u} \\
    \includegraphics[height=.26\linewidth]{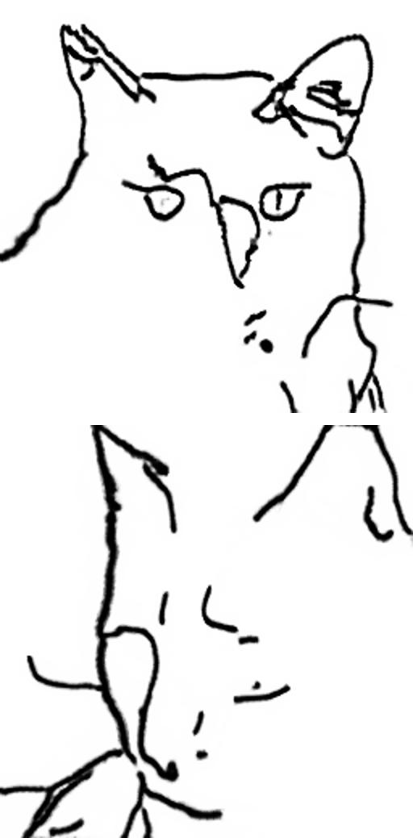} &
    \includegraphics[height=.26\linewidth]{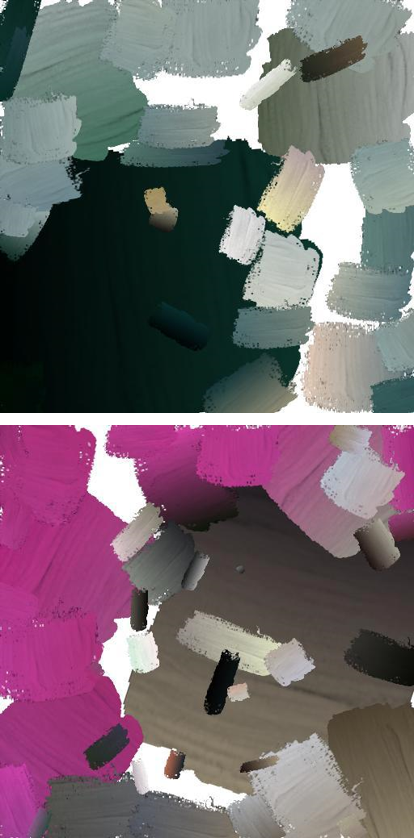} &
    \includegraphics[height=.26\linewidth]{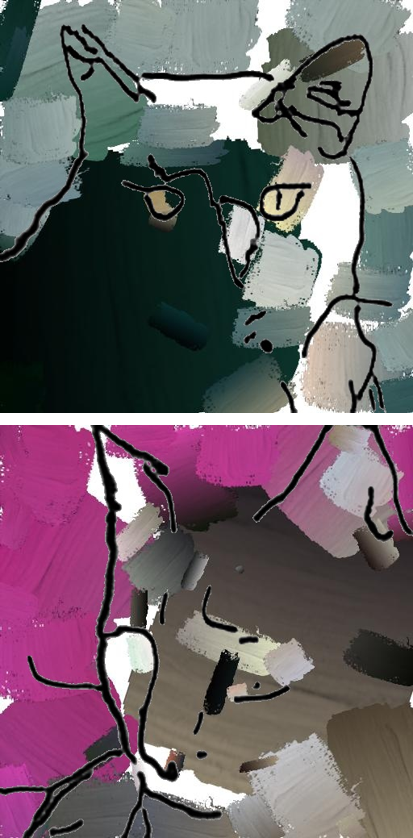} & & &
    \includegraphics[height=.26\linewidth]{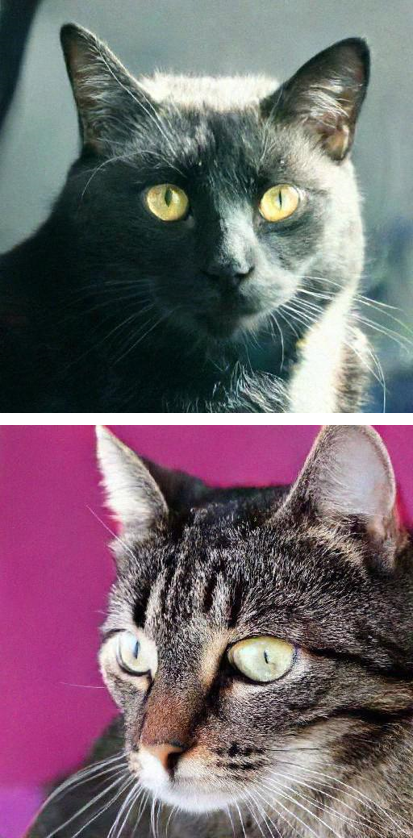} &
    \includegraphics[height=.26\linewidth]{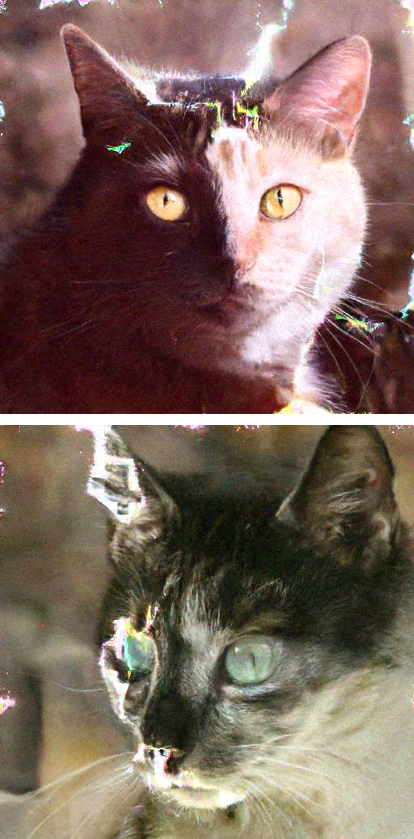} &
    \includegraphics[height=.26\linewidth]{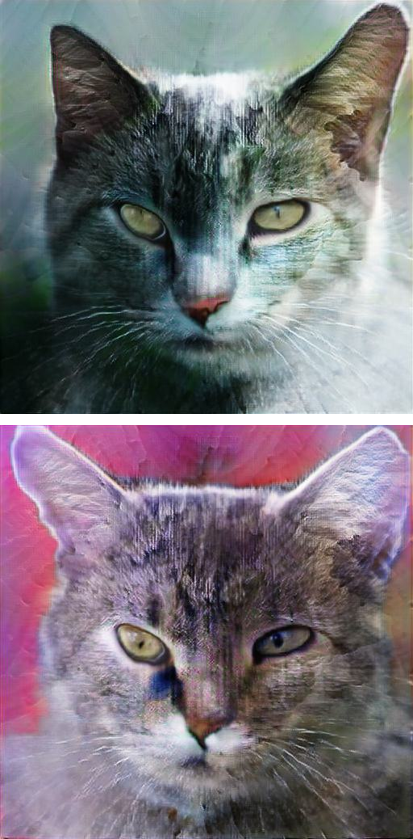} &
    \includegraphics[height=.26\linewidth]{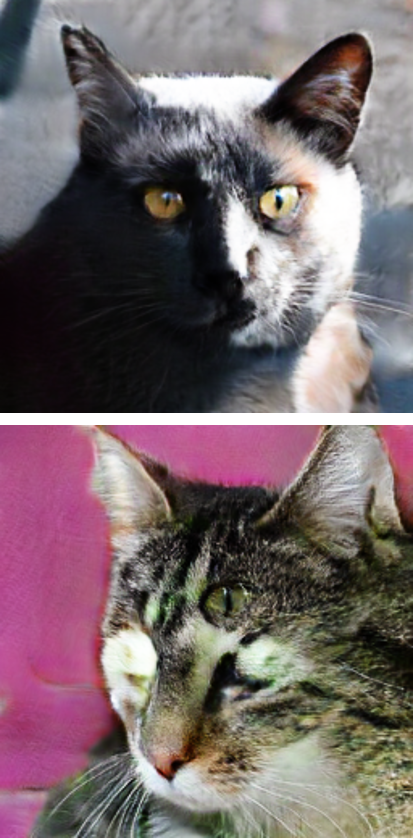} \\
    \includegraphics[height=.26\linewidth]{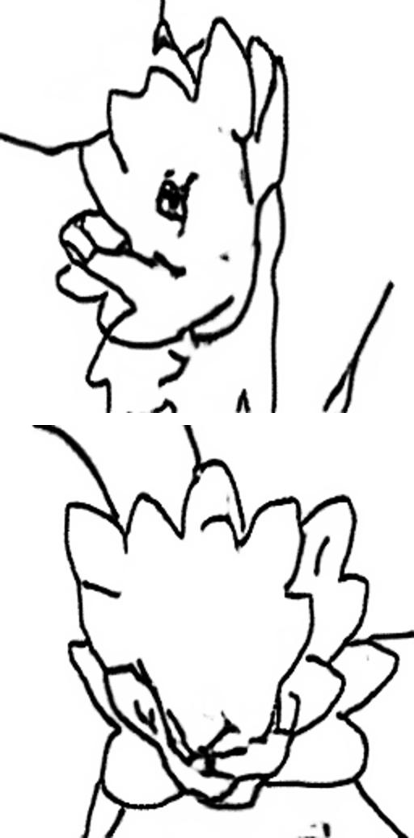} &
    \includegraphics[height=.26\linewidth]{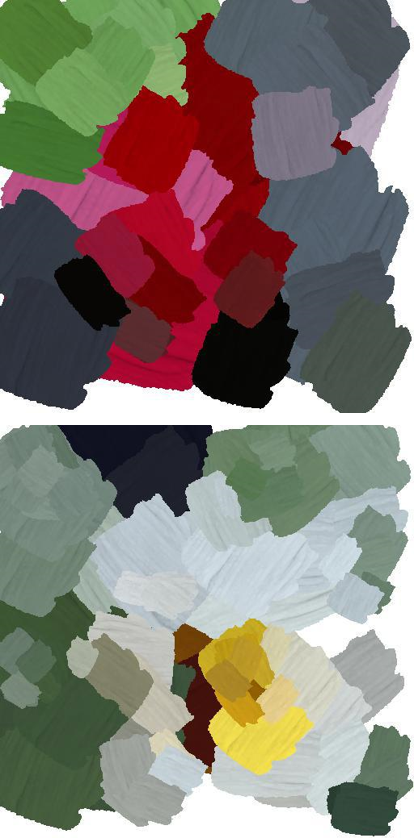} &
    \includegraphics[height=.26\linewidth]{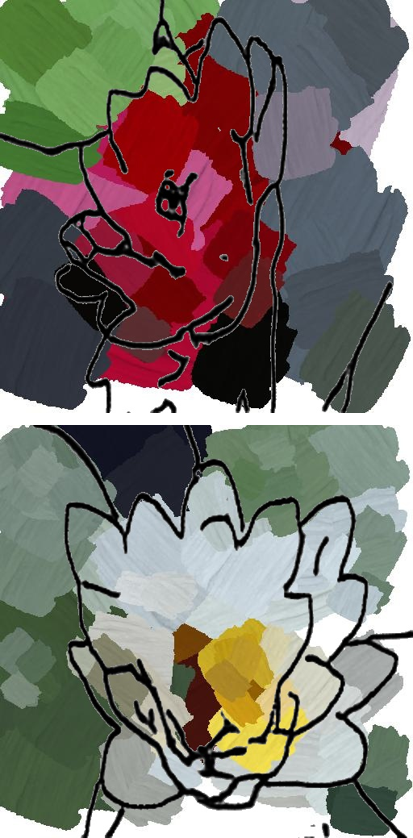} & & &
    \includegraphics[height=.26\linewidth]{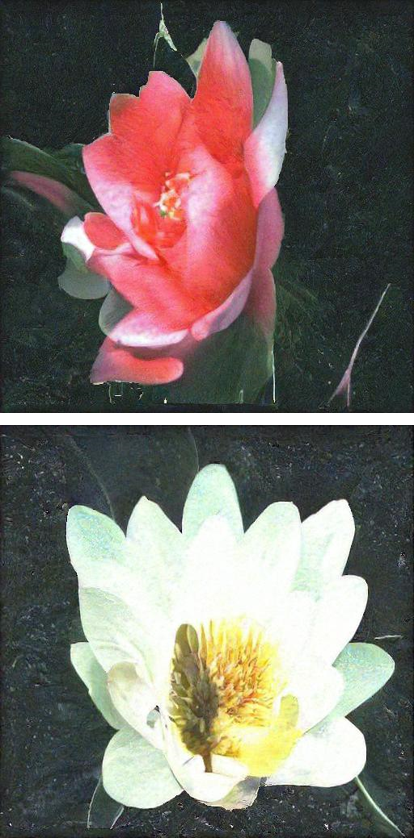} &
    \includegraphics[height=.26\linewidth]{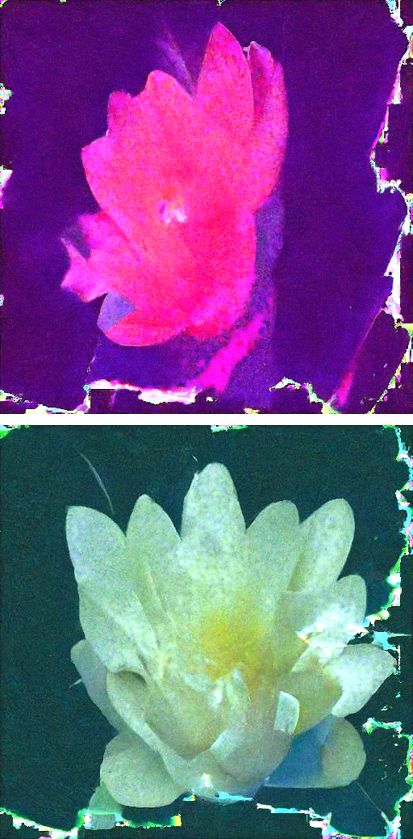} &
    \includegraphics[height=.26\linewidth]{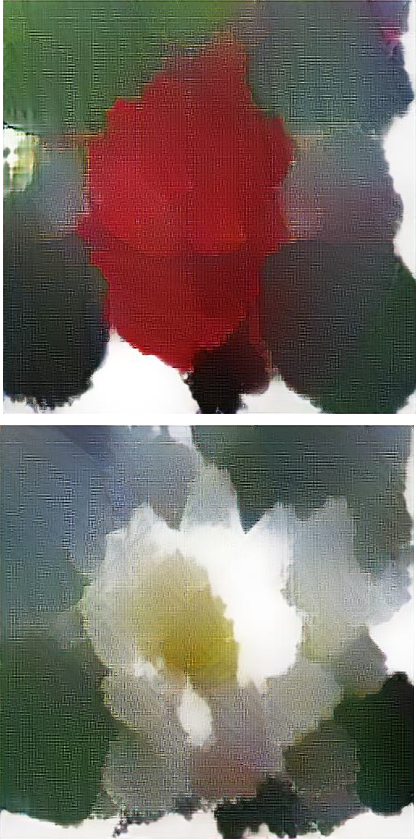} &
    \includegraphics[height=.26\linewidth]{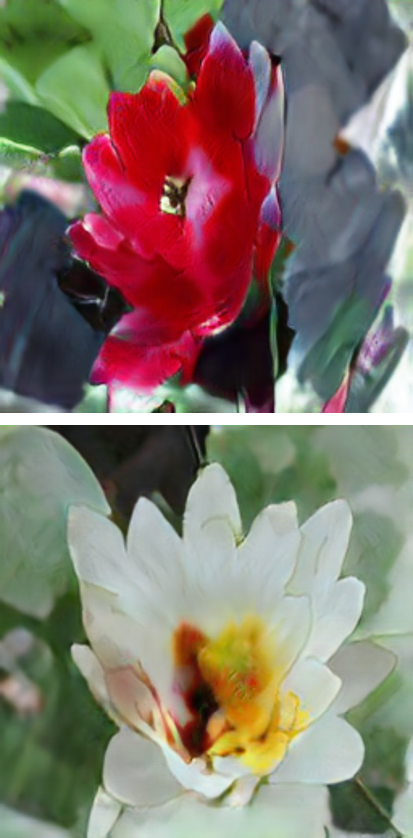} \\
    \includegraphics[height=.26\linewidth]{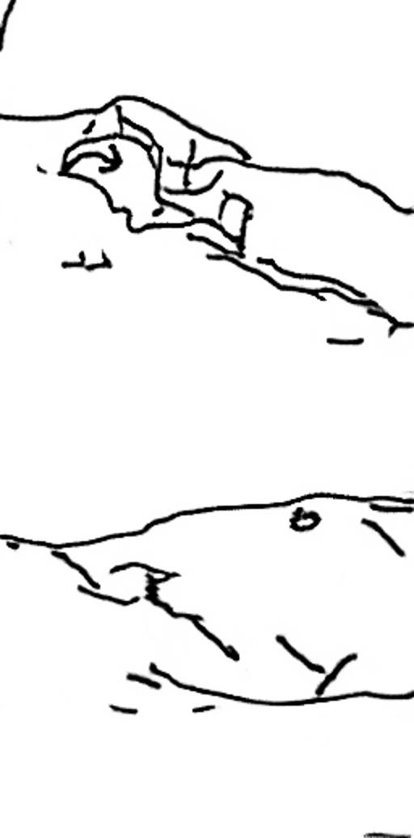} &
    \includegraphics[height=.26\linewidth]{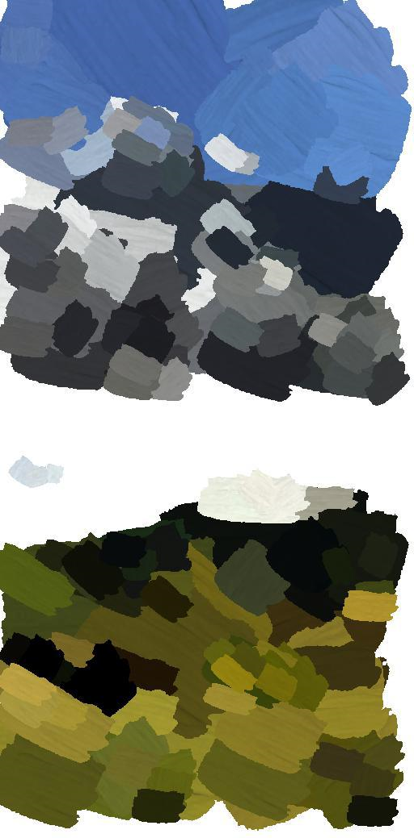} &
    \includegraphics[height=.26\linewidth]{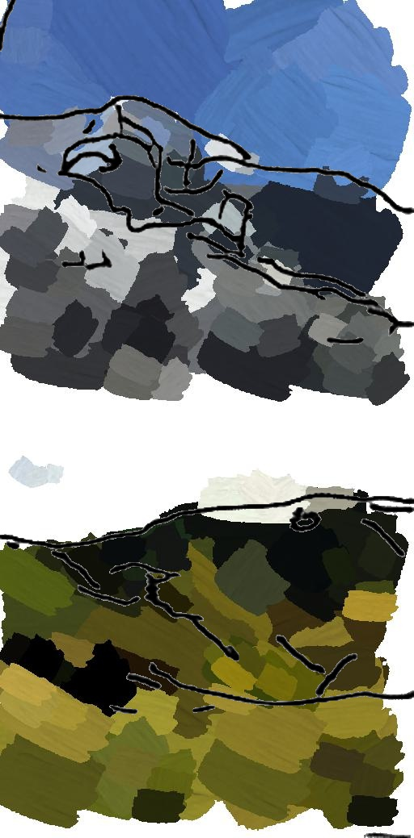} & & &
    \includegraphics[height=.26\linewidth]{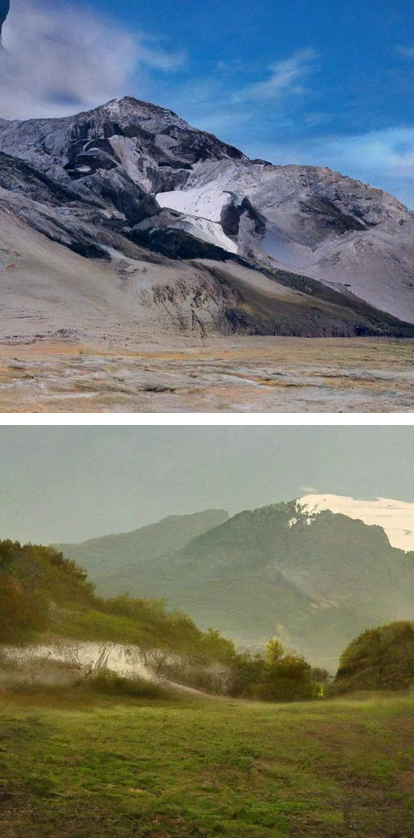} &
    \includegraphics[height=.26\linewidth]{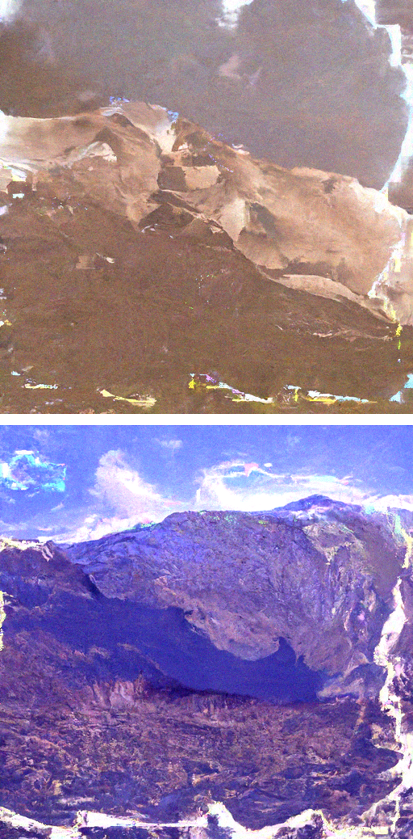} &
    \includegraphics[height=.26\linewidth]{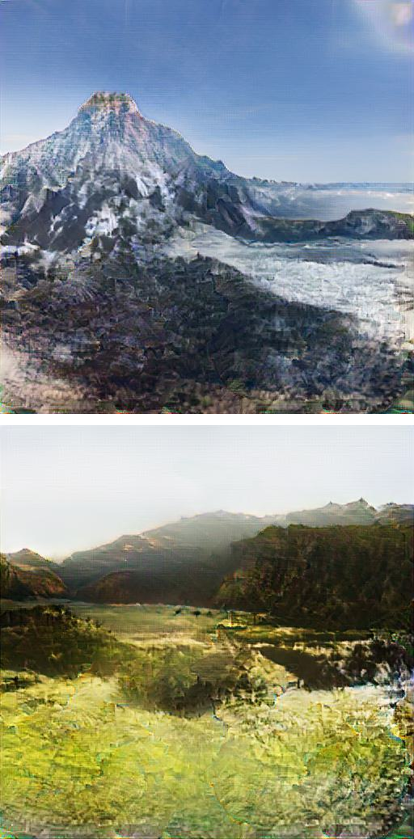} &
    \includegraphics[height=.26\linewidth]{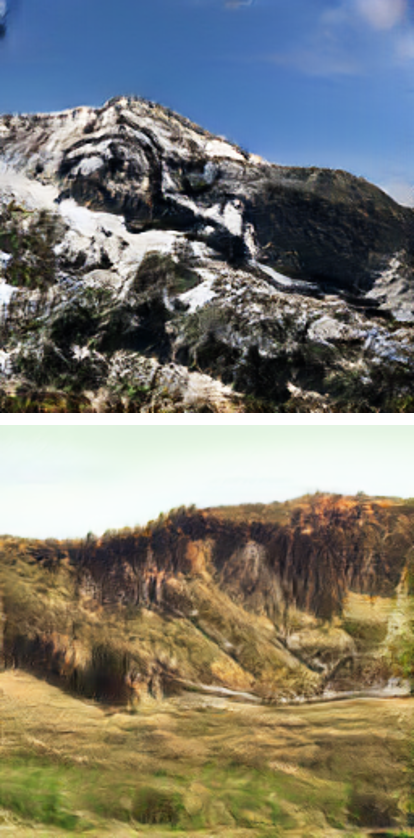} \\

    \end{tabular}
    \vspace{-2mm}
\caption{
\textbf{Qualitative comparisons.}
We present results from different approaches on the (\textit{top two rows})~AFHQ, (\textit{middle two rows})~Oxford Flower, and (\textit{bottom two rows})~Landscapes datasets.
U-GAT-IT~\cite{kim2019u}, as an image-to-image translation method, takes as input the combination of sketches and strokes (third column).
SDEdit~\cite{meng2021sdedit}, SSS2IS~\cite{liu2021self} and our model take the contour and color as separate inputs (the leftmost two columns).
}
\label{fig:visual-qual}
\vspace{\figmargin}
\end{figure*}
To separately control the guidance level of the sketch and stroke conditions, we leverage classifier-free guidance~\cite{ho2021classifier} and modify it for two-dimensional guidance.  
In practice, we adopt a two-stage training strategy.
First, we train the model with complete sketches and strokes as conditions. Then we fine-tune the model by randomly replacing 30\% of each condition with an image filled with gray pixels, denoted as $\boldsymbol{\emptyset}$, for unconditional representation.
During sampling, the ratio between the degree of faithfulness to strokes and sketches is controlled through the following linear combination with two guidance scales $s_\text{sketch}$ and $s_\text{stroke}$:
\vspace{\eqmargin}
\begin{equation} \label{eq-trade-off}
\begin{split}
&\hat{\epsilon}_\theta(x_t, t, \textbf{c}_\text{sketch}, \textbf{c}_\text{stroke}) 
= \hat{\epsilon}_\theta(x_t, t,  \boldsymbol{\emptyset}, \boldsymbol{\emptyset}) \\
&+ s_\text{sketch}(\hat{\epsilon}_\theta(x_t, t,  \textbf{c}_\text{sketch}, \boldsymbol{\emptyset}) - \hat{\epsilon}_\theta(x_t, t,  \boldsymbol{\emptyset}, \boldsymbol{\emptyset})) \\
&+ s_\text{stroke}(\hat{\epsilon}_\theta(x_t, t,  \boldsymbol{\emptyset}, \textbf{c}_\text{stroke}) - \hat{\epsilon}_\theta(x_t, t,  \boldsymbol{\emptyset}, \boldsymbol{\emptyset})).
\end{split}
\vspace{\eqmargin}
\end{equation}
With this formulation, our model supports multi-guidance on a single diffusion model.

\subsection{Realism Control}
\vspace{\subsecmargin}
\label{realism}
In reality, sketches and strokes provided by users are usually inconsistent to the real images. 
Therefore, it is essential to provide the control over how faithful the output should be to the inputs. In other words, how realistic the output should be.
We then provide realism control in addition to the two-dimensional classifier-free guidance with sketch and stroke information.
We apply iterative latent variable refinement~\cite{choi2021ilvr} to refine each intermediate transition in the generative process with a downsampled reference image. 
%
The proposed realism control allows additional trade-off  between consistency to the provided strokes/sketches and the distance to target data distribution (i.e. real images).
Let $LP$ represents a linear low pass filtering operation which performs downsampling to a transformed size \textbf{N} and upsampling back.
Given a realism scale $s_\text{realism} \sim [0, 1]$ as an indication of the transformed size \textbf{N} and a reference image combining sketch and stroke information $\textbf{c}_\text{comb}$ of size $\textbf{m}*\textbf{m}$, the realism adjustment during the conditioning generative process at timestep $t$ can be expressed as:
\vspace{\eqmargin}
\begin{equation} \label{eqn:ilvr}
\begin{split}
&\tilde{x}_{t-1} \sim \hat{p}_\theta(\tilde{x}_{t-1} \vert x_t, \textbf{c}_\text{sketch}, \textbf{c}_\text{stroke}), \\
& x_{t-1} \vcentcolon= \tilde{x}_{t-1} - {LP}_\textbf{N}(\tilde{x}_{t-1}) + {LP}_\text{N}( {\textbf{c}_\text{comb}}_{t-1}), \\
& \text{in which } \textbf{N} = -s_\text{realism}(\nicefrac{\textbf{m}}{8}-1) + (\nicefrac{\textbf{m}}{8})+k
\end{split}
\vspace{\eqmargin}
\end{equation}
where ${\textbf{c}_\text{comb}}_{t-1} \sim q({\textbf{c}_\text{comb}}_{t-1} \vert \textbf{c}_{\text{comb}_0})$ with $\textbf{c}_{\text{comb}_0} = \textbf{c}_\text{comb}$ showing that ${\textbf{c}_\text{comb}}_{t-1}$ is sampled following Eq.
~\ref{eqn:xt} as gradually injecting noise into $\textbf{c}_\text{comb}$ by $t-1$ steps. In details, as $x_{t-1}$ can be seen as combining the high-frequency contents of $\tilde{x}_{t-1}$ (produced by $\tilde{x}_{t-1} - {LP}_\textbf{N}(\tilde{x}_{t-1})$) with the low-frequency contents of the corrupted reference ${\textbf{c}_\text{comb}}_{t-1}$, the downsampled size \textbf{N} controlled by $s_\text{realism}$ determines how faithful the output should be to the reference (on the other hand, the tendency of the synthesized output towards the target distribution). The detailed discussion and explanation for the computation of \textbf{N} is provided in the supplementary materials.\par
\vspace{-1mm}
The overall three-dimensional control of our proposed framework is illustrated in Figure~\ref{fig:method}, in which it is realized via the combination of the sketch- and stroke-guidance with the realism control.

\begin{figure*}[t!]
\centering
\setlength\tabcolsep{1.5pt} 
\begin{tabular}{c:c|c:c}
Input & Multimodal Results & Input & Multimodal Results\\
\includegraphics[width=.24\linewidth]{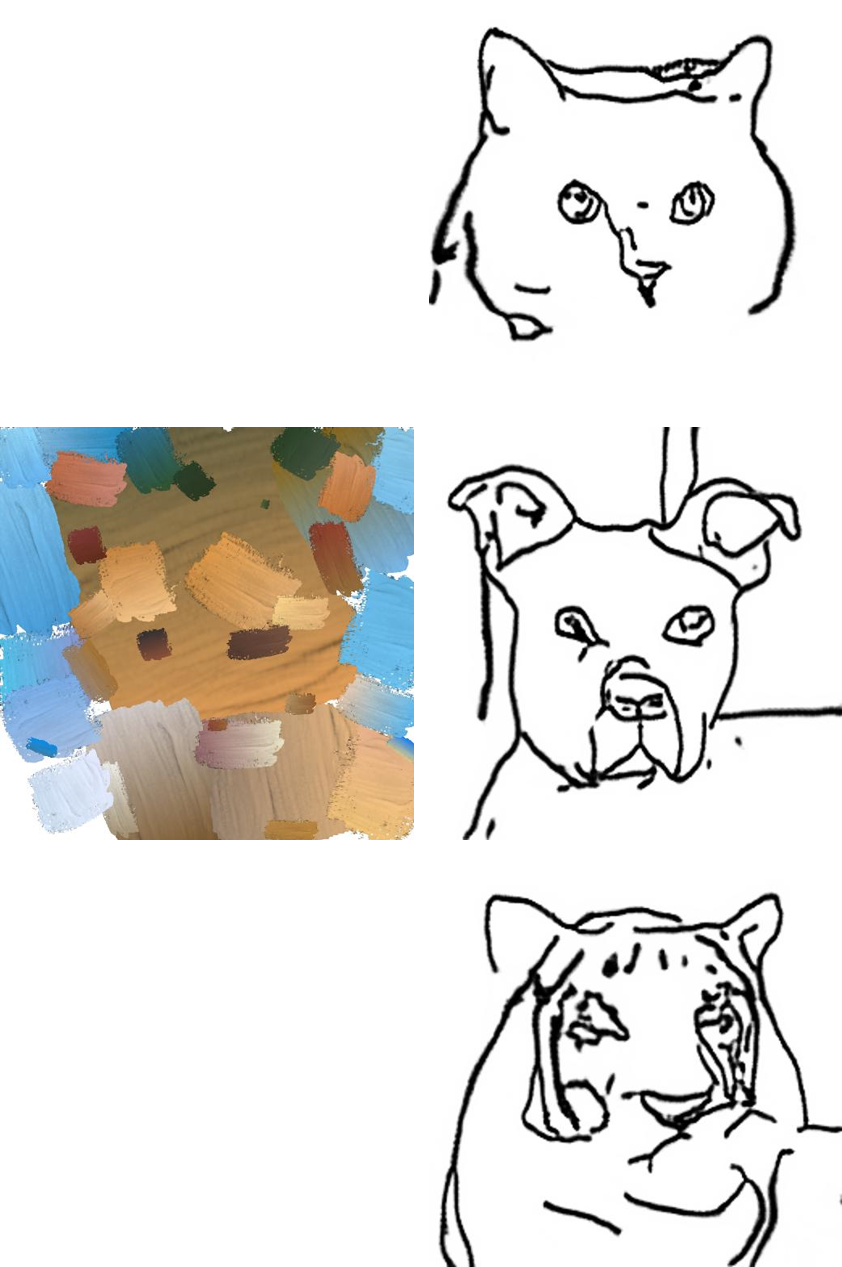} &
\includegraphics[width=.24\linewidth]{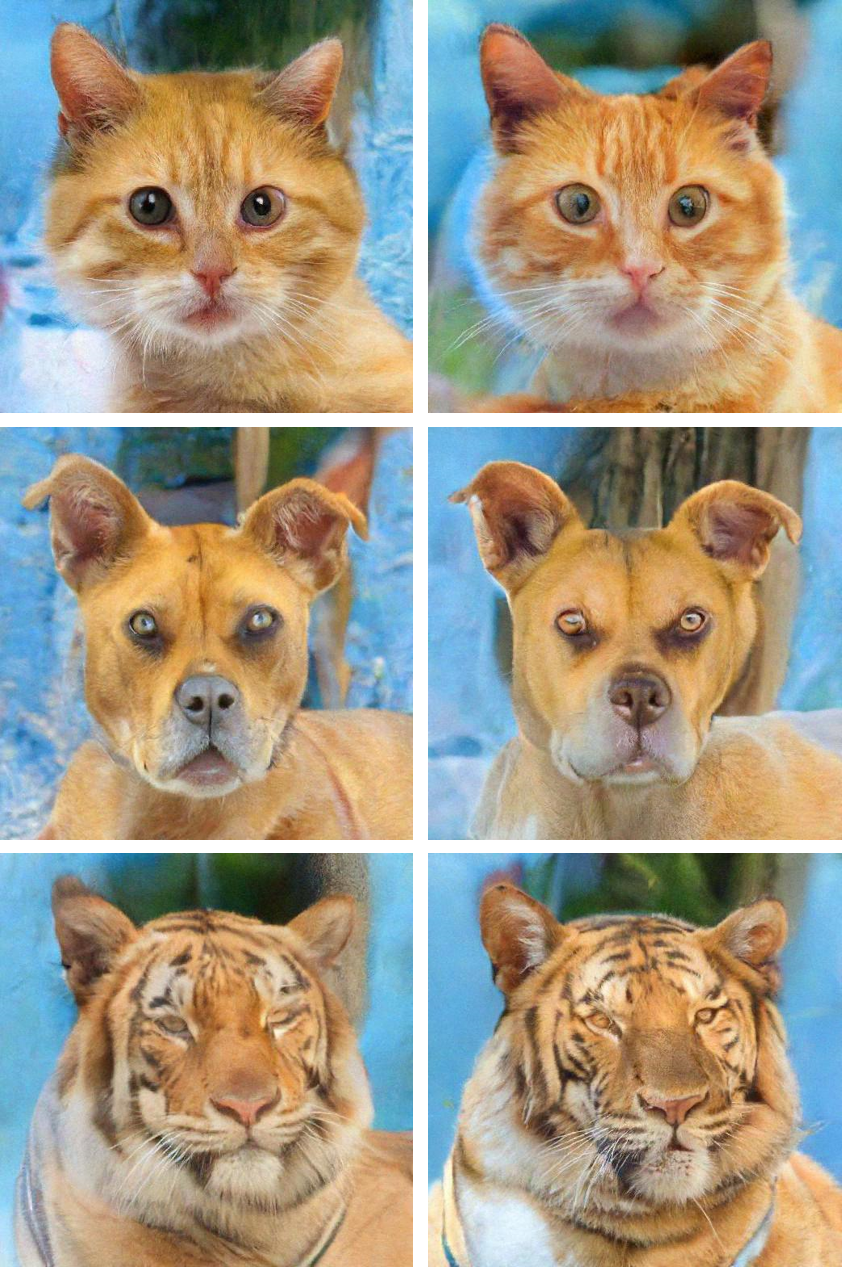} &
\includegraphics[width=.24\linewidth]{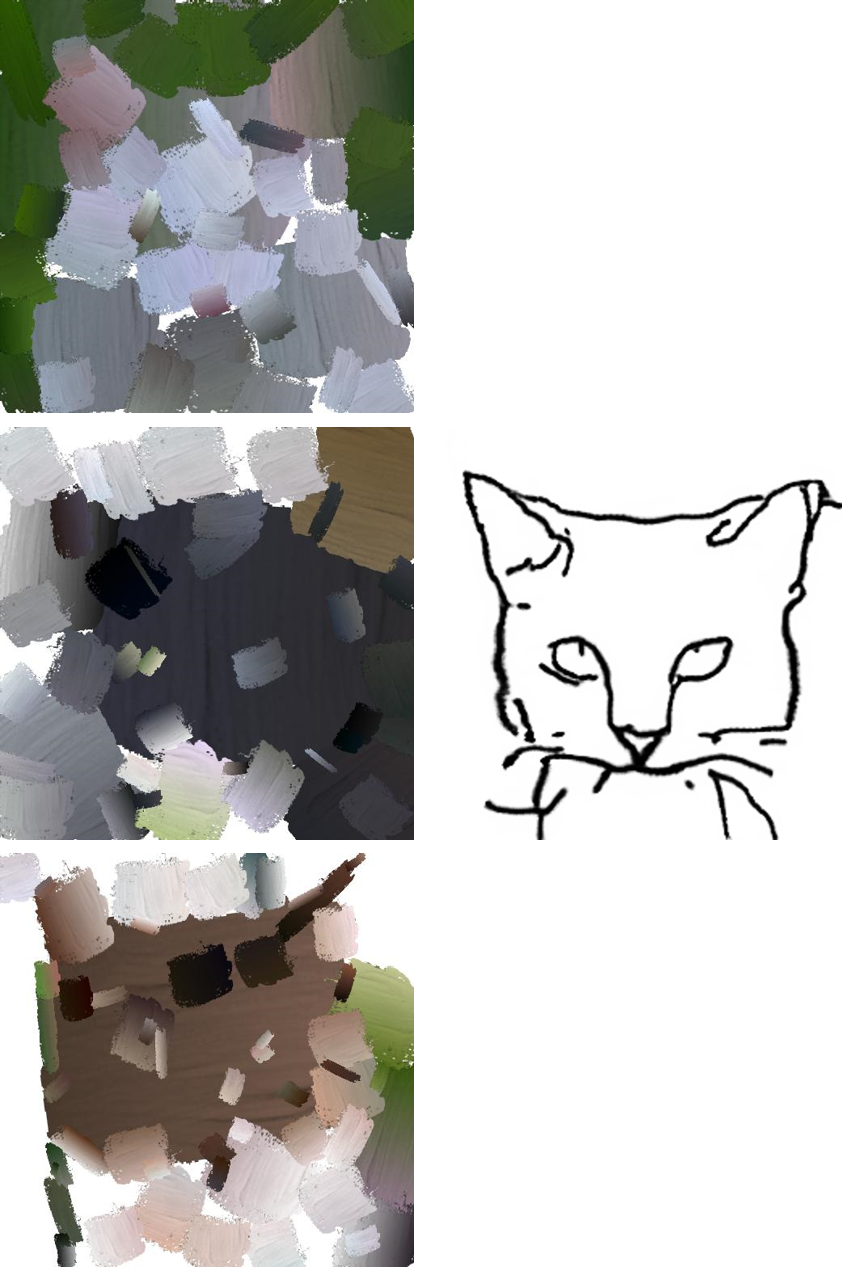} &
\includegraphics[width=.24\linewidth]{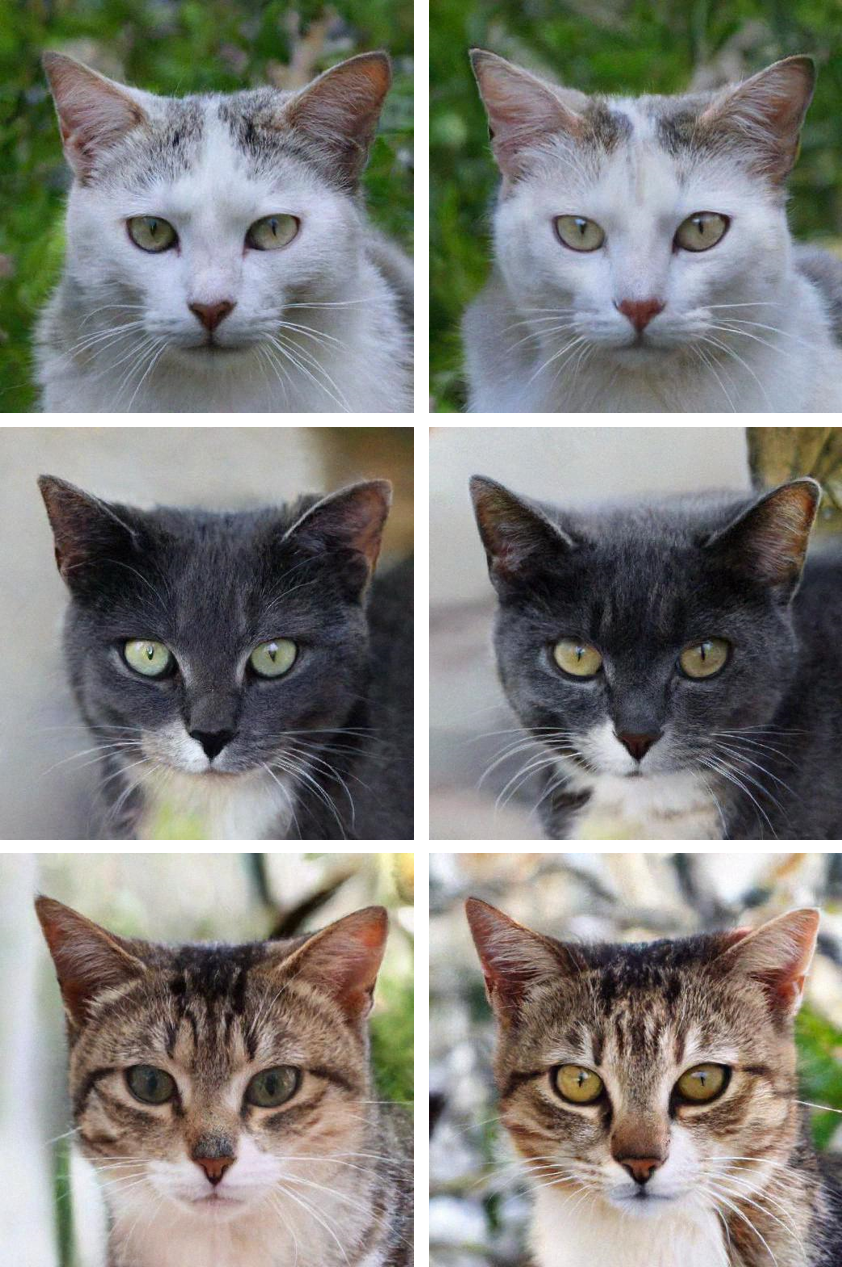} 
\end{tabular}
\vspace{-1em}
\caption{
\textbf{Multi-modal and multi-domain generation.}
The proposed approach 1) produces multi-modal results from the same set of input data, 2) understands the implicit class information from the input sketch image (as shown in the left hand side of the figure), and 3) is robust to \textit{un-aligned} sketch stroke input data (note the input sketch and stroke are extracted from different source images in this example).
}
\label{fig:visual-multi}
\vspace{-1em}
\end{figure*}
\section{Experiments}
\vspace{\secmargin}
We conduct extensive qualitative and quantitative experiments to validate the effectiveness of the proposed \modelName method on the task of image generation from stroke and sketch.
First, we compare our approach with several recent state-of-the-art frameworks, and demonstrate the three-dimensional control (contour, color, realism) over the generation process.
Second, we show two applications: multi-conditioned local editing and region-sensitive stroke-to-image generation.
Finally, we discuss the trade-off and the interaction between the three controllable dimensions.
%
%
%
\vspace{-2em}
\paragraph{Datasets.} We conduct experiments using the AFHQ~\cite{choi2020stargan}, Landscapes~\cite{skorokhodov2021aligning} and Oxford Flower~\cite{nilsback2008automated} datasets.
We use Photo-sketching~\cite{li2019photo} to generate the black sketches, and the stylized neural painting~\cite{zou2021stylized} as well as the paint transformer~\cite{liu2021paint} model to synthesize the colored strokes for all the datasets.
%
%
%
We provide more data preparation details in the supplementary document.
\vspace{-5mm}
\paragraph{Compared methods.}
We compare our method with three recent state-of-the-art frameworks on image generation via the stroke and sketch task:
\begin{compactitem}
\item \textbf{U-GAT-IT}~\cite{kim2019u} is a recent image-to-image translation approach. To leverage U-GAT-IT, we overlay the black sketches and the colored stroke to form the drawing image, which is considered to belong to the source domain. The corresponding photo-realistic image is then treated as the target domain image.
\item \textbf{SSS2IS}~\cite{liu2021self} is a self-supervised GAN-based scheme that takes as input a black sketch and a style image. 
We retrain the model by replacing the input style images with a colored stroke image, and computing the regression loss between the real image and the autoencoder output.
\item \textbf{SDEdit}~\cite{meng2021sdedit} is a diffusion-based algorithm for the stroke-to-image generation task. 
To involve the sketch guidance, we retrain the model to take the sketch as the conditional signal by concatenating the sketch image with the original input of the U-Net network.
\end{compactitem}
%
%
%
%
%
%

\begin{figure}[t!]
	\centering
	\subfloat[Multi-conditioned local editing.]{%
    	\centering
        \setlength\tabcolsep{1.5pt} 
        \begin{tabular}{c:c|cc}
        Original & Input & Multimodal Results \\
        \includegraphics[height=.71\linewidth]{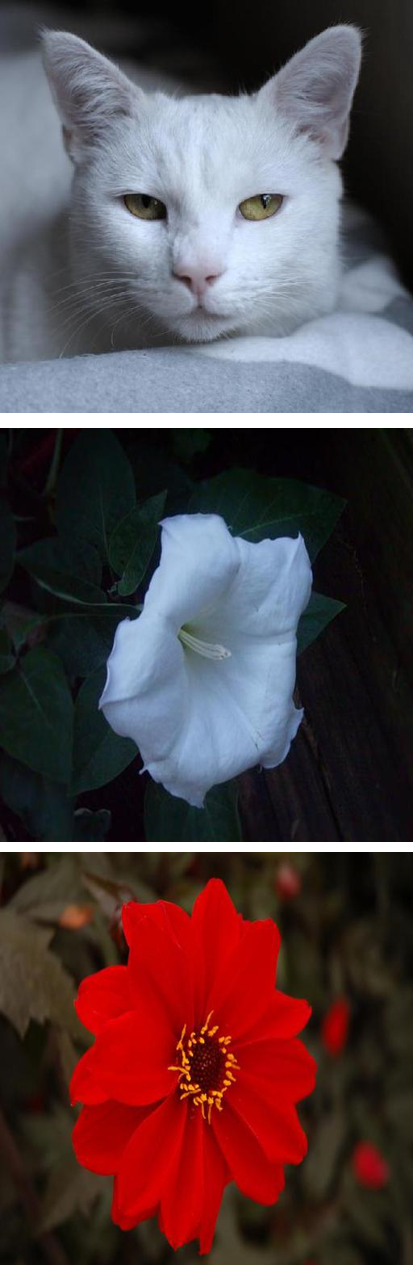} &
        \includegraphics[height=.71\linewidth]{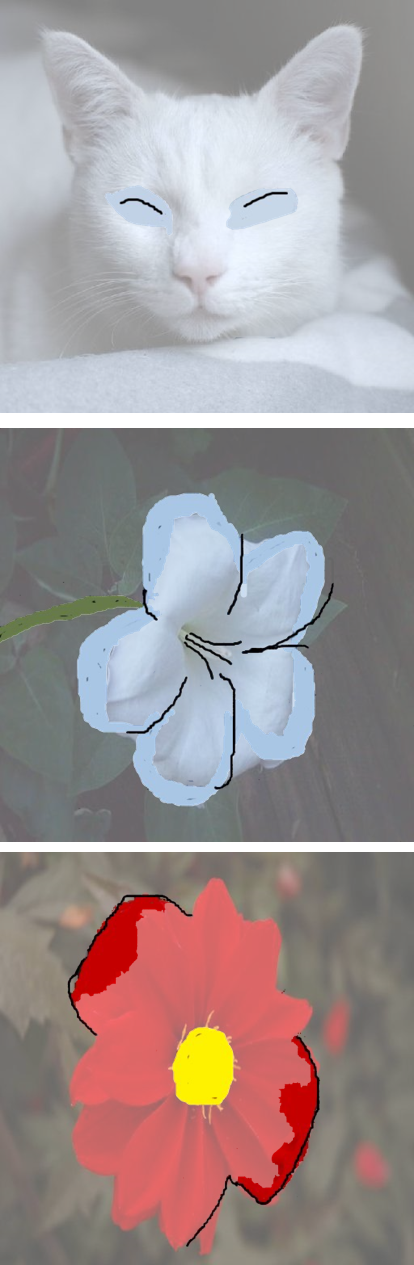} &
        \includegraphics[height=.71\linewidth]{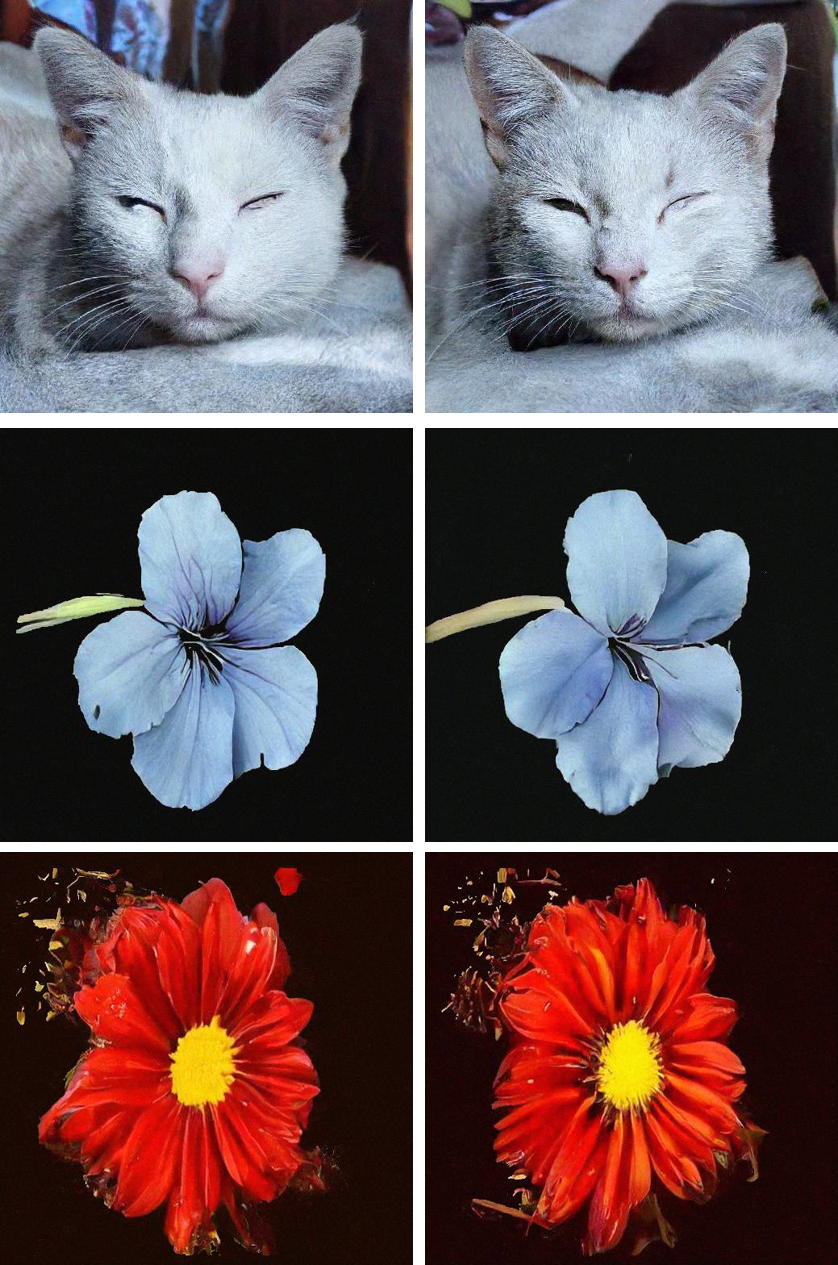} &
        \end{tabular}
	}
    \qquad
    \vspace{\figmargin}
    \subfloat[Region-sensitive stroke-to-image.]{%
        \centering
        \setlength\tabcolsep{1.5pt} 
        \begin{tabular}{c|c}
        Input & Multimodal Results \\
        \includegraphics[height=.71\linewidth]{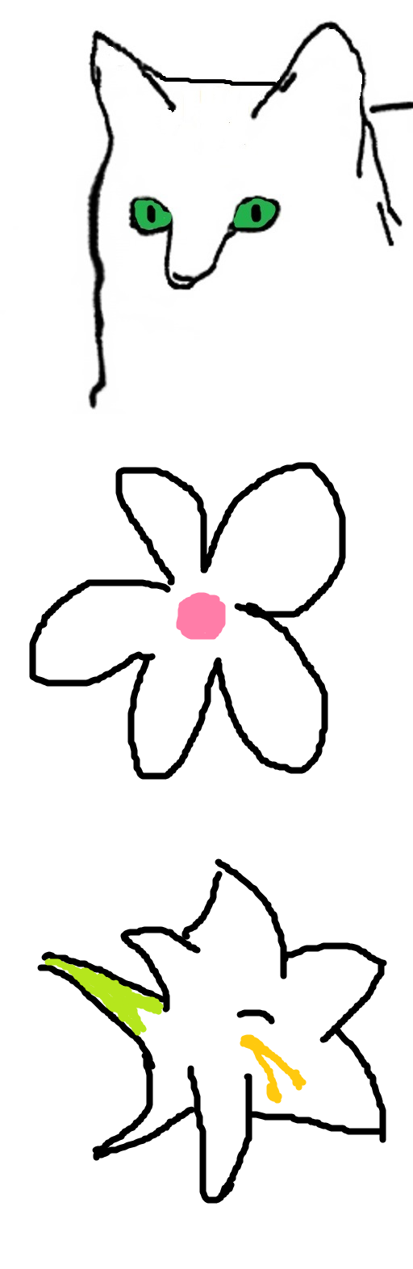} &
        \includegraphics[height=.71\linewidth]{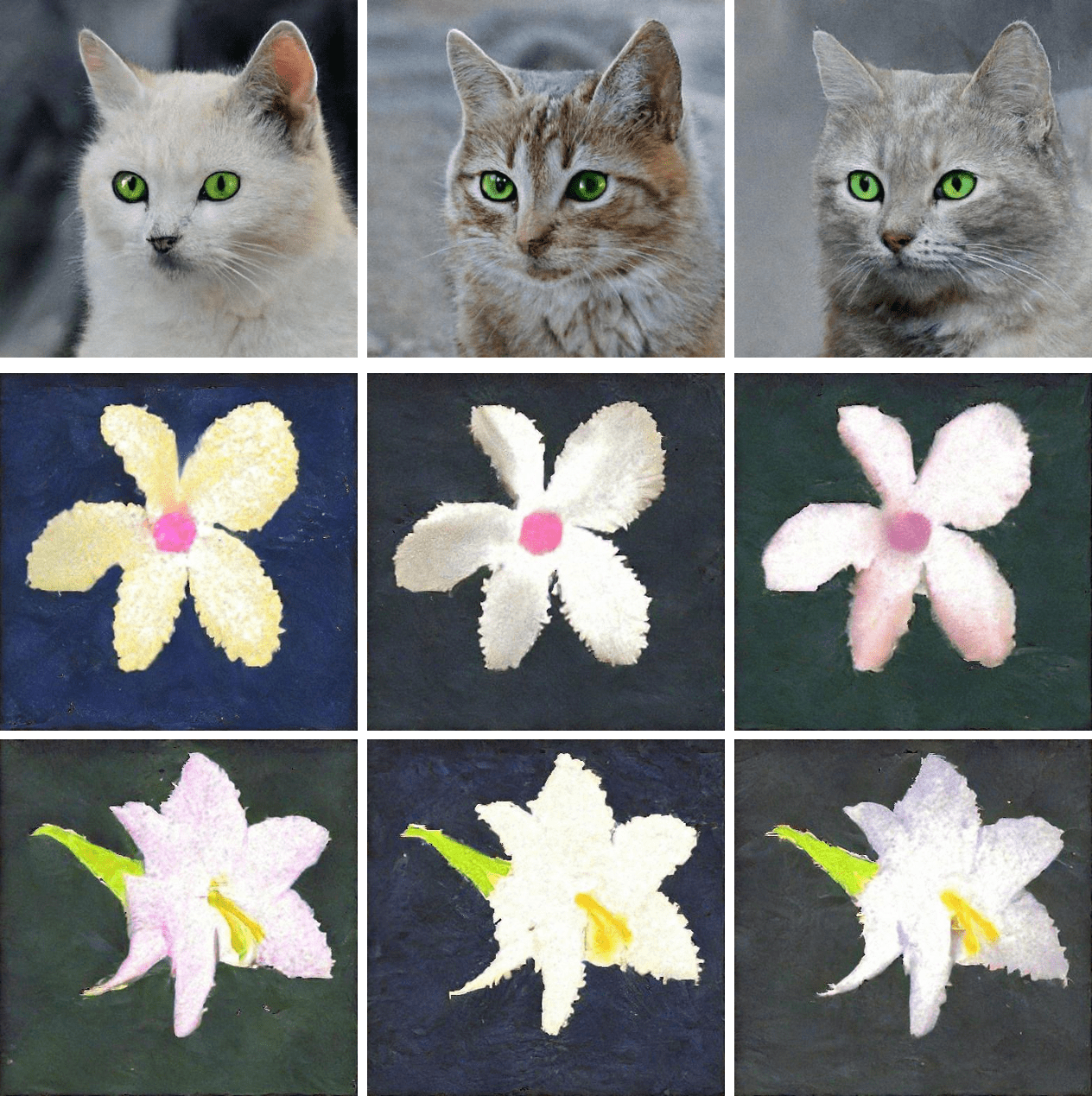} 
        \end{tabular}
	}
	\vspace{-1mm}
    \caption{
    \textbf{Applications.} (a) By drawing the new contour or color on an existing image, the proposed model enables the mask-free image editing. 
    (b) With the partial colored stoke as the input, the proposed method synthesizes more diverse contents in the non-colored region.
    Here we use a cat contour\protect\footnotemark and hand-drawing flowers as examples.
    }
    \label{fig:visual-app}
    \vspace{-1em}
\end{figure}



\footnotetext{Source from https://thenounproject.com/icon/cat-975212/}
\begin{table}[t!]
\centering
\small
\tabcolsep=0.15cm
\caption{
\textbf{Quantitative comparisons.}
We use the FID ($\downarrow$) metric to measure the generated image quality, and the LPIPS ($\downarrow$) score  to evaluate the consistency between the synthesized images and the input sketches.
}
\vspace{-3mm}
\label{tab:fid}
\begin{tabular}{lcccccc} 
\toprule
 & \multicolumn{2}{c}{\textbf{AFHQ-cat}} & \multicolumn{2}{c}{\textbf{Flowers}} & \multicolumn{2}{c}{\textbf{LHQ}} \\ 
\cmidrule(l){2-3}\cmidrule(l){4-5}\cmidrule(l){6-7}
 & FID & LPIPS & FID & LPIPS & FID & LPIPS \\ 
\cmidrule{1-7}
U-GAT-IT & 24.75 & 0.185 & \textbf{74.27} & 0.207 & \textbf{36.93} & 0.188 \\
SSS2IS & 85.48 & 0.23 & 275.24 & 0.227 & 62.25 & 0.143 \\
SDEdit & 30.55 & 0.178 & 138.97 & 0.196 & 84.67 & 0.15 \\ 
\midrule
\textbf{Ours} & \textbf{15.27} & \textbf{0.148} & 83.12 & \textbf{0.125} & 38.83 & \textbf{0.117} \\
\bottomrule
\end{tabular}
\vspace{-1em}
\end{table}


\begin{figure}[t]
    \centering
    \includegraphics[width=\linewidth]{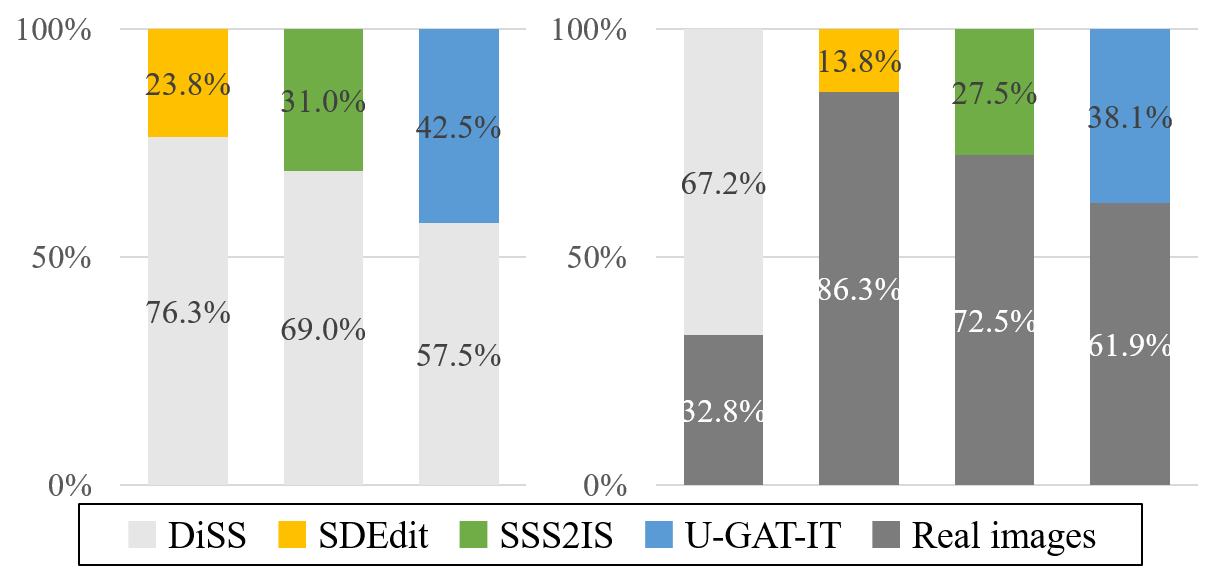}
    \vspace{-2em}
    \caption{
    \textbf{User preference study.} We conduct the study that asks participants to select results (based on images generated from AFHQ-cat and Landscapes datasets) that are \textit{more realistic}. 
    The number indicates the percentage of preference for that particular pair-wise comparison. 
    %
    }
    \label{fig:user-study}
    \vspace{\figmargin}
    \vspace{-2mm}
\end{figure}
\begin{figure*}[t]
    \centering
    \setlength\tabcolsep{1.5pt} 
    \begin{tabular}{c:cccccc}
    Input & Realism = 1.0 & Realism = 0.8 & Realism = 0.6 & Realism = 0.4 & Realism = 0.2 & Realism = 0.0 \\
    \includegraphics[height=.135\linewidth,width=.135\linewidth]{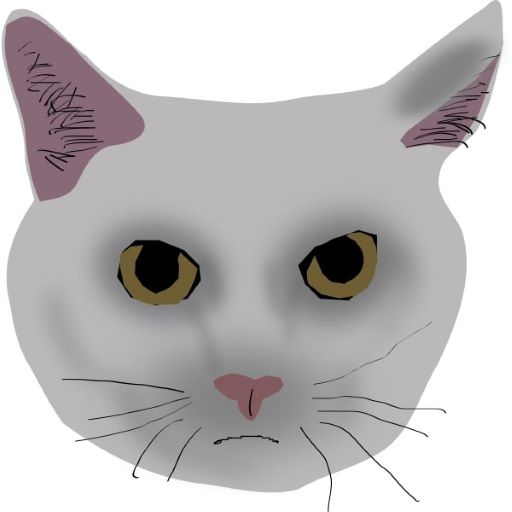} &
    \includegraphics[height=.135\linewidth,width=.135\linewidth]{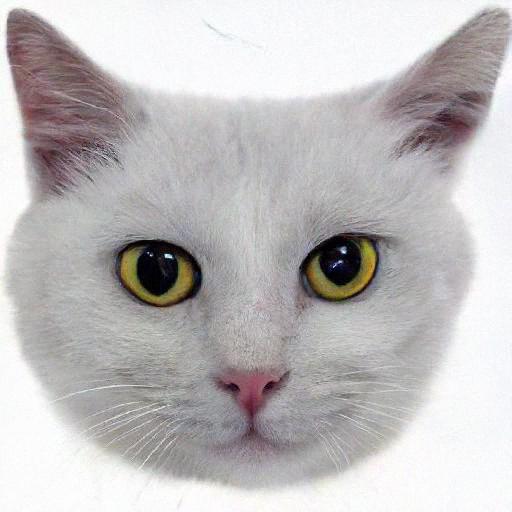} &
    \includegraphics[height=.135\linewidth,width=.135\linewidth]{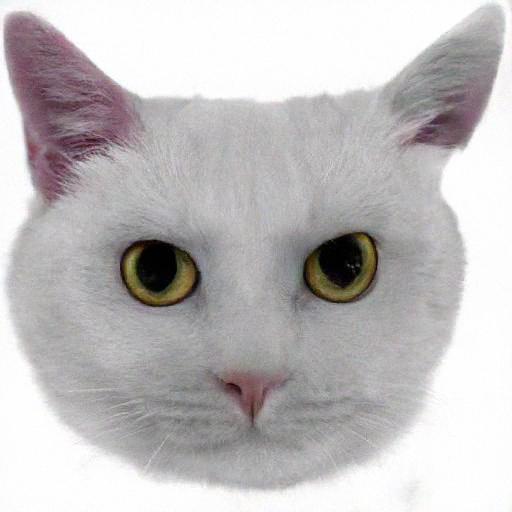} &
    \includegraphics[height=.135\linewidth,width=.135\linewidth]{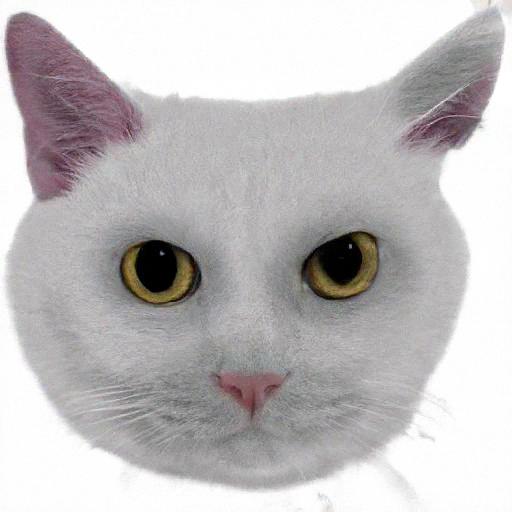} &
    \includegraphics[height=.135\linewidth,width=.135\linewidth]{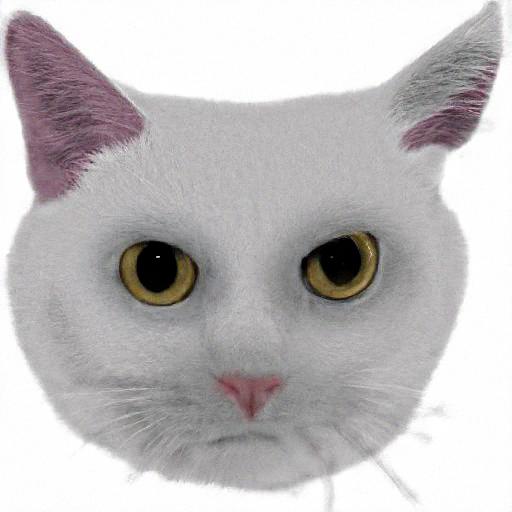} &
    \includegraphics[height=.135\linewidth,width=.135\linewidth]{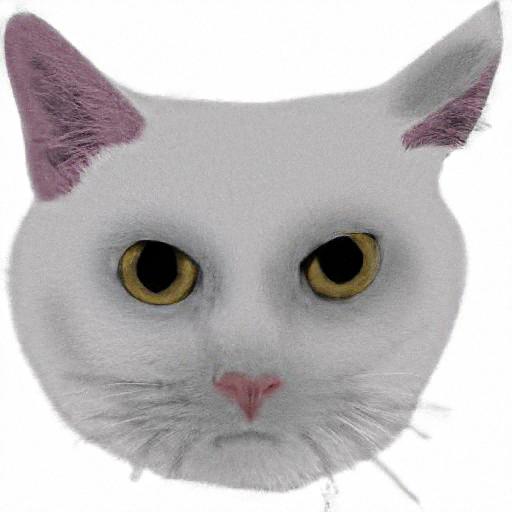} &
    \includegraphics[height=.135\linewidth,width=.135\linewidth]{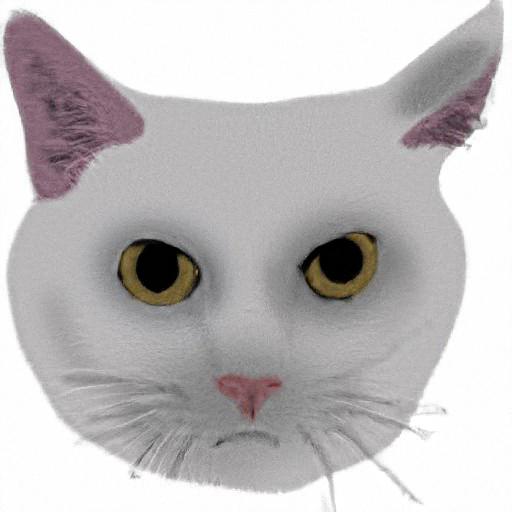} \\
    \rowcolor[rgb]{0.753,0.753,0.753} LPIPS & 0.509 & 0.451 & 0.417 & 0.400 & 0.348 & 0.310 \\
    & & & & & & \\
    \includegraphics[height=.135\linewidth,width=.135\linewidth]{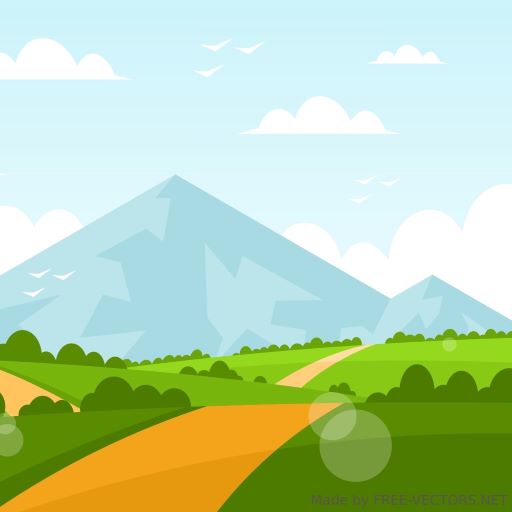} &
    \includegraphics[height=.135\linewidth,width=.135\linewidth]{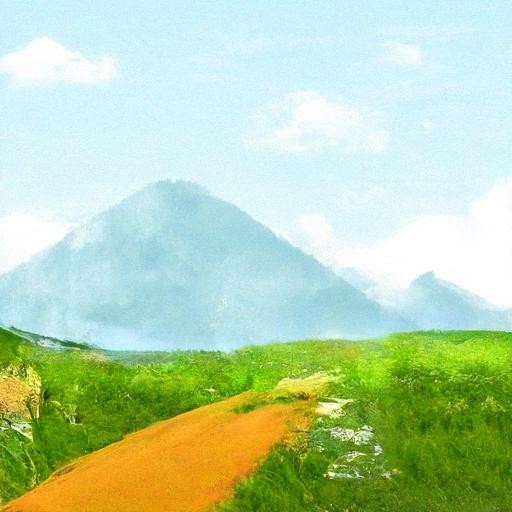} &
    \includegraphics[height=.135\linewidth,width=.135\linewidth]{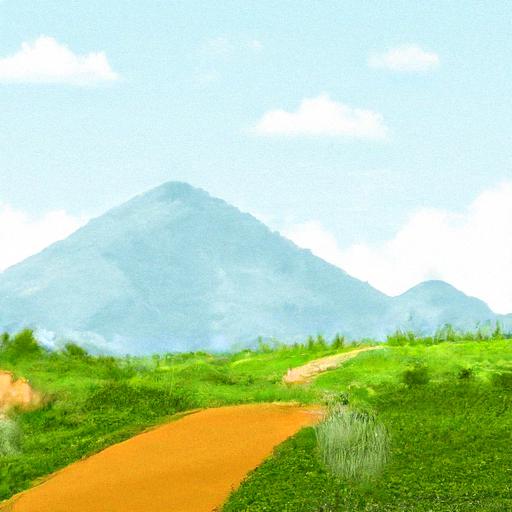} &
    \includegraphics[height=.135\linewidth,width=.135\linewidth]{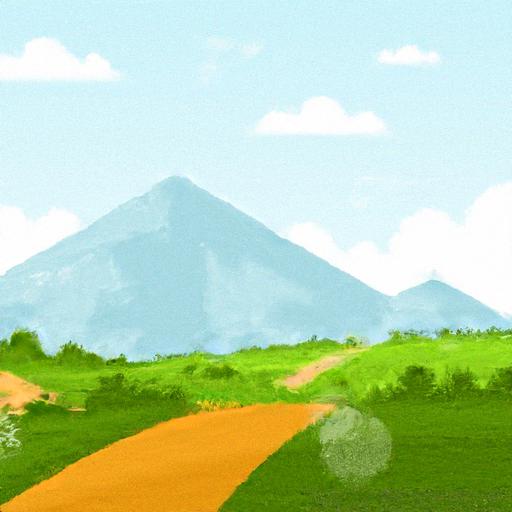} &
    \includegraphics[height=.135\linewidth,width=.135\linewidth]{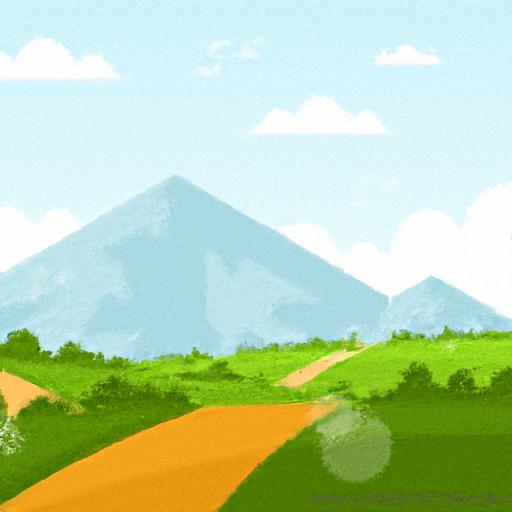} &
    \includegraphics[height=.135\linewidth,width=.135\linewidth]{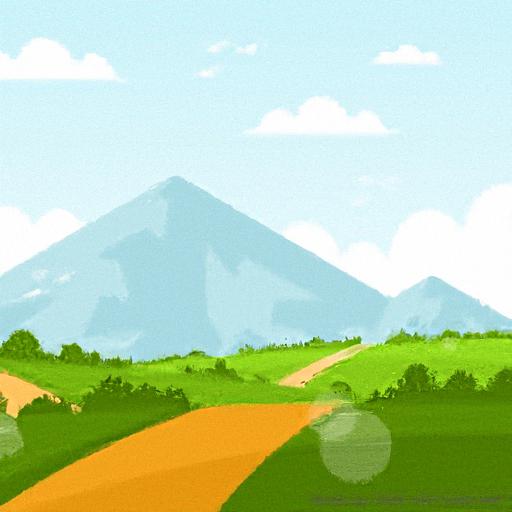} &
    \includegraphics[height=.135\linewidth,width=.135\linewidth]{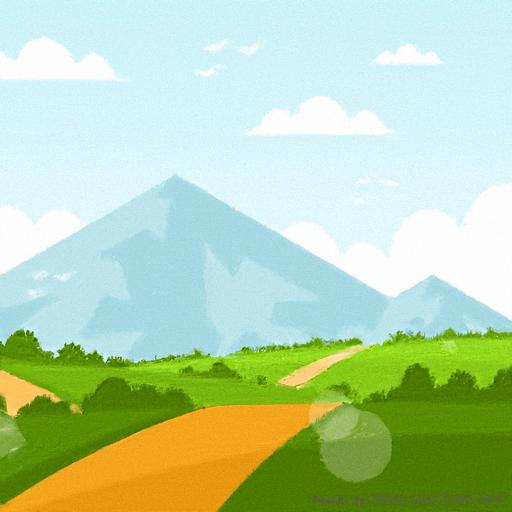} \\
    \rowcolor[rgb]{0.753,0.753,0.753} LPIPS & 0.421 & 0.364 & 0.271 & 0.232 & 0.190 & 0.158 \\
    \end{tabular}
    \vspace{\figmargin}
\caption{
\textbf{Trade-off between realism and consistency to image guidance.}
We demonstrate the trade-off between the image realism and the correspondence to the input guidance, where the realism scale is varied from low ($0.0$, \textit{right}) to high ($1.0$, \textit{left}).
We also show the LPIPS scores between the generated image and the input guidance.
Both the object-level (a cat drawing\protect\footnotemark) and scene-level (a landscape painting\protect\footnotemark) input guidance images are used in this experiment.
}
\label{fig:trade-off-realism}
\vspace{\figmargin}
\vspace{-3mm}
\end{figure*}

\footnotetext{Source from https://freesvg.org/kocka}
\footnotetext{Source from https://free-vectors.net/nature/green-field-vector}
\subsection{Qualitative Evaluation}
\vspace{\subsecmargin}
\paragraph{Adaptively-realistic image generation from sketch and stroke .}
We present the qualitative comparisons between the proposed \modelName and other methods in Figure~\ref{fig:visual-qual}.
Compared to the other frameworks, the proposed \modelName approach produces more realistic results on the object-level (cats and flowers) and scene-level (landscapes) datasets.
Moreover, the images generated by our scheme faithfully correspond to the input contour and color information.
%
%
It is also noteworthy that our method is robust to different levels of details provided by the contour image.
For example, in the second row of Figure~\ref{fig:visual-qual}, the proposed \modelName still synthesizes photo-realistic result even the contour image does not indicate the eye position of the cat.
%
%
%
Finally, we demonstrate the variation produced by changing the three controllable scales in Figure~\ref{fig:teaser} (sketch/stroke), Figure~\ref{fig:trade-off-realism} (realism), and Figure~\ref{fig:sketch-stroke-trade-off} (sketch/stroke).
\vspace{-2em}
\paragraph{Multi-modal multi-domain translation.}
As the input only contains rough contour and colored stroke information, our \modelName approach is capable of synthesizing multiple (i.e. multimodal) image generation results (based on different initial randomly-drawn noises $x_T$ and stochastic sampling procedure).
The results are shown in Figure~\ref{fig:visual-multi}.
Note all the images (cats, dogs, and wild animals) are synthesized from the same trained model.
This suggests that the proposed model is able to understand the implicit category information from the input sketches.
%
%
%
In addition to the diverse generation results, since the input sketch and stroke images are not extracted from the same source image, we demonstrate that our method is also robust to the \textit{un-aligned} sketch-stroke input data.

\vspace{-1.5em}
\paragraph{Applications.}
The proposed \modelName approach not only offers the three-dimensional control over the generation process, but also enables two interesting applications: multi-conditioned local editing and region-sensitive stroke-to-image generation.
Note that we do \textit{not} retrain our model, but only design a specific inference algorithm for these two applications.
We provide the details in the supplementary document.
First, we present the visual editing results in Figure~\ref{fig:visual-app} (a).
%
Our model enables flexible local manipulation on an existing image, which refers to both the hand-drawn contour and colored strokes.
%
%
Secondly, we demonstrate the region-sensitive stroke-to-image generation results in Figure~\ref{fig:visual-app} (b).
The proposed approach can take partial-sketch as input and produces results that 1) match the appearance in the region of partial-sketch and 2) exhibit multiple plausible contents in the non-colored region.

\subsection{Quantitative Evaluation}
\vspace{\subsecmargin}
\paragraph{Image quality and correspondence to input sketch.}

We use the Fréchet Inception Distance (FID)~\cite{heusel2017gans} to measure the realism of the generated images.
To evaluate whether the synthesized images correspond to the input sketch, we compute the Learned Perceptual Image Patch Similarity (LPIPS)~\cite{zhang2018unreasonable} score on the \textit{sketch} level.
Specifically, we calculate the similarity between the input sketch and the sketch inferred from the generated image (via Photo-sketching~\cite{li2019photo}).
Lower FID and LPIPS values indicate better perceptual quality and correspondence, respectively.
%
The quantitative results in Table~\ref{tab:fid} show that our method performs favorably against other representative approaches.

\vspace{\paramargin}
\vspace{-1em}
\paragraph{User preference study.}
To further understand the visual quality of images generated from sketches and strokes, we conduct a user study (with more than 80 candidates in total) by pairwise comparison.
We use the results generated from the AFHQ-cat and Landscapes datasets.
Given a randomly-swapped pair of images sampled from real images and images generated from various methods, we ask the participants to choose the image which is \textit{more realistic}.
Figure~\ref{fig:user-study} presents the statistics of users' preferences. 
The results validate the effectiveness of the proposed approach.

\vspace{\paramargin}
\vspace{-1em}
\paragraph{Realism vs. correspondence to input guidance.}
Figure~\ref{fig:trade-off-realism} demonstrates the trade-off between the generated image realism and the correspondence between the generated image and the input guidance.
We change the realism scale from low ($0.0$) to high ($1.0$) in this experiment.
%
%
\begin{figure}[t]
    \centering
    \includegraphics[width=\linewidth]{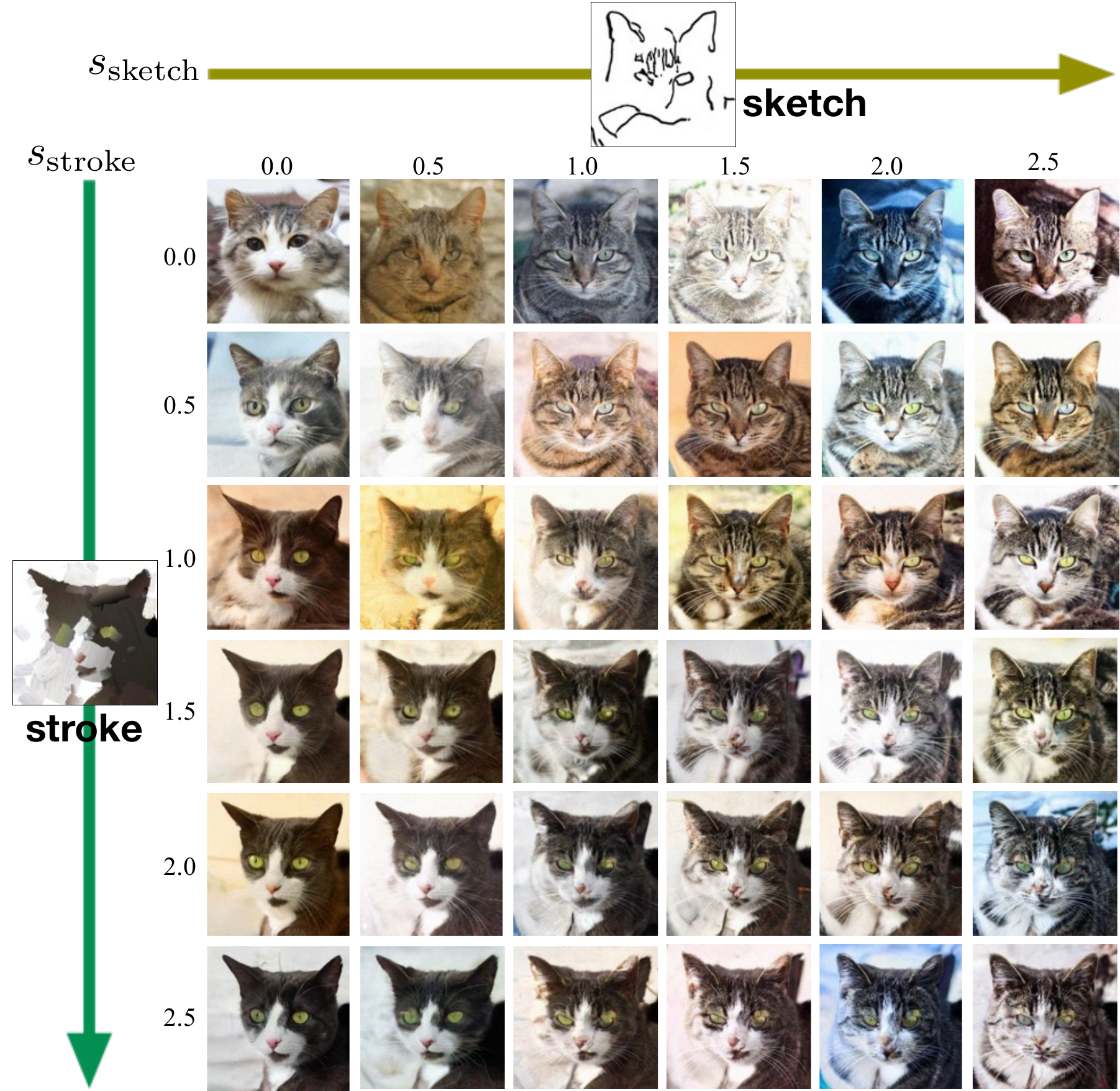}
    \vspace{-5mm}
    \caption{
    \textbf{Qualitative results of using different stroke and sketch scales.}
    The top-left corner show the results generated without guidance.
    Stronger scale values lead to results which are more consistent to the input guidance.
    }
    \label{fig:sketch-stroke-trade-off}
    \vspace{\figmargin}
    \vspace{-4mm}
\end{figure}
\begin{figure}[t]
    \centering
    \includegraphics[width=0.8\linewidth]{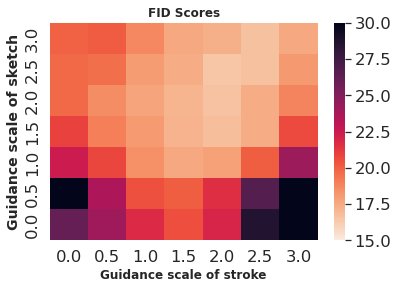}
    \vspace{-2mm}
    \caption{
    \textbf{Impact of various stroke and sketch scales on generated image quality.}
    We report the FID scores to indicate the generated image quality.
    The results suggest that setting the stroke and sketch scale values to be in interval of $[1.5, 2.5]$ lead to the best image quality.
    }
    \label{fig:heatmap}
    \vspace{\figmargin}
    \vspace{-0.5em}
\end{figure}

\vspace{\paramargin}
\vspace{-1em}
\paragraph{Controlling stroke and sketch scales.}
We conduct an ablation study to understand the impact of using different stroke and sketch scales.
Figure~\ref{fig:sketch-stroke-trade-off} shows the qualitative results, and Figure~\ref{fig:heatmap} reports the FID scores computed using the AFHQ-cat dataset.
%
The results show that we can obtain the best generated image quality by setting the sketch and stroke scale values in the interval of $[1.5, 2.5]$.
More ablation study results about stroke, sketch and realism scales are provided in the supplementary document.
%
%
%
\section{Conclusion}
\vspace{\secmargin}
In this work, we introduce \modelName, a versatile and flexible framework that synthesizes photo-realistic images from the sketch and colored stroke guidance.
Our method uses 1) two-directional classifier-free guidance and 2) iterative latent variable refinement to offer the three-dimensional control (sketch, colored stroke, realism) over the image generation process.
Extensive experimental results verify the effectiveness of the proposed approach against several representative schemes.
%
%
%
%
Furthermore, we demonstrate that the proposed \modelName framework enables more interesting applications, such as mask-free local editing and region-sensitive stroke-to-image generation.

\clearpage
\title{Adaptively-Realistic Image Generation\\from Stroke and Sketch with Diffusion Model\\\textit{Supplementary Materials}}
\maketitle
\section{Dataset and Preprocessing Details}
\subsection{Dataset.}
{We consider three datasets: AFHQ~\cite{choi2020stargan}, Landscapes~\cite{skorokhodov2021aligning}, and Oxford Flower~\cite{nilsback2008automated} in our experiments. 
AFHQ dataset has three domains: 5153 training and 500 testing images of``cat'', 4739 training and
500 testing images of ``dog'', and 4738 training and 500 testing images of ``wildlife'' (e.g. tiger, lion, wolf, etc).
Oxford Flower dataset contains 8189 images (split to 7189 images for training and 1000 for testing), and Landscapes (LHQ) dataset has 90000 images (split to 81000 images for training, and 9000 for testing).
}
\subsection{Generating Sketch-stroke Pairs from Given Datasets.}
{ For each image in the aforementioned three datasets, we prepare the corresponding sketch and stroke images for training and testing in our proposed task. \par
%
\paragraph{Sketch Generation.} We utilize the official pretrained model of Photo-sketching~\cite{li2019photo}, a GAN-based network extracting the contour drawings from the given image, to obtain the paired sketch data for images of the datasets used in our experiments.
%
Noting that we perform additional foreground extraction using GrabCut algorithm in OpenCV specifically for the Oxford Flower images before applying the sketch generation, as we find that the images in this dataset usually have many leaves behind the main flowers which would cause some distraction for capturing outlines of the main object.
\paragraph{Stroke Generation.} We do image-to-painting translation via applying two methods, Stylized Neural Painting~\cite{zou2021stylized} and Paint Transformer~\cite{liu2021paint}, in order to generate the paired stroke data for images of the datasets used in our experiments. 
%
Both Stylized Neural Painting and Paint Transformer are state-of-the-art image-to-painting frameworks and able to produce a sequence of meaningful and faithful stroke prediction.
%
Considering that practical stroke images manually created by human users usually contain only coarse features, we generate stroke data by randomly selecting the intermediate canvas produced during the progressive procedure of Stylized Neural Painting or Paint Transformer.
We apply Stylized Neural Painting on the AFHQ dataset and Paint Transformer on the datasets with more images (i.e. Oxford Flower and Landscapes) since Paint Transformer is less time-consuming.}
\subsection{Generating Sketch-stroke Pairs from Custom Input Images.}\label{sec:custom-preprocess}
{
Given the custom input images $\textbf{c}_{\text{comb}}$ which are the synthetic ones (e.g. the input images shown in Figure.7 of our main manuscript), as they are different from the real images (e.g. what we have in the AFHQ, Landscapes, and Oxford Flower datasets), we perform the following operations (different from the way that we described in the previous subsection 1.2) in order to extract black-white sketches $\textbf{c}_{\text{sketch}}$ and colored stroke images $\textbf{c}_{\text{stroke}}$. Firstly, we adopt the GrabCut algorithm in OpenCV to extract and concentrate on the foreground/main object (where the subordinate parts/fragments around the image border, which are likely to be background, are removed). Next, we utilize Canny algorithm to detect the edges, followed by finding the contour information upon the edges via~\cite{suzuki1985topological} (findContours function in OpenCV), to obtain the black-white sketch $\textbf{c}_\text{sketch}$. As for the colored strokes $\textbf{c}_\text{stroke}$, they are generated by making the contour pixels in white upon the input image $\textbf{c}_\text{comb}$.\par
For the two applications (i.e. multi-conditioned local editing and region-sensitive stroke-to-image) unleashed by our proposed \modelName, as users would provide their input $\textbf{c}_{\text{comb}}$ via directly drawing on the top of the original image, we adopt another procedure to extract the sketch $\textbf{c}_\text{sketch}$ and stroke $\textbf{c}_\text{stroke}$: Firstly, we apply a thresholding operation (grayscale value $>50$ to white; otherwise black) on the input to obtain the black-white sketches $\textbf{c}_\text{sketch}$; then, we extract the colored strokes $\textbf{c}_\text{stroke}$ by replacing the black pixels with white color, which is achieved by the bitwise AND operation on the input image $\textbf{c}_\text{comb}$ and a binary mask (thresholding on the saturation value in which those pixels with saturation $>0$ are labelled as 1, otherwise 0).

%
%
%
%
}
\section{Implementation Details}
{
We implement the models with Pytorch. 
The implementation details of \modelName are provided in Section~\ref{DiSS}, in which we describe the network architecture, the settings of hyper-parameters, how we design the computation for the downsampling size \textbf{N} used in realism control, and the algorithm of the adaptively-realistic image generation from stroke and sketch (cf. the first paragraph in Sec. 4.1 of our main manuscript).
For the two applications that our \modelName unleashes, i.e. multi-conditioned local editing and region-sensitive stroke-to-image, we explain the details and provide their algorithms in Section~\ref{application}.
}
\subsection{\modelName} \label{DiSS}
\paragraph{Network Architecture.}
{ We modify the UNet model in~\cite{dhariwal2021diffusion} to realize the posterior prediction in our sketch- and stroke-guided diffusion model (i.e. $\hat{\epsilon}_\theta(x_t, t, \textbf{c}_\text{sketch}, \textbf{c}_\text{stroke})$ in Sec. 3.2 of our main manuscript).
%
The basic UNet model is constructed with a sequence of residual layers and downsampling convolutions as encoder, followed by a sequence of residual layers and the corresponding upsampling convolutions as decoder, with skip connections linking the intermediate layers with the same spatial size. 
After firstly being proposed in~\cite{ho2020denoising}, the UNet model for diffusions is further improved by~\cite{dhariwal2021diffusion} with higher sampling quality. 
We then modify it by extending the input channel from 3 to 7, which allows the concatenation between the input image and the additional two conditions, i.e. sketches (1-channel) and strokes (3-channel).
} 
\paragraph{Settings of Hyper-parameters.}
{
We apply the same settings of hyper-parameters among the three datasets, as shown below:
\begin{table}[h!]
\begin{tabular}{@{}lc@{}}
\toprule
                     & \multicolumn{1}{l}{\begin{tabular}[c]{@{}l@{}}AFHQ Cat 512$\times$512\\ Oxford Flower 512$\times$512\\ Landscapes-HQ 512$\times$512\end{tabular}} \\ \midrule
Diffusion steps      & 1000                                                                                                               \\
Noise schedule       & linear                                                                                                             \\
Channels             & 128                                                                                                                \\
Depth                & 3                                                                                                                  \\
Channels multiple    & 0.5, 1, 1, 2, 2, 4, 4                                                                                              \\
Heads channels       & 64                                                                                                                 \\
Attention resolution & 32, 16, 8                                                                                                          \\
BigGANup/downsample  & yes                                                                                                                \\
Dropout              & 0.0                                                                                                                \\
Batchsize            & 2                                                                                                                  \\
Learning rate        & 1e-4      \\
\bottomrule

\end{tabular}
\caption{
{{
\textbf{The settings of hyper-parameters among the three datasets.}
We apply the same settings of hyper-parameters for the three datasets on our diffusion models.
}}}
\label{fig:hyperparameters}
\vspace{-2em}
\end{table}
}
\paragraph{Realism Control.}
{
The basic concept of our realism control is mainly inherited from ILVR~\cite{choi2021ilvr}, which utilizes a low-pass filter\protect\footnotemark to operate a downsampling procedure on the given reference $\textbf{c}_\text{comb}$ (from the original image size $\textbf{m} \times \textbf{m}$ to size $\textbf{N} \times \textbf{N}$ and upsampling back to the original size).
The key difference between the realism control of \modelName and ILVR is that the realism control of \modelName permits the downsampling to an arbitrary size \textbf{N} (from Equation 10 in the main manuscript), while ILVR performs downsampling to $\textbf{m}/2^{s}$ ($s$ is a non-negative integer) in which it only allows specific sizes. \par

\footnotetext{https://github.com/assafshocher/ResizeRight}
To enable a continuous realism scale $s_{\text{realism}} \sim [0.0,1.0]$, the higher (respectively lower) values of $s_{\text{realism}}$ should be corresponding to a smaller (respectively larger) downsampling size \textbf{N} in order to realize a trade-off between the realism and the consistency for the synthesized output image.
The computation \textbf{N} with respect to the corresponding realism scale $s_{\text{realism}}$ is provided in Equation 10 of the main manuscript, in which we assume an affine relation between \textbf{N} and $s_{\text{realism}}$, 
i.e. $\textbf{N}=s_{\text{realism}}\times a+b$ where $a$ and $b$ are scalars. In the following we provide the detailed explanation on how such computation is derived.
%
Basically, when $s_{\text{realism}}=1.0$ (the most realistic and the least consistent to $\textbf{c}_\text{comb}$), we would like to force the transformed size $\textbf{N}=1$, which passes the least information of the reference image $\textbf{c}_\text{comb}$ during the filtering;
On the other hand, we make the transformed size $\textbf{N}=\textbf{m}$ (size remains the same, passes the most information of the reference image $\textbf{c}_\text{comb}$ during the filtering) when $s_{\text{realism}}=0.0$ (the least realistic and the most consistent to $\textbf{c}_\text{comb}$). In order to fulfil such purpose, we hence set $a=-(\textbf{m}-1)$ and $b=\textbf{m}$ in which the formulation becomes:
\begin{equation}\label{eqn:not-optimized}
\begin{split}
&\textbf{N} = -s_\text{realism}(\nicefrac{\textbf{m}}{1}-1) + (\nicefrac{\textbf{m}}{1}).
\end{split}
\end{equation}\par
However, in practice, we discover that when applying $s_\text{realism}=[0.0, 0.8]$, the results are highly consistent to $\textbf{c}_\text{comb}$.
Consequently, we substitute the divisor in the formulation above with $8$ to achieve adaptively-photorealistic translation with $s_\text{realism}=[0.0, 1.0]$.
Furthermore, we append a constant term $k$ to adapt on different datasets, $k=0$ for the object-level dataset (AFHQ, flowers) and $k=16$ for the scene-level dataset (landscapes).
The final formulation hence becomes:
\begin{equation}\label{eqn:not-optimized}
\begin{split}
&\textbf{N} = -s_\text{realism}(\nicefrac{\textbf{m}}{8}-1) + (\nicefrac{\textbf{m}}{8})+k
\end{split}
\end{equation}
}

\paragraph{Algorithm.}
{ The detailed algorithm for realizing the overall ``adaptively-realistic image generation from stroke and sketch'', what our \modelName does, is presented in Algorithm~\ref{alg:diss}.}
\begin{algorithm}[htb] 
\caption{\modelName}
\label{alg:diss} 
\begin{algorithmic}[1] 
    \STATE \textbf{Input:} Input custom image $\textbf{c}_{\text{comb}}$
    \STATE \textbf{Output:} Generated image ${x} = {x}_0$ 
    \STATE Extract $\textbf{c}_{\text{comb}} \rightarrow \textbf{c}_\text{sketch}$, $\textbf{c}_\text{stroke}$
	\STATE Sample ${x}_T \sim \mathcal{N}(0, \textbf{I})$
	\STATE \textbf{For} $t = T \text{,...,} 1$ \textbf{do}
	\STATE \;\;\; $\tilde{x}_{t-1} \sim \hat{p}_\theta(\tilde{x}_{t-1} \vert x_{t}, \textbf{c}_\text{sketch}, \textbf{c}_\text{stroke})$
	\STATE \;\;\; $\textbf{c}_{\text{comb}_{t-1}} \sim q(\textbf{c}_{\text{comb}_{t-1}} \vert \textbf{c}_{\text{comb}_0})$ \;\;\;\; 
	\COMMENT{$\textbf{c}_{\text{comb}_0} = \textbf{c}_{\text{comb}}$}
	\STATE \;\;\; $x_{t-1} \leftarrow \tilde{x}_{t-1} - {LP}_\textbf{N}(\tilde{x}_{t-1}) + {LP}_\text{N}( \textbf{c}_{\text{comb}_{t-1}})$
	\STATE \textbf{End for} 
	\STATE \textbf{Return} $x_0$

\end{algorithmic}
\end{algorithm}

\subsection{Applications} \label{application}
{
As described in the main manuscript, our proposed \modelName without any model retraining is able to unleash two applications, i.e. multi-conditioned local editing and region-sensitive stroke-to-image, in which here we provide their detailed descriptions in the following as well as their algorithms (Algorithm~\ref{alg:multi-conditioned-local-editing} and~\ref{alg:region-sensitive-stroke-to-image} respectively). 
\begin{itemize}
    \item \textbf{Multi-conditioned Local Editing.} 
    { Our proposed method can naturally realise both sketch and stroke local editing simultaneously on an existing image. 
    As illustrated in Algorithm~\ref{alg:multi-conditioned-local-editing}, we can handle an real image with guided local sketch and stroke drew on top of it as input, via extracting the corresponding sketch part and stroke part followed by performing sketch and stroke modifications with our proposed two-directional classifier-free guidance and realism control. 
    We achieve this application mainly owing to the three-dimensional control technique that enables both sketch and stroke guidance while the realism control keeps the unmodified region of images and enhance the edited parts at the same time.}
    \item \textbf{Region-sensitive stroke-to-image.} 
    { \modelName enables the partial color-conditioning on the specified regions and provides variations on the white colored regions. 
    Algorithm~\ref{alg:region-sensitive-stroke-to-image} shows that after performing the two-directional classifier-free guidance, we apply an additional step before latent variable refinement at each time step to enforce the consistency of partial stroke. 
    Specifically, we mask the non-colored regions to allow the variations and append the noisy colored information to the current latent image (line 9).
    Also note that here we use the stroke condition only, instead of the combination of sketch and stroke, to refine the current step image to further strengthen the partial stroke guidance.}
\end{itemize}
}
\vspace{-1em}
\section{User-Friendly Interface of \modelName}
\begin{figure*}[ht!]
    \centering
    \includegraphics[width=.85\textwidth]{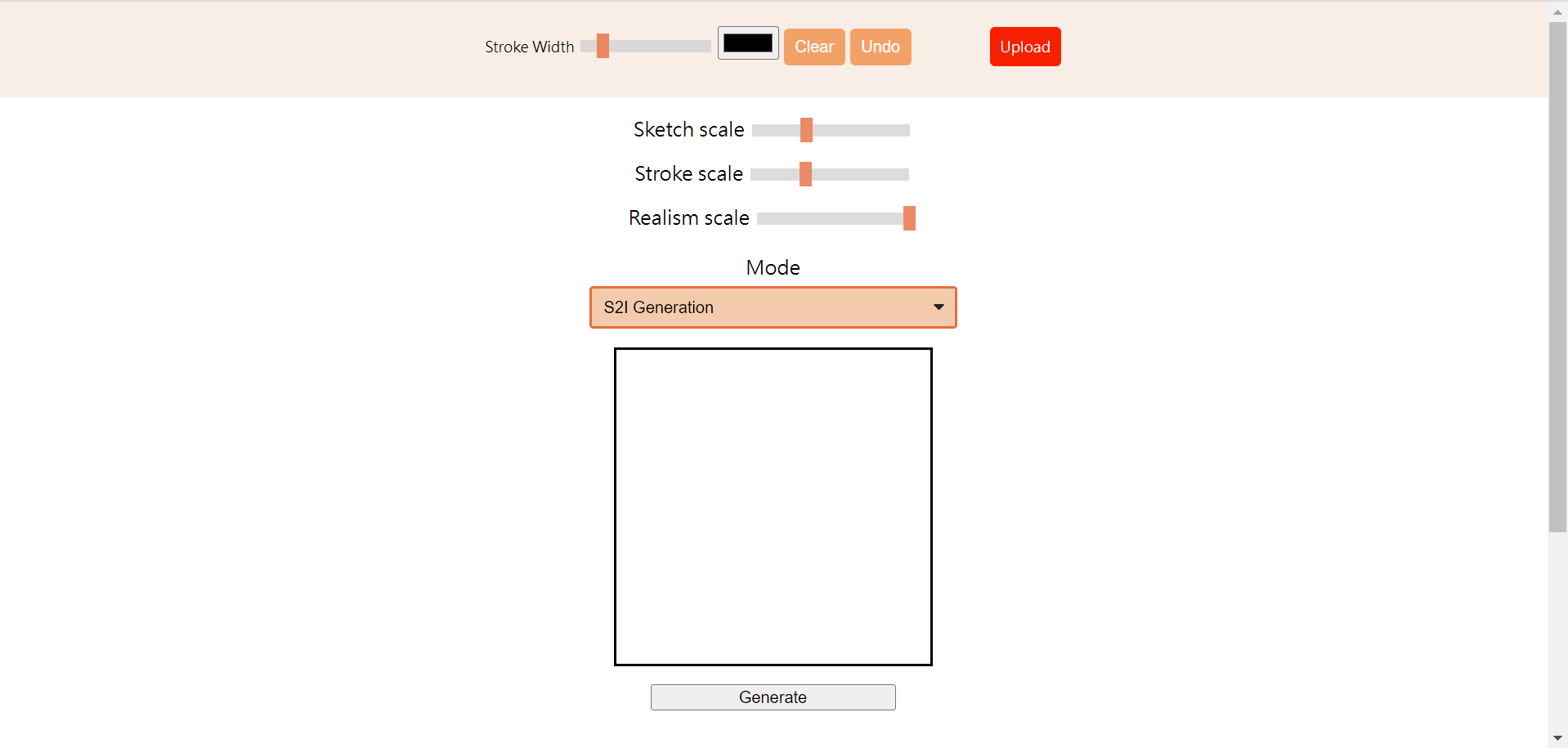}
    \includegraphics[width=.85\textwidth]{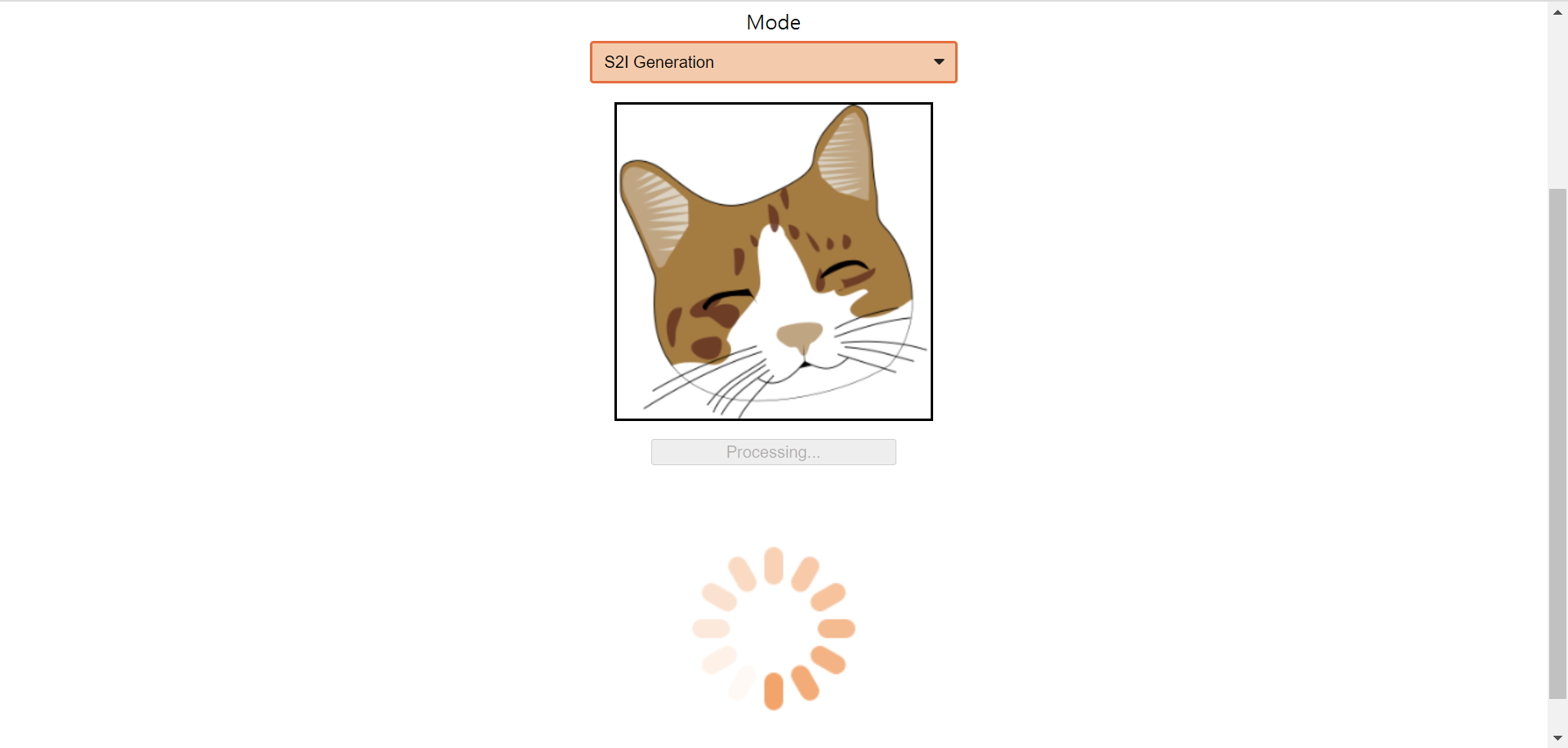}
    \includegraphics[width=.85\textwidth]{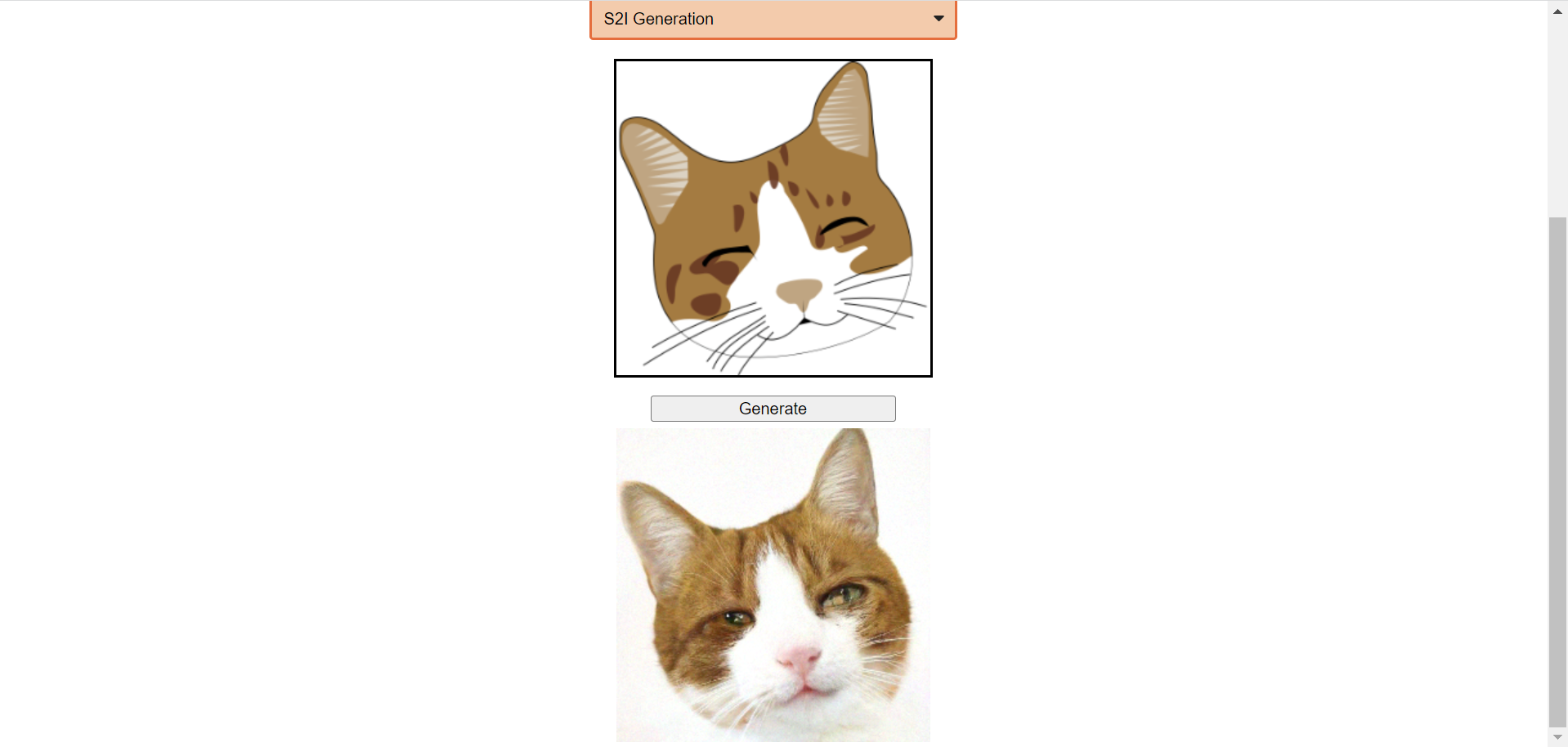}
    \caption{
    \color{black}{
    \textbf{\modelName User Interface.}
    We build a user interface enabling all the applications of \modelName. The screenshots from top to bottom respectively show the initial state, the processing state after creating a drawing, and the final generation result.}}
    \label{fig:interface}
\end{figure*}
\begin{figure*}[t!]
    \centering
    \includegraphics[width=.85\linewidth]{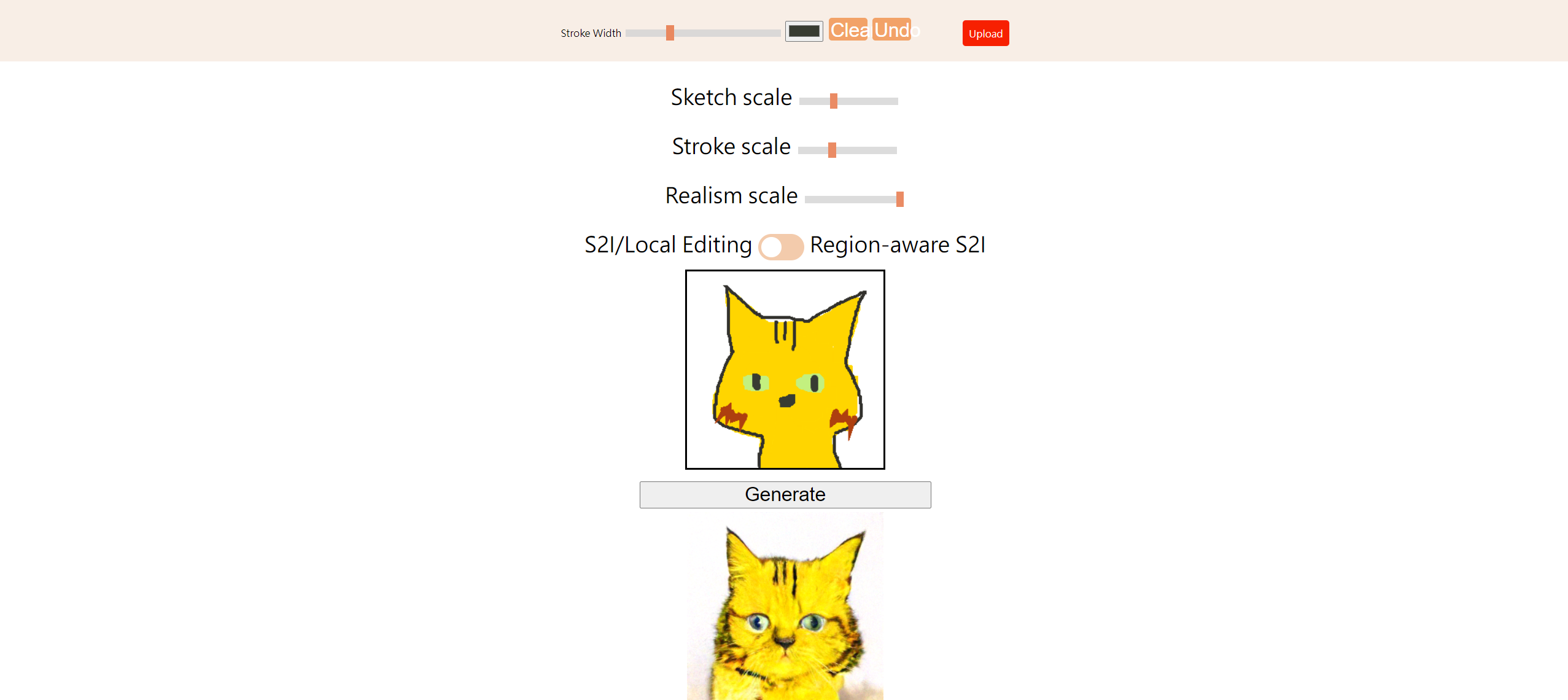}

    \caption{
    \textbf{\modelName User Interface with Hand-Drawing.}
    We play DiSS on the user interface by drawing on our own.}
    \label{fig:interface-hand-drawing}
    \vspace{2em}
\end{figure*}
{
We provide a user-friendly webpage that enables user to create their own drawing and performs all \modelName applications (three dimensional control image generation, multi-conditioned local editing, and region-sensitive stroke-to-image).
We present the screenshots of running our user interface in Figure~\ref{fig:interface}, which sequentially demonstrates the initial state (cf. top subfigure), processing state after user creating a drawing, and the final generated result.
Our interface allows the users to draw on their own (an example of this is provided in Figure~\ref{fig:interface-hand-drawing}) or simply upload an image.
By clicking ``Generate" button, the loading icon will show up and \modelName starts processing the input image (cf. the subfigure in the middle).
The results will show on the webpage when the generation is done, the screenshot is shown in the bottom subfigure.
}

\begin{algorithm*}[ht!] 
\caption{Application: Multi-conditioned Local Editing.}
\label{alg:multi-conditioned-local-editing} 
\begin{algorithmic}[1] 
    \STATE \textbf{Input:} Original image with hand-drawing editing, $\acute{x}$
    \STATE \textbf{Output:} Edited image ${x}_\text{edited} = {x_\text{edited}}_0$ 
    \STATE Extract $\acute{x} \rightarrow \textbf{c}_\text{sketch}$: local sketch editing, $\textbf{c}_\text{stroke}$: stroke editing + the original image
    \STATE $R$: range of the timestep refinement
	\STATE Sample ${x_\text{edited}}_T \sim \mathcal{N}(0, \textbf{I})$
	\STATE \textbf{For} $t = T \text{,...,} 1$ \textbf{do}
	\STATE \;\;\; $\tilde{x}_{\text{edited}_{t-1}} \sim \hat{p}_\theta(\tilde{x}_{\text{edited}_{t-1}} \vert x_{\text{edited}_t}, \textbf{c}_\text{sketch}, \textbf{c}_\text{stroke})$
	\STATE \;\;\; \textbf{If} $t > R$ \textbf{do}
	\STATE \;\;\; \;\;\; $\textbf{c}_{\text{comb}_{t-1}} \sim q(\textbf{c}_{\text{comb}_{t-1}} \vert \textbf{c}_{\text{comb}_0})$ \;\;\;\; 
	\COMMENT{$\textbf{c}_{\text{comb}_0} = \textbf{c}_\text{comb}$}
	\STATE \;\;\; \;\;\; $x_{\text{edited}_{t-1}} \leftarrow \tilde{x}_{\text{edited}_{t-1}} - {LP}_\textbf{N}(\tilde{x}_{\text{edited}_{t-1}}) + {LP}_\text{N}( \textbf{c}_{\text{comb}_{t-1}})$
	\STATE \textbf{End for} 
	\STATE \textbf{Return} ${x_\text{edited}}_0$

\end{algorithmic}
\end{algorithm*}
\begin{algorithm*}[ht!] 
\caption{Application: Region-sensitive Stroke-To-Image.} 
\label{alg:region-sensitive-stroke-to-image} 
\begin{algorithmic}[1] 
    \STATE \textbf{Input:} Guided condition $\textbf{c}_\text{comb}$ with sketch and partial colored stroke
    \STATE \textbf{Output:} Region-sensitive stroke-to-image $x = x_0$ 
	\STATE Extract $\textbf{c}_\text{comb}$ $\rightarrow$  $\textbf{c}_\text{sketch}$, $\textbf{c}_\text{stroke}$ (Section~\ref{sec:custom-preprocess})
    \STATE $R$: range of the timestep refinement
	\STATE mask: 0 on colored region of $\textbf{c}_\text{stroke}$; otherwise 1
	\STATE Sample $x_T \sim \mathcal{N}(0, \textbf{I})$
	\STATE \textbf{For} $t = T \text{,...,} 1$ \textbf{do}
	\STATE \;\;\; $\tilde{x}_{t-1} \sim \hat{p}_\theta(\tilde{x}_{t-1} \vert x_t, \textbf{c}_\text{sketch}, \textbf{c}_\text{stroke})$
	\STATE \;\;\; \textbf{If} $t > R$ \textbf{do}
	\STATE \;\;\; \;\;\; ${\textbf{c}_\text{stroke}}_{t-1} \sim q({\textbf{c}_\text{stroke}}_{t-1} \vert \textbf{c}_{\text{stroke}_0})$ 
	\COMMENT{$\textbf{c}_{\text{stroke}_0} = \textbf{c}_\text{stroke}$}
	\STATE \;\;\; \;\;\; $\tilde{x}_{t-1} \leftarrow \text{mask} \times \tilde{x}_{t-1} + (1-\text{mask}) \times {\textbf{c}_\text{stroke}}_{t-1}$ \;\;\;\;  
	\COMMENT{ append $\textbf{c}_{\text{stroke}_{t-1}}$ on $\tilde{x}_{t-1}$}
	\\
	\STATE \;\;\; \;\;\; $x_{t-1} \leftarrow \tilde{x}_{t-1} - {LP}_\textbf{N}(\tilde{x}_{t-1}) + {LP}_\text{N}( \textbf{c}_{\text{stroke}_{t-1}})$
	\STATE \textbf{End for} 
	\STATE \textbf{Return} $x_0$

\end{algorithmic}
\end{algorithm*}
\section{Additional Results}
\begin{figure*}[!ht]
	\centering
	\subfloat[Multi-conditioned Local Editing.]{%
    	\centering
        \setlength\tabcolsep{1.5pt} 
        \begin{tabular}{c:c|cc}
        Original & Input & Multimodal Results \\
        \includegraphics[height=.5\textwidth]{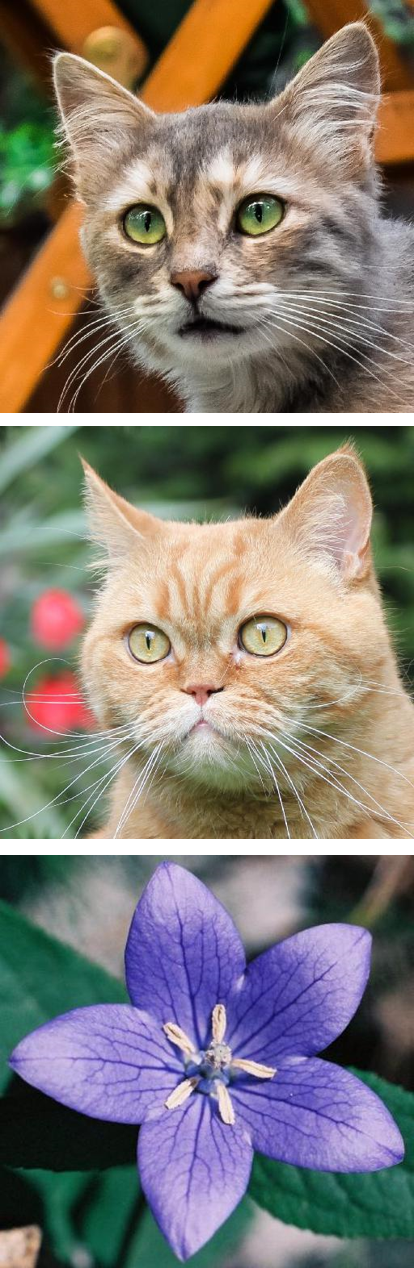} &
        \includegraphics[height=.5\textwidth]{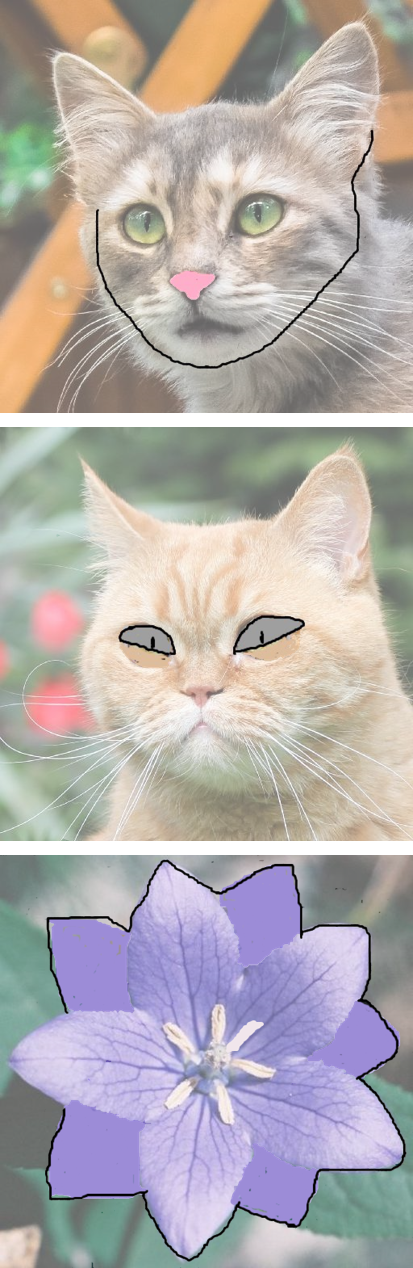} &
        \includegraphics[height=.5\textwidth]{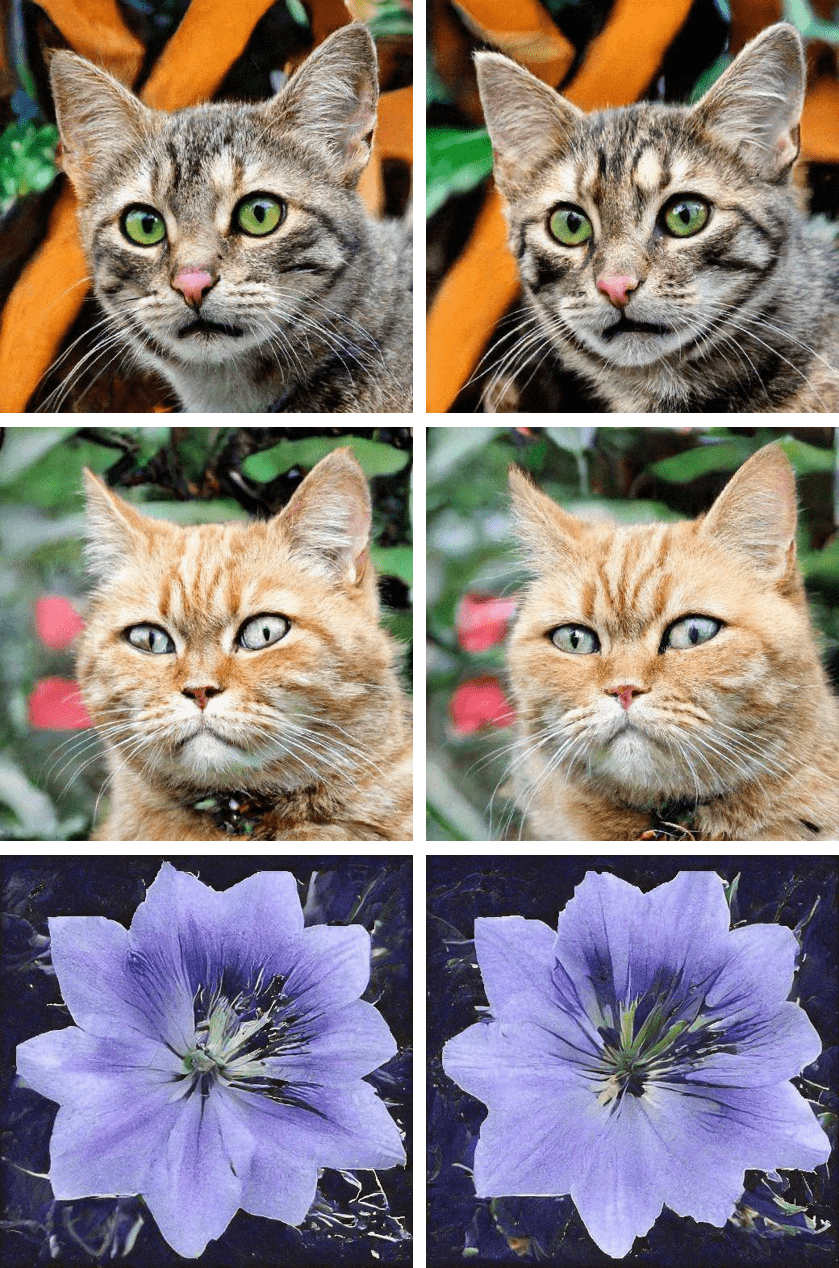} &
        \end{tabular}
	}
	\quad
    \subfloat[Region-sensitive Stroke-to-Image.]{%
        \centering
        \setlength\tabcolsep{1.5pt} 
        \begin{tabular}{c|c}
        Input & Multimodal Results \\
        \includegraphics[height=.5\textwidth]{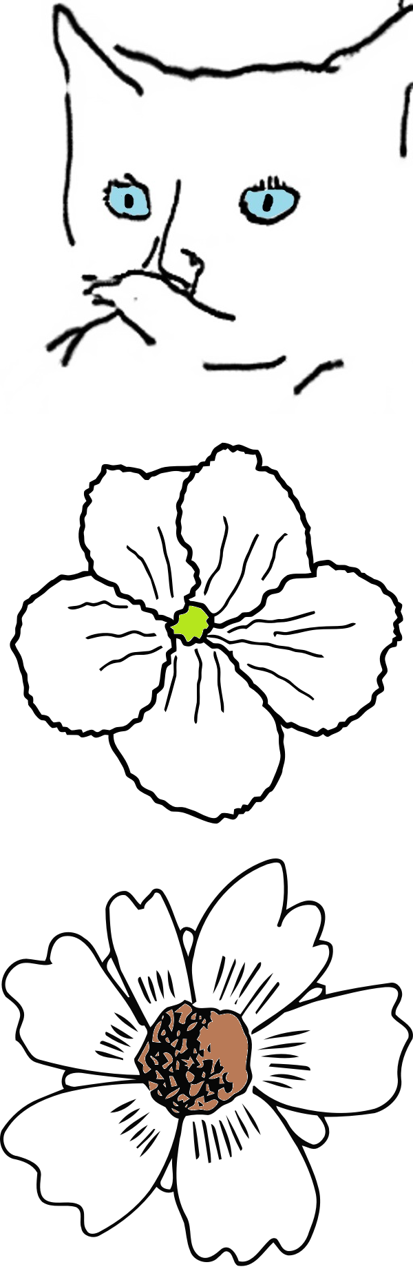} &
        \includegraphics[height=.5\textwidth]{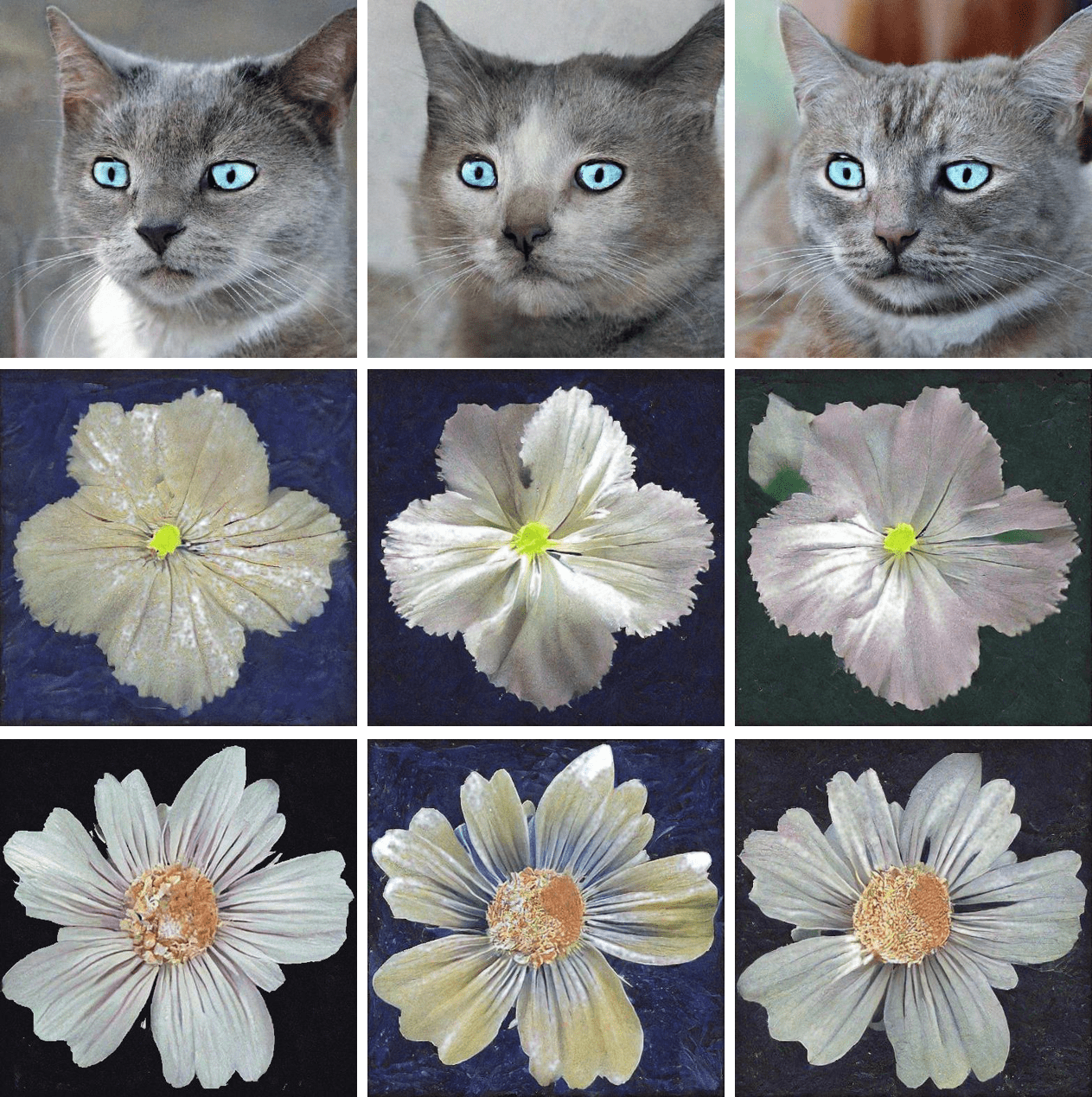} 
        \end{tabular}
	}
    \caption{
    \textbf{More Applications results.} 
    (a) By drawing the new contour or color on an existing image, the proposed model enables the mask-free image editing. 
    (b) With the partial colored stoke as the input, the proposed method synthesizes more diverse contents in the non-colored region.
    Here we use cats and flowers\protect\footnotemark\protect\footnotemark as examples.
    }
    \label{fig:supp-app}
\end{figure*}

\begin{figure*}[t!]
    \centering
    \setlength\tabcolsep{1.5pt} 
    \begin{tabular}{c:cccccc}
    Input & Realism = 1.0 & Realism = 0.8 & Realism = 0.6 & Realism = 0.4 & Realism = 0.2 & Realism = 0.0 \\
    \includegraphics[height=.135\linewidth,width=.135\linewidth]{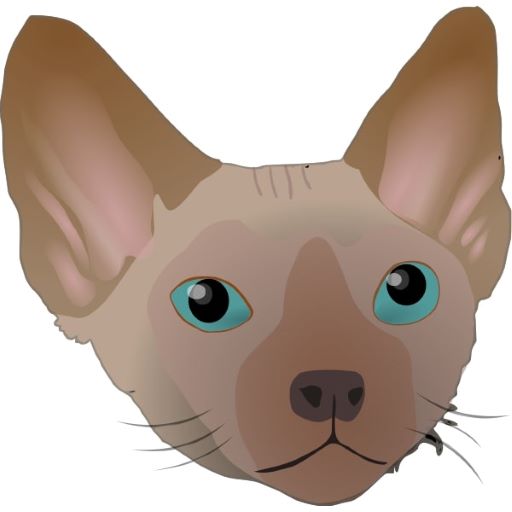} &
    \includegraphics[height=.135\linewidth,width=.135\linewidth]{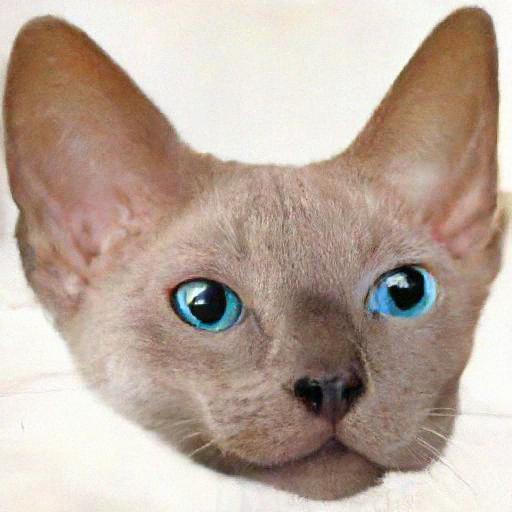} &
    \includegraphics[height=.135\linewidth,width=.135\linewidth]{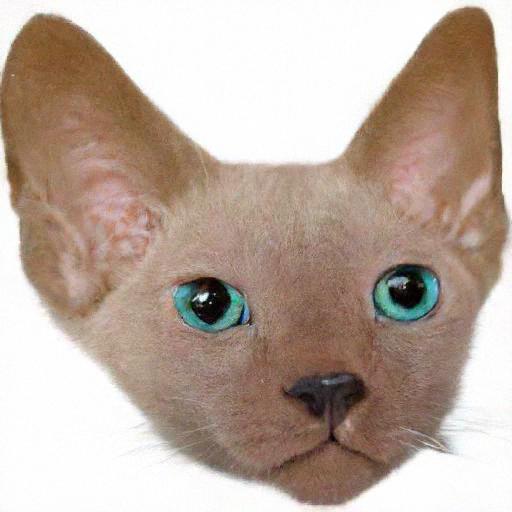} &
    \includegraphics[height=.135\linewidth,width=.135\linewidth]{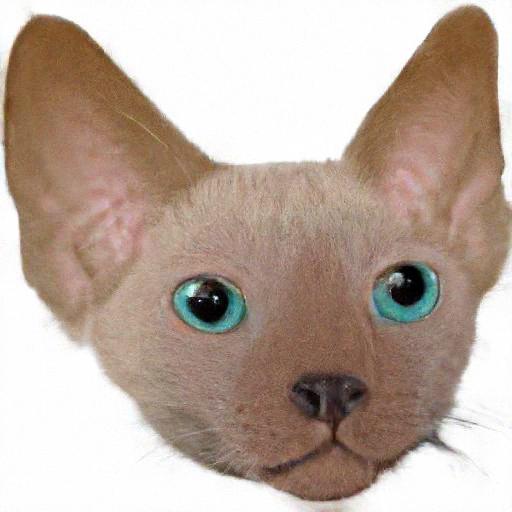} &
    \includegraphics[height=.135\linewidth,width=.135\linewidth]{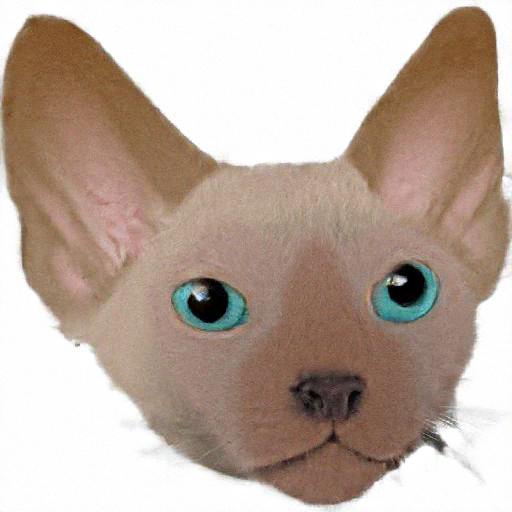} &
    \includegraphics[height=.135\linewidth,width=.135\linewidth]{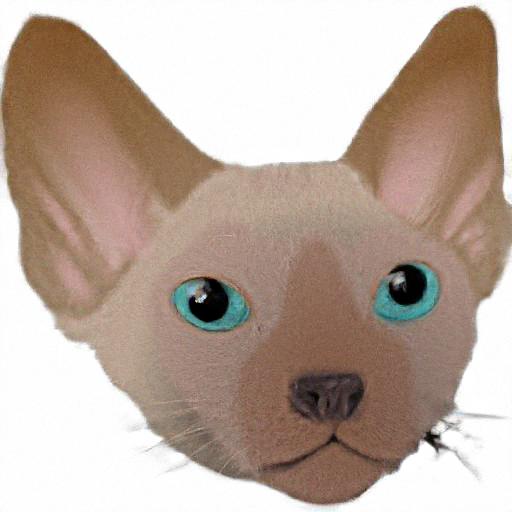} &
    \includegraphics[height=.135\linewidth,width=.135\linewidth]{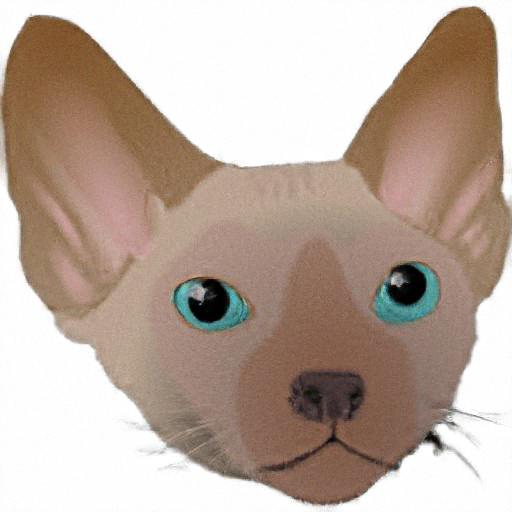} \\
    \rowcolor[rgb]{0.753,0.753,0.753} LPIPS & 0.499 & 0.421 & 0.396 & 0.342 & 0.322 & 0.271 \\
    & & & & & & \\
    \includegraphics[height=.135\linewidth,width=.135\linewidth]{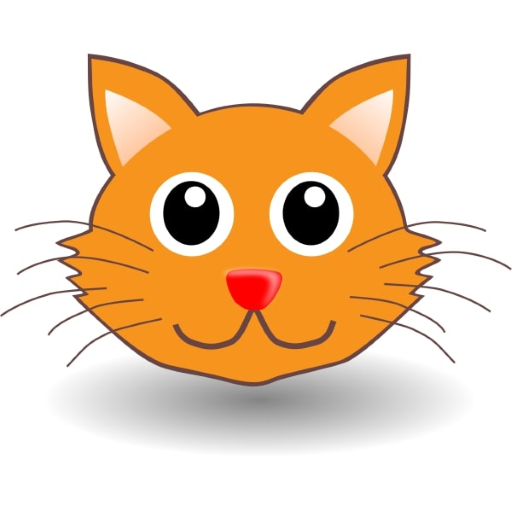} &
    \includegraphics[height=.135\linewidth,width=.135\linewidth]{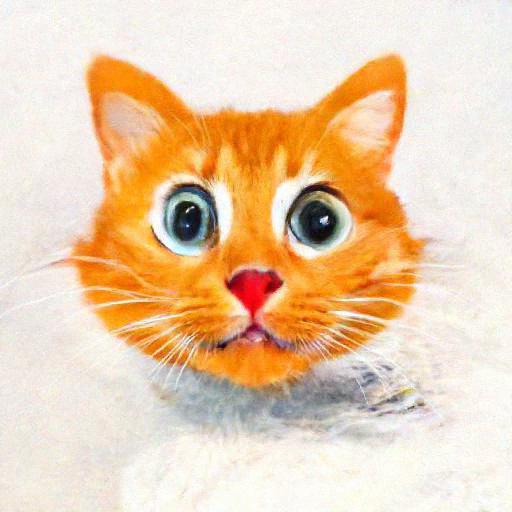} &
    \includegraphics[height=.135\linewidth,width=.135\linewidth]{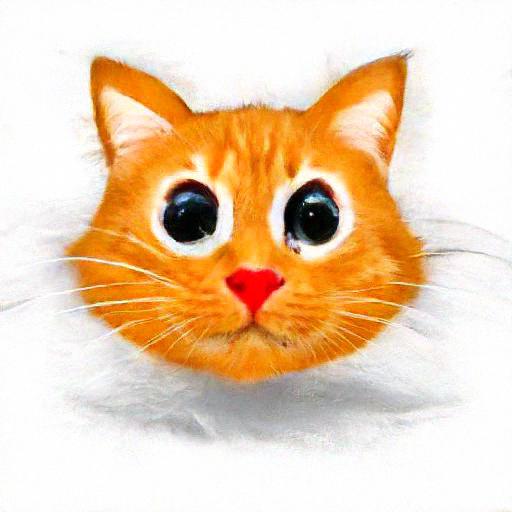} &
    \includegraphics[height=.135\linewidth,width=.135\linewidth]{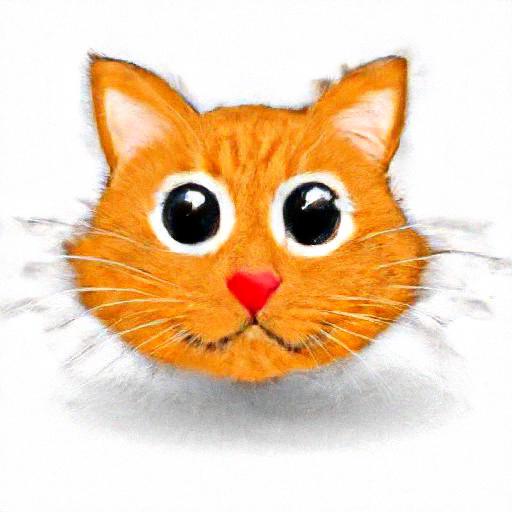} &
    \includegraphics[height=.135\linewidth,width=.135\linewidth]{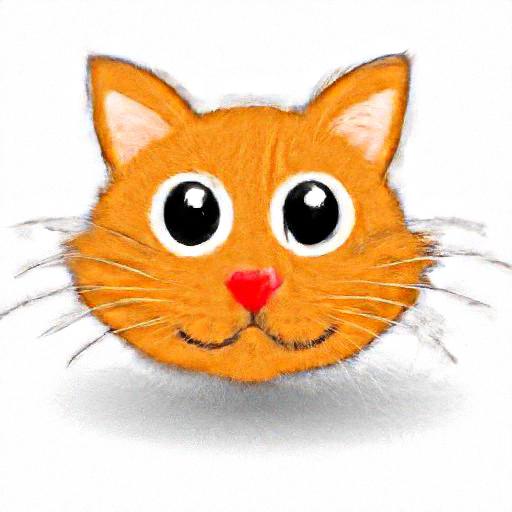} &
    \includegraphics[height=.135\linewidth,width=.135\linewidth]{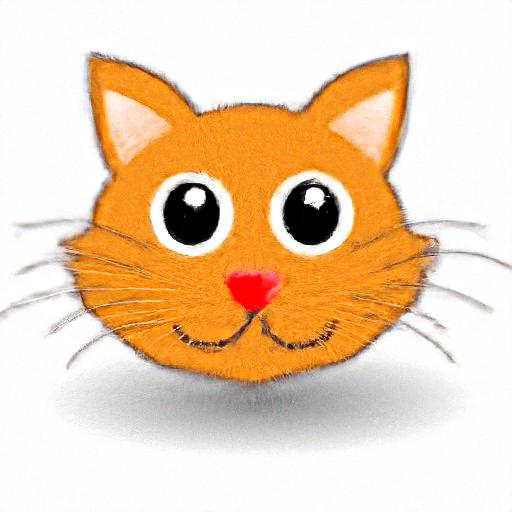} &
    \includegraphics[height=.135\linewidth,width=.135\linewidth]{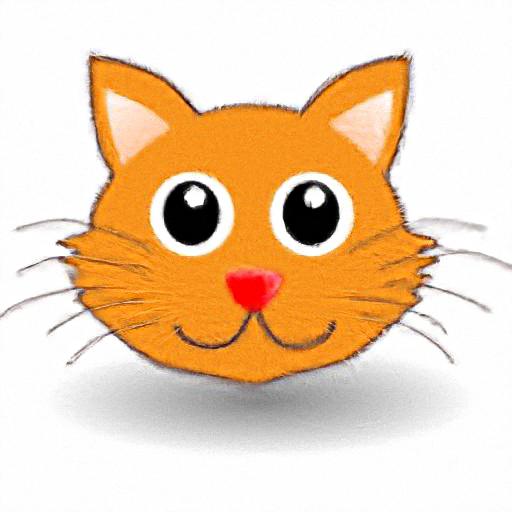} \\
    \rowcolor[rgb]{0.753,0.753,0.753} LPIPS & 0.504 & 0.386 & 0.343 & 0.277 & 0.257 & 0.187 \\
    & & & & & & \\
    \includegraphics[height=.135\linewidth,width=.135\linewidth]{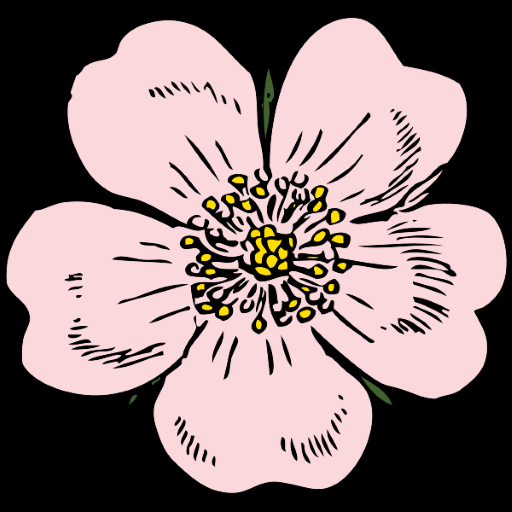} &
    \includegraphics[height=.135\linewidth,width=.135\linewidth]{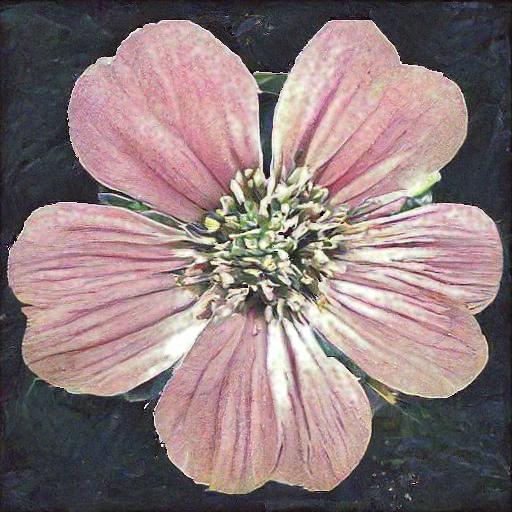} &
    \includegraphics[height=.135\linewidth,width=.135\linewidth]{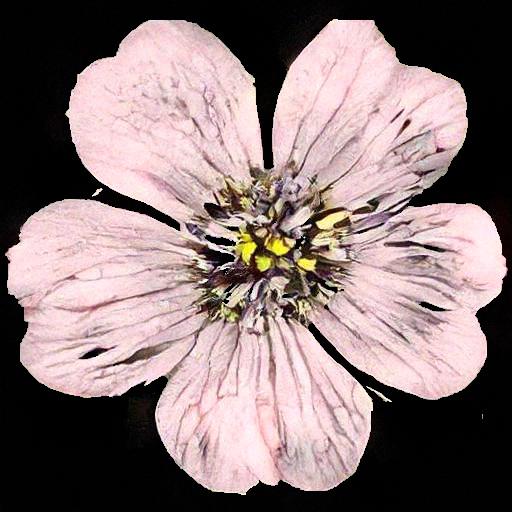} &
    \includegraphics[height=.135\linewidth,width=.135\linewidth]{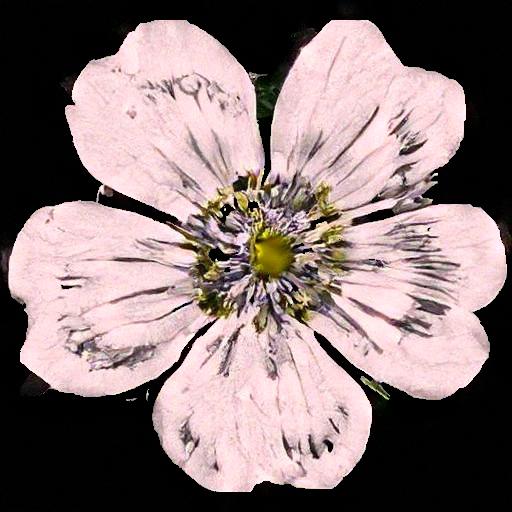} &
    \includegraphics[height=.135\linewidth,width=.135\linewidth]{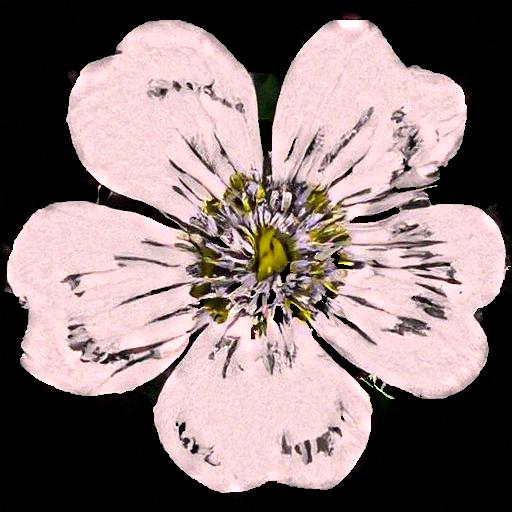} &
    \includegraphics[height=.135\linewidth,width=.135\linewidth]{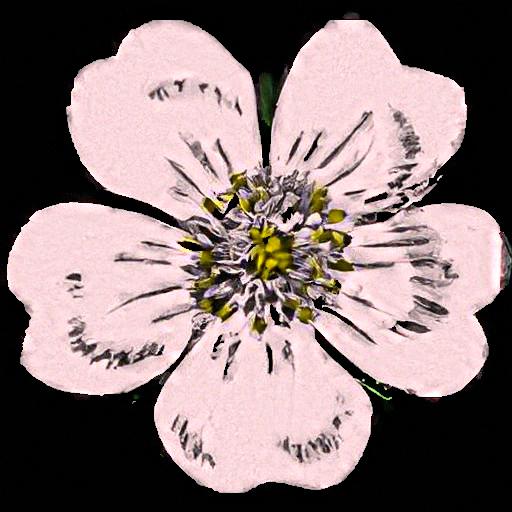} &
    \includegraphics[height=.135\linewidth,width=.135\linewidth]{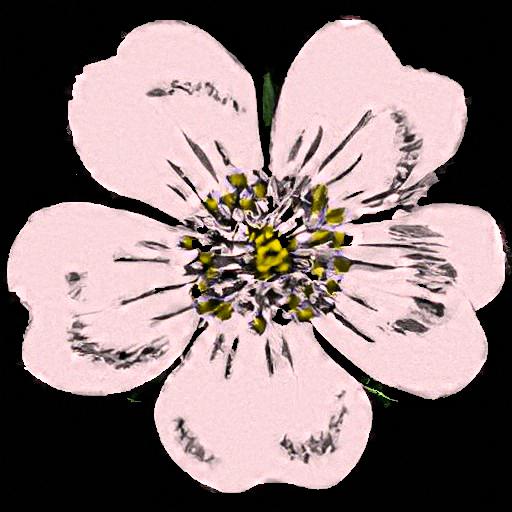} \\
    \rowcolor[rgb]{0.753,0.753,0.753} LPIPS & 0.563 & 0.359 & 0.315 & 0.275 & 0.239 & 0.216 \\
    & & & & & & \\
    \includegraphics[height=.135\linewidth,width=.135\linewidth]{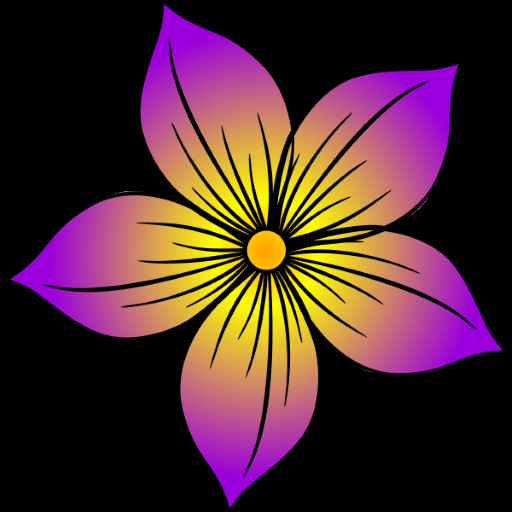} &
    \includegraphics[height=.135\linewidth,width=.135\linewidth]{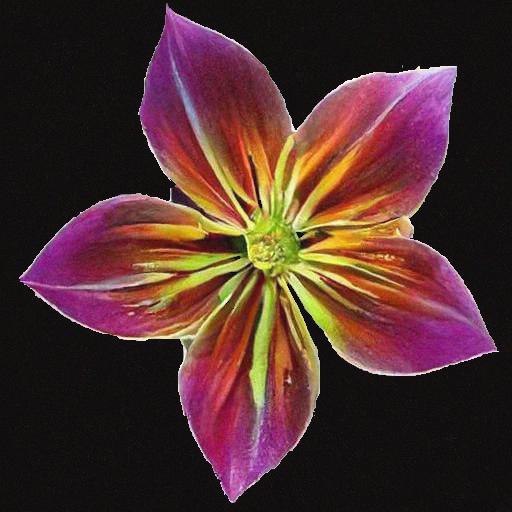} &
    \includegraphics[height=.135\linewidth,width=.135\linewidth]{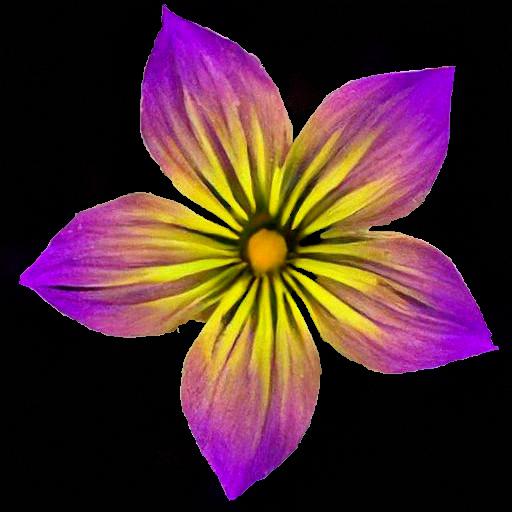} &
    \includegraphics[height=.135\linewidth,width=.135\linewidth]{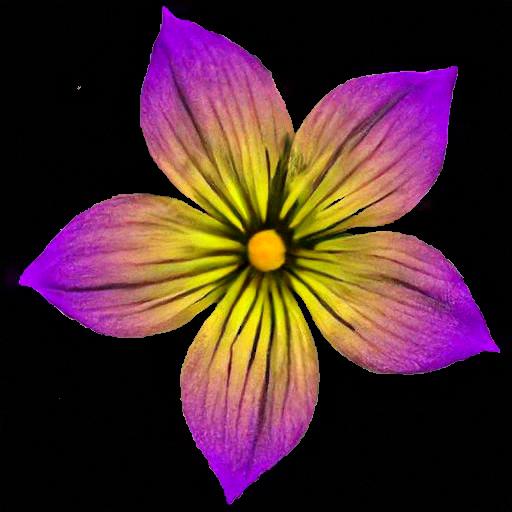} &
    \includegraphics[height=.135\linewidth,width=.135\linewidth]{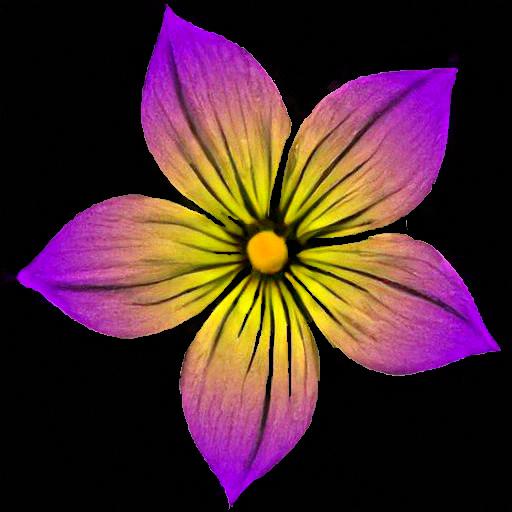} &
    \includegraphics[height=.135\linewidth,width=.135\linewidth]{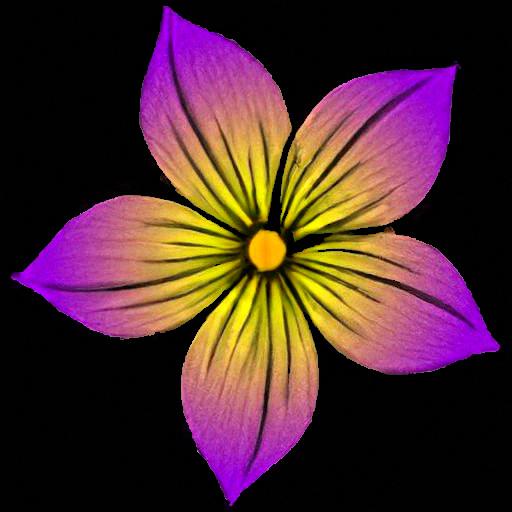} &
    \includegraphics[height=.135\linewidth,width=.135\linewidth]{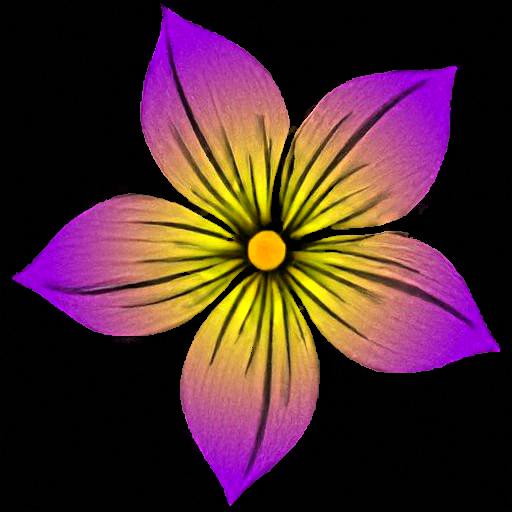} \\
    \rowcolor[rgb]{0.753,0.753,0.753} LPIPS & 0.362 & 0.242 & 0.239 & 0.198 & 0.170 & 0.149 \\
    & & & & & & \\
    \includegraphics[height=.135\linewidth,width=.135\linewidth]{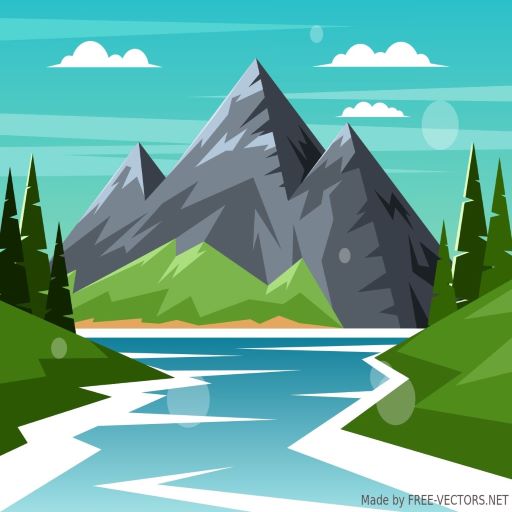} &
    \includegraphics[height=.135\linewidth,width=.135\linewidth]{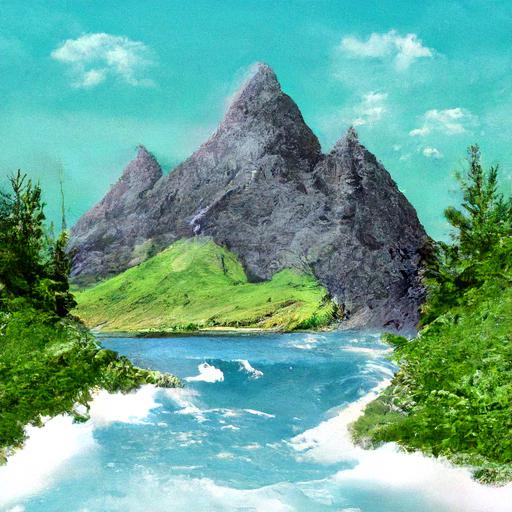} &
    \includegraphics[height=.135\linewidth,width=.135\linewidth]{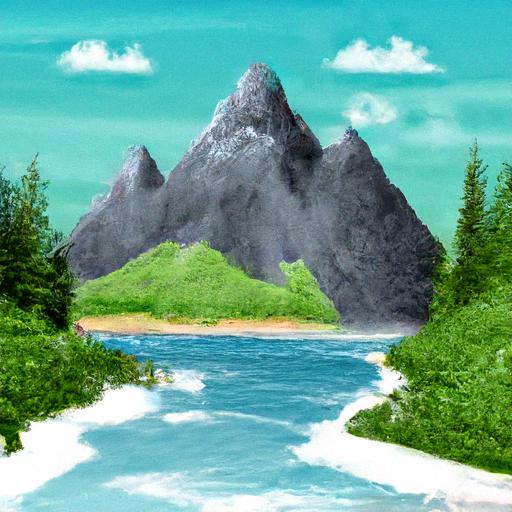} &
    \includegraphics[height=.135\linewidth,width=.135\linewidth]{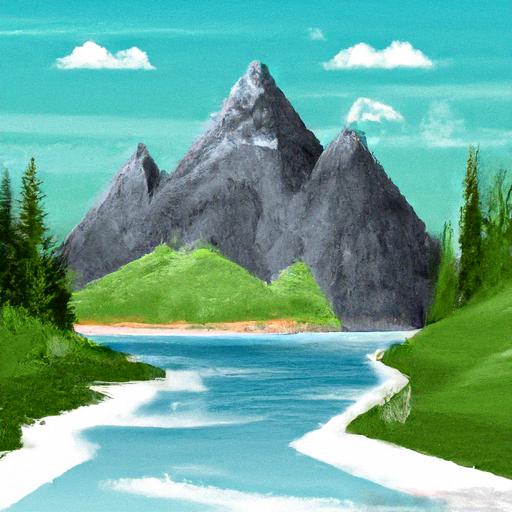} &
    \includegraphics[height=.135\linewidth,width=.135\linewidth]{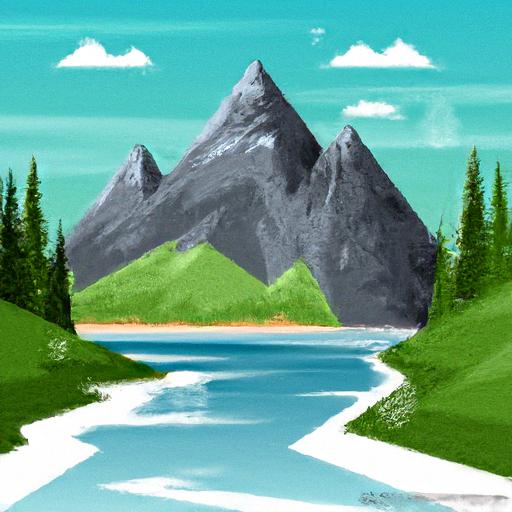} &
    \includegraphics[height=.135\linewidth,width=.135\linewidth]{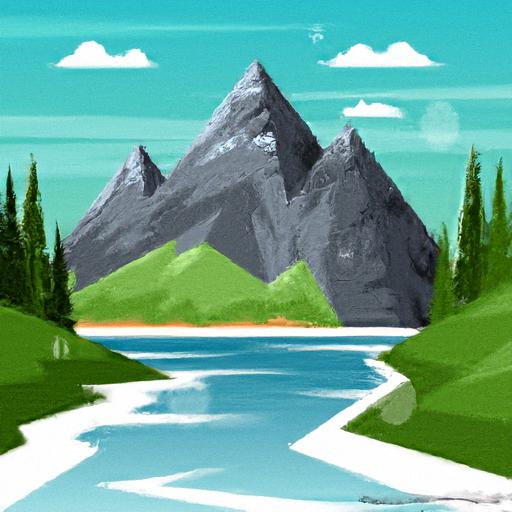} &
    \includegraphics[height=.135\linewidth,width=.135\linewidth]{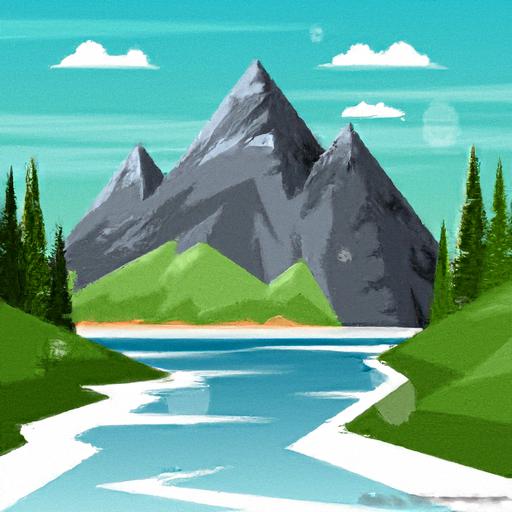} \\
    \rowcolor[rgb]{0.753,0.753,0.753} LPIPS & 0.608 & 0.522 & 0.412 & 0.353 & 0.276 & 0.210 \\
    & & & & & & \\
    \includegraphics[height=.135\linewidth,width=.135\linewidth]{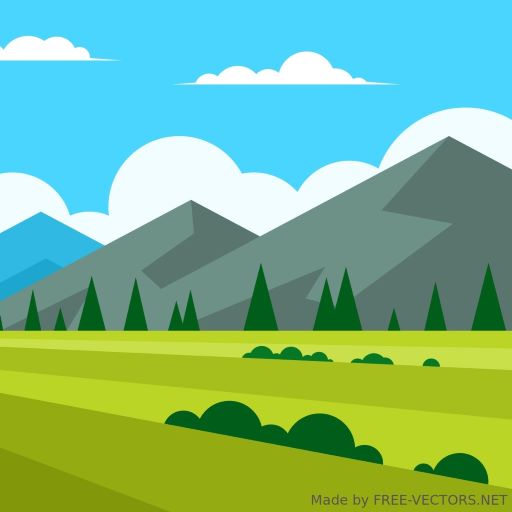} &
    \includegraphics[height=.135\linewidth,width=.135\linewidth]{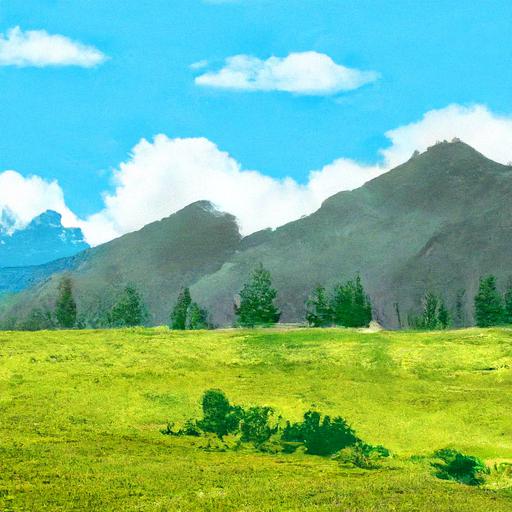} &
    \includegraphics[height=.135\linewidth,width=.135\linewidth]{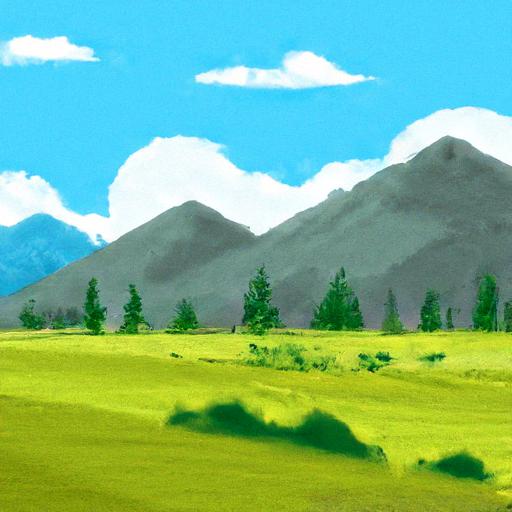} &
    \includegraphics[height=.135\linewidth,width=.135\linewidth]{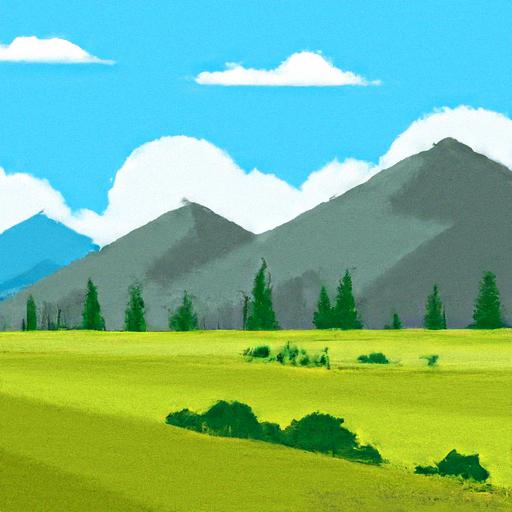} &
    \includegraphics[height=.135\linewidth,width=.135\linewidth]{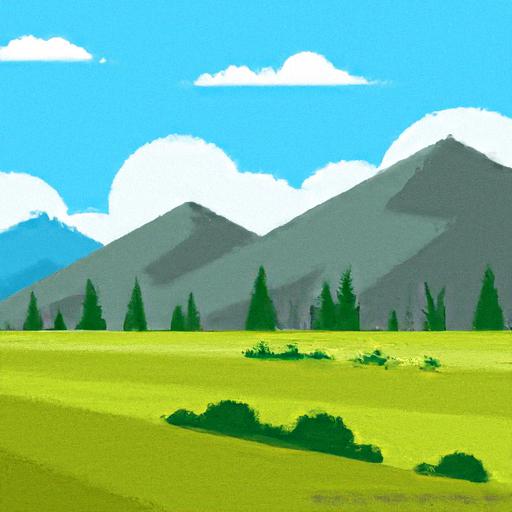} &
    \includegraphics[height=.135\linewidth,width=.135\linewidth]{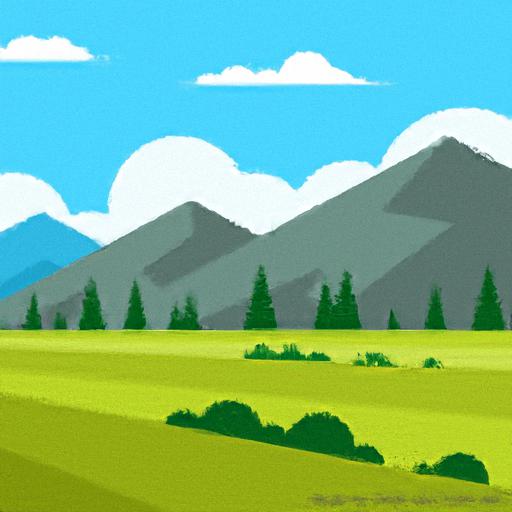} &
    \includegraphics[height=.135\linewidth,width=.135\linewidth]{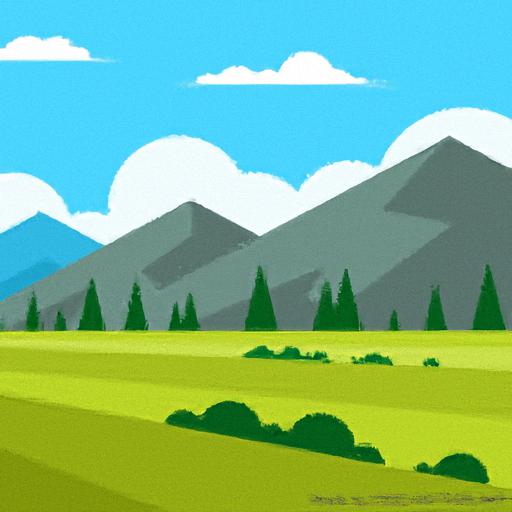} \\
    \rowcolor[rgb]{0.753,0.753,0.753} LPIPS & 0.503 & 0.360 & 0.277 & 0.216 & 0.186 & 0.144 \\
    \end{tabular}
\caption{
\textbf{Trade-off between realism and consistency to image guidance.}
We demonstrate the trade-off between the image realism and the correspondence to the input guidance, where the realism scale is varied from low ($0.0$, \textit{right}) to high ($1.0$, \textit{left}).
We also show the LPIPS scores between the generated image and the input guidance.
Both the object-level and scene-level input guidance images\protect\footnotemark\protect\footnotemark\protect\footnotemark\protect\footnotemark\protect\footnotemark\protect\footnotemark are used in this experiment.
}
\label{fig:supp-trade-off-realism}
\end{figure*}

\begin{figure*}[t!]
	\centering
    	\includegraphics[width=\linewidth]{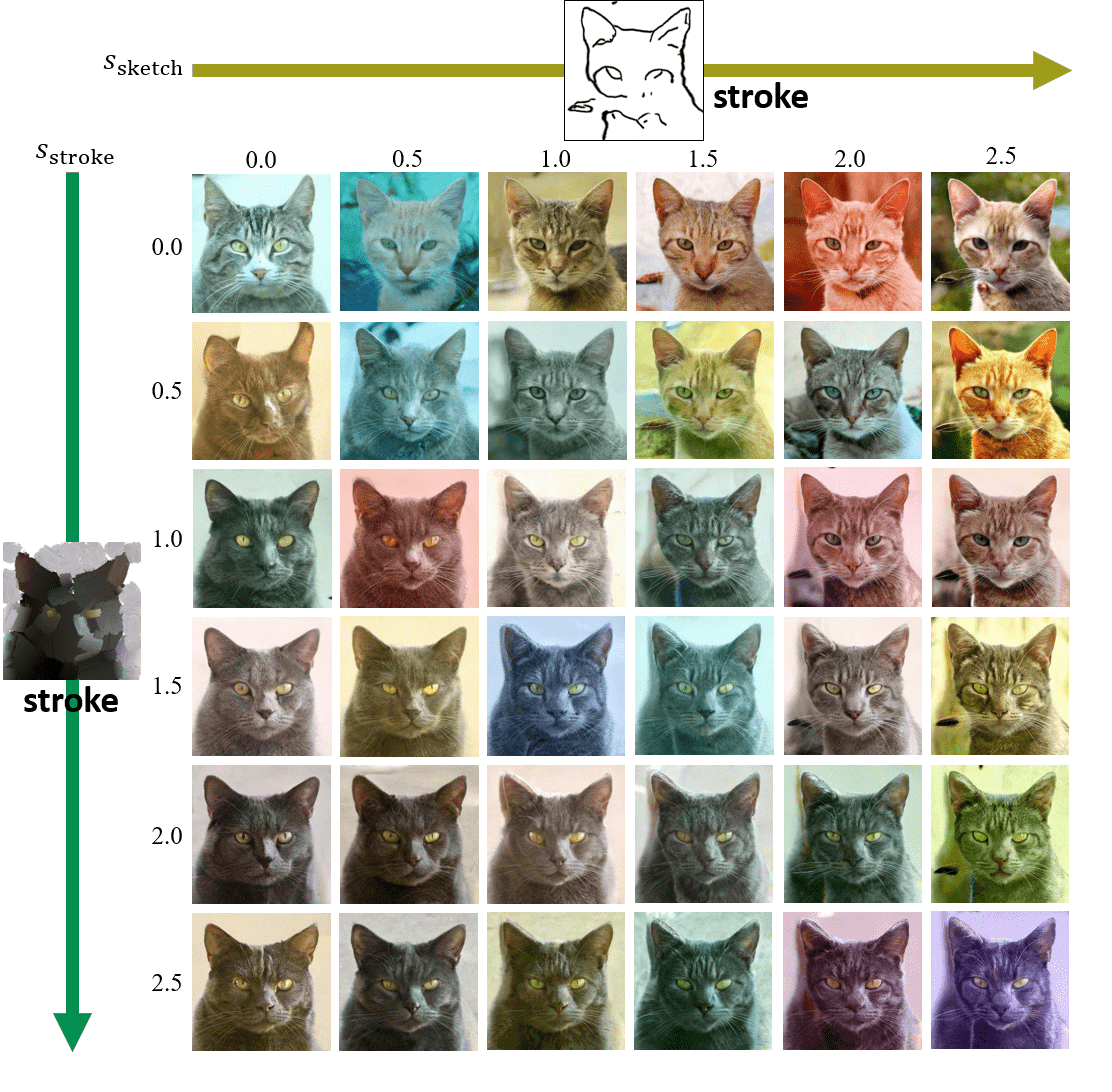}
    \caption{
    \textbf{Qualitative results on AFHQ cat dataset of using different stroke and sketch scales.} The top-left corner show the results generated without guidance. Stronger scale values lead to results
    which are more consistent to the input guidance.} 
    \label{supp-sketch-stroke-tradeoff}
\end{figure*}

More qualitative results of two applications enabled by our proposed \modelName, i.e. multi-conditioned local editing and region-sensitive stroke-to-image, are shown in Figure~\ref{fig:supp-app}. Also, more qualitative results demonstrating the trade-off between realism and consistency, together with the corresponding LPIPS score, are provided in Figure~\ref{fig:supp-trade-off-realism}. Lastly, we provide another set of qualitative results of using different stroke and sketch scales on Lanscapes dataset in Figure~\ref{supp-sketch-stroke-tradeoff}.

\footnotetext[2]{https://freesvg.org/1497040842}
\footnotetext[3]{https://freesvg.org/1528308068-65465}

\footnotetext[4]{https://freesvg.org/dior-2}
\footnotetext[5]{https://freesvg.org/funny-cat-head-vector-illustration}
\footnotetext[6]{https://www.nicepng.com/maxp/u2q8e6e6y3t4r5i1/} 
\footnotetext[7]{https://freesvg.org/1532149926}
\footnotetext[8]{https://free-vectors.net/nature/river-vector}
\footnotetext[9]{https://free-vectors.net/nature/summer-mountain-landscape-vector}

{\small
\bibliographystyle{ieee_fullname}
\bibliography{egbib}

\begin{thebibliography}{10}\itemsep=-1pt

\bibitem{chen2018sketchygan}
Wengling Chen and James Hays.
\newblock Sketchygan: Towards diverse and realistic sketch to image synthesis.
\newblock In {\em IEEE Conference on Computer Vision and Pattern Recognition
  (CVPR)}, 2018.

\bibitem{cheng2020segvae}
Yen-Chi Cheng, Hsin-Ying Lee, Min Sun, and Ming-Hsuan Yang.
\newblock Controllable image synthesis via segvae.
\newblock In {\em European Conference on Computer Vision}, 2020.

\bibitem{choi2021ilvr}
Jooyoung Choi, Sungwon Kim, Yonghyun Jeong, Youngjune Gwon, and Sungroh Yoon.
\newblock {ILVR}: Conditioning method for denoising diffusion probabilistic
  models.
\newblock In {\em IEEE International Conference on Computer Vision (ICCV)},
  2021.

\bibitem{choi2020stargan}
Yunjey Choi, Youngjung Uh, Jaejun Yoo, and Jung-Woo Ha.
\newblock {StarGAN v2}: Diverse image synthesis for multiple domains.
\newblock In {\em IEEE Conference on Computer Vision and Pattern Recognition
  (CVPR)}, 2020.

\bibitem{dhariwal2021diffusion}
Prafulla Dhariwal and Alexander Nichol.
\newblock Diffusion models beat {GAN}s on image synthesis.
\newblock In {\em Advances in Neural Information Processing Systems (NeurIPS)},
  2021.

\bibitem{goodfellow2014generative}
Ian Goodfellow, Jean Pouget-Abadie, Mehdi Mirza, Bing Xu, David Warde-Farley,
  Sherjil Ozair, Aaron Courville, and Yoshua Bengio.
\newblock Generative adversarial nets.
\newblock In {\em Advances in Neural Information Processing Systems (NeurIPS)},
  2014.

\bibitem{heusel2017gans}
Martin Heusel, Hubert Ramsauer, Thomas Unterthiner, Bernhard Nessler, and Sepp
  Hochreiter.
\newblock {GAN}s trained by a two time-scale update rule converge to a local
  nash equilibrium.
\newblock In {\em Advances in Neural Information Processing Systems (NeurIPS)},
  2017.

\bibitem{ho2020denoising}
Jonathan Ho, Ajay Jain, and Pieter Abbeel.
\newblock Denoising diffusion probabilistic models.
\newblock In {\em Advances in Neural Information Processing Systems (NeurIPS)},
  2020.

\bibitem{ho2021classifier}
Jonathan Ho and Tim Salimans.
\newblock Classifier-free diffusion guidance.
\newblock In {\em NeurIPS Workshop on Deep Generative Models and Downstream
  Applications}, 2021.

\bibitem{isola2017image}
Phillip Isola, Jun-Yan Zhu, Tinghui Zhou, and Alexei~A Efros.
\newblock Image-to-image translation with conditional adversarial networks.
\newblock In {\em IEEE Conference on Computer Vision and Pattern Recognition
  (CVPR)}, 2017.

\bibitem{kim2019u}
Junho Kim, Minjae Kim, Hyeonwoo Kang, and Kwanghee Lee.
\newblock {U-GAT-IT}: Unsupervised generative attentional networks with
  adaptive layer-instance normalization for image-to-image translation.
\newblock In {\em International Conference on Learning Representations (ICLR)},
  2020.

\bibitem{DRIT}
Hsin-Ying Lee, Hung-Yu Tseng, Jia-Bin Huang, Maneesh~Kumar Singh, and
  Ming-Hsuan Yang.
\newblock Diverse image-to-image translation via disentangled representations.
\newblock In {\em European Conference on Computer Vision}, 2018.

\bibitem{li2019photo}
Mengtian Li, Zhe Lin, Radomir Mech, Ersin Yumer, and Deva Ramanan.
\newblock Photo-sketching: Inferring contour drawings from images.
\newblock In {\em IEEE Winter Conference on Applications of Computer Vision
  (WACV)}, 2019.

\bibitem{liu2021self}
Bingchen Liu, Yizhe Zhu, Kunpeng Song, and Ahmed Elgammal.
\newblock Self-supervised sketch-to-image synthesis.
\newblock In {\em AAAI Conference on Artificial Intelligence (AAAI)}, 2021.

\bibitem{liu2020unsupervised}
Runtao Liu, Qian Yu, and Stella~X Yu.
\newblock Unsupervised sketch to photo synthesis.
\newblock In {\em European Conference on Computer Vision (ECCV)}, 2020.

\bibitem{liu2021paint}
Songhua Liu, Tianwei Lin, Dongliang He, Fu Li, Ruifeng Deng, Xin Li, Errui
  Ding, and Hao Wang.
\newblock Paint transformer: Feed forward neural painting with stroke
  prediction.
\newblock In {\em IEEE International Conference on Computer Vision (ICCV)},
  2021.

\bibitem{meng2021sdedit}
Chenlin Meng, Yang Song, Jiaming Song, Jiajun Wu, Jun-Yan Zhu, and Stefano
  Ermon.
\newblock {SDEdit}: Image synthesis and editing with stochastic differential
  equations.
\newblock In {\em International Conference on Learning Representations (ICLR)},
  2022.

\bibitem{nichol2021improved}
Alexander~Quinn Nichol and Prafulla Dhariwal.
\newblock Improved denoising diffusion probabilistic models.
\newblock In {\em International Conference on Machine Learning (ICML)}, 2021.

\bibitem{nilsback2008automated}
Maria-Elena Nilsback and Andrew Zisserman.
\newblock Automated flower classification over a large number of classes.
\newblock In {\em Sixth Indian Conference on Computer Vision, Graphics \& Image
  Processing}, 2008.

\bibitem{park2019SPADE}
Taesung Park, Ming-Yu Liu, Ting-Chun Wang, and Jun-Yan Zhu.
\newblock Semantic image synthesis with spatially-adaptive normalization.
\newblock In {\em Proceedings of the IEEE Conference on Computer Vision and
  Pattern Recognition}, 2019.

\bibitem{sangkloy2017scribbler}
Patsorn Sangkloy, Jingwan Lu, Chen Fang, Fisher Yu, and James Hays.
\newblock Scribbler: Controlling deep image synthesis with sketch and color.
\newblock In {\em IEEE Conference on Computer Vision and Pattern Recognition
  (CVPR)}, 2017.

\bibitem{skorokhodov2021aligning}
Ivan Skorokhodov, Grigorii Sotnikov, and Mohamed Elhoseiny.
\newblock Aligning latent and image spaces to connect the unconnectable.
\newblock In {\em IEEE International Conference on Computer Vision (ICCV)},
  2021.

\bibitem{suzuki1985topological}
Satoshi Suzuki et~al.
\newblock Topological structural analysis of digitized binary images by border
  following.
\newblock {\em Computer vision, graphics, and image processing}, 1985.

\bibitem{wang2021sketch}
Sheng-Yu Wang, David Bau, and Jun-Yan Zhu.
\newblock Sketch your own {GAN}.
\newblock In {\em IEEE Conference on Computer Vision and Pattern Recognition
  (CVPR)}, 2021.

\bibitem{zhang2018unreasonable}
Richard Zhang, Phillip Isola, Alexei~A Efros, Eli Shechtman, and Oliver Wang.
\newblock The unreasonable effectiveness of deep features as a perceptual
  metric.
\newblock In {\em IEEE Conference on Computer Vision and Pattern Recognition
  (CVPR)}, 2018.

\bibitem{zhu2017unpaired}
Jun-Yan Zhu, Taesung Park, Phillip Isola, and Alexei~A Efros.
\newblock Unpaired image-to-image translation using cycle-consistent
  adversarial networks.
\newblock In {\em IEEE International Conference on Computer Vision (ICCV)},
  2017.

\bibitem{zhu2017toward}
Jun-Yan Zhu, Richard Zhang, Deepak Pathak, Trevor Darrell, Alexei~A Efros,
  Oliver Wang, and Eli Shechtman.
\newblock Toward multimodal image-to-image translation.
\newblock In {\em Advances in Neural Information Processing Systems}, 2017.

\bibitem{zou2021stylized}
Zhengxia Zou, Tianyang Shi, Shuang Qiu, Yi Yuan, and Zhenwei Shi.
\newblock Stylized neural painting.
\newblock In {\em IEEE Conference on Computer Vision and Pattern Recognition
  (CVPR)}, 2021.

\end{thebibliography}
}

\end{document}